\documentclass[10pt,twocolumn,letterpaper]{article}

\usepackage[pagenumbers]{cvpr}
\usepackage[utf8]{inputenc}
\usepackage[T1]{fontenc}
\usepackage{url}
\usepackage{booktabs}
\usepackage{amsfonts}
\usepackage{nicefrac}
\usepackage{microtype}
\usepackage{xcolor}

\usepackage{url}
\usepackage{amsmath}
\usepackage{amssymb}
\usepackage{mathtools}
\usepackage{amsthm}
\usepackage{multirow,tabularx}
\usepackage{longtable}
\usepackage{booktabs,longtable}
\usepackage{hhline}
\usepackage{microtype}
\usepackage{graphicx}
\usepackage{booktabs}
\usepackage{algorithmic}
\usepackage{algorithm}
\usepackage{pifont}
\usepackage{enumitem}
\usepackage{tcolorbox}
\usepackage{wrapfig}
\usepackage{amsmath}
\usepackage{bm}
\usepackage{fontawesome}
\usepackage{cancel}
\usepackage[normalem]{ulem}

\newcommand{\Def}[0]{\mathrel{\mathop:}=}

\usepackage{pifont}

\usepackage{color, colortbl}
\definecolor{Gray}{gray}{0.93}
\definecolor{Orange}{rgb}{1,0.5,0}
\definecolor{Green}{rgb}{0,0.80,0}
\definecolor{Blue}{rgb}{0,0,0.92}
\definecolor{Red}{rgb}{0.90,0,0}
\definecolor{DGray}{gray}{0.83}
\definecolor{LightCyan}{rgb}{0.88,1,1}

\definecolor{bluegray}{rgb}{0.4, 0.6, 0.8}
\definecolor{ceruleanblue}{rgb}{0.16, 0.32, 0.75}

\newcommand{\Retrain}{{\text{Retrain}}}

\newcommand{\thetau}{{\btheta_\mathrm{u}}}
\newcommand{\thetao}{{\btheta_\mathrm{o}}}

\usepackage{amsmath,amsfonts,bm}

\def\eqref#1{(\ref{#1})}
\def\1{\bm{1}}

\newcommand{\Dr}{\mathcal{D_{\mathrm{r}}}}
\newcommand{\Df}{\mathcal{D_{\mathrm{f}}}}
\newcommand{\Dt}{\mathcal{D_{\mathrm{t}}}}

\DeclareMathAlphabet{\mathsfit}{\encodingdefault}{\sfdefault}{m}{sl}
\SetMathAlphabet{\mathsfit}{bold}{\encodingdefault}{\sfdefault}{bx}{n}

\DeclareMathOperator*{\minimize}{\text{minimize}}

\newcommand{\btheta}{{\boldsymbol{\theta}}}

\newcommand{\bdelta}{\boldsymbol\delta}

\usepackage[font={footnotesize}]{caption}
\usepackage{wrapfig}
\usepackage{siunitx}
\usepackage{makecell}
\DeclareCaptionFont{red}{\color{red}}   % new floats use \captionsetup{font+=red} (caption resets the color)
\definecolor{fvCifar}{RGB}{0,90,181}    % blue:   CIFAR-10 DDPM paragraph + upper half of Tab. gen_class
\definecolor{fvDit}{RGB}{0,128,96}      % green:  ImageNet DiT paragraph + lower half of Tab. gen_class
\definecolor{fvSd}{RGB}{204,102,0}      % orange: nudity-removal paragraph + I2P/Sens./CLIP/FID columns of Tab. gen_sd
\definecolor{fvAtk}{RGB}{120,60,160}    % purple: adversarial-prompt paragraph + Pre/ASR columns of Tab. gen_sd

\definecolor{cvprblue}{rgb}{0.21,0.49,0.74}
\usepackage[pagebackref,breaklinks,colorlinks,allcolors=cvprblue]{hyperref}
\def\confName{CVPR}
\def\confYear{2027}

\title{
Forget Vectors at Play: Universal Input Perturbations
Driving Machine Unlearning in Image Classification and Generation}

\author{Changchang Sun$^{1}\thanks{Work conducted during the visit to the OPTML Group@Michigan State University.}$ ~~~~Ren Wang$^{2}$ ~~~~Yihua Zhang$^3$  ~~~~Jinghan Jia$^3$ \\ ~~~~Jiancheng Liu$^3$ ~~~~Gaowen Liu$^4$ ~~~~Yan Yan$^1$  ~~~~Sijia Liu $^3$ \\
  $^1$University of Illinois Chicago
  ~~~~~ $^2$Illinois Institute of Technology\\
 $^3$Michigan State University	
  ~~~~~ $^4$Cisco Research
  \\
}

\newcommand{\scc}[1]{\textcolor{red}{#1}}

\definecolor{mg}{RGB}{255, 220, 220}
\definecolor{mr}{RGB}{255, 240, 200}

\definecolor{star}{RGB}{0, 147, 0}

\definecolor{mb}{RGB}{255, 160, 130}
\definecolor{mpink}{RGB}{200, 170, 200}
\definecolor{mpurple}{RGB}{160, 195, 160}

\definecolor{mgap}{RGB}{130, 50, 70}

\newcommand{\cb}[2]{
  \colorbox{#1}{\parbox[c][0.16cm][c]{2.5cm}{\centering #2}}
}

\newcommand{\cbs}[2]{%
  \colorbox{#1}{\parbox[c][0.16cm][c]{0.9cm}{\centering #2}}%
}

\newcommand{\cbsr}[2]{%
  \colorbox{#1}{\parbox[c][0.009cm][c]{0.7cm}{\centering #2}}%
}
\newcommand{\cbsg}[2]{%
  \colorbox{#1}{\parbox[c][0.009cm][c]{0.5cm}{\centering #2}}%
}
\begin{document}
\maketitle
\begin{abstract}
Machine unlearning (MU), which seeks to erase the influence of specific unwanted data from already-trained models, is becoming increasingly vital in model editing, particularly to comply with evolving data regulations like the ``right to be forgotten''. Conventional MU approaches are predominantly model-based, typically requiring retraining or fine-tuning the model's weights to meet unlearning requirements. In this work, we approach the MU problem from an input perturbation-based perspective, where the model weights remain intact throughout the unlearning process. We demonstrate the existence of a proactive input-based unlearning strategy, referred to \textit{forget vector}, which can be generated as an input-agnostic data perturbation and remains as effective as model-based approximate unlearning approaches in both image classification and image generation. 
For \textit{image classification}, we demonstrate that forget vector can induce effective class-wise and
random data forgetting in frozen classifiers while preserving performance on retained data. 
We also explore \textit{forget vector arithmetic}, whereby multiple class-specific forget vectors can be combined through simple operations (\textit{e.g.}, linear combinations) to generate new forget vectors for unseen unlearning tasks, such as forgetting arbitrary subsets across classes. 
For \textit{image generation}, we extend forget vector to conditional diffusion models by jointly perturbing the noisy input and conditioning embeddings of frozen denoisers, enabling targeted removal of unwanted classes and concepts from generated outputs.
Extensive experiments on both image classification and image generation validate the effectiveness and adaptability of the forget vector, showcasing its competitive performance relative to state-of-the-art model-based methods while achieving superior parameter efficiency.
\end{abstract}

\section{Introduction}
To prevent unauthorized use of personal or sensitive data after training and comply with legislation such as the ``right to be forgotten'' \citep{hoofnagle2019european}, machine unlearning (\textbf{MU}) has garnered increasing attention as a solution to various challenges in vision tasks \citep{golatkar2020eternal,poppi2023multi,warnecke2021machine,fan2023salun}.
In essence, it initiates a reverse learning process to erase the impact of unwanted data (\textit{e.g.}, specific data points, classes, or knowledge) from an already-trained model, while still preserving its utility for information not targeted by an unlearning request.
Based on the guarantees provided for data removal from already-trained models, existing MU methods can be broadly categorized into two approaches: \textit{exact unlearning} \citep{dong2024idea,guo2019certified,thudi2022necessity} and \textit{approximate unlearning} \citep{graves2021amnesiac,thudi2022unrolling,becker2022evaluating,izzo2021approximate,cha2024learning, chen2023boundary, tarun2023fast}. The former guarantees the complete and verifiable removal of targeted data, typically achieved by retraining the model from scratch with the data to be forgotten excluded from the training set, a process we refer to as \textit{\Retrain}.
However, due to the high computational overhead, research has increasingly focused on approximate unlearning methods, which seek to achieve efficient unlearning without requiring full retraining.

Approximate unlearning strikes a balance between computational efficiency and effective data removal, making it practical for many real-world applications. Most existing approximate unlearning techniques are \textit{model-based}, updating the model's weights within a limited number of training iterations to eliminate the influence of specific unwanted data, thus avoiding a full retraining process. Representative methods in this category include fine-tuning approaches~\citep{warnecke2021machine,perifanis2024sftc}, gradient ascent techniques~\citep{thudi2022unrolling,chen2024machine}, and influence function-based methods~\citep{golatkar2021mixed,golatkar2020eternal}.

Although the model-based unlearning methods have made significant strides, they often overlook the \textit{data-based} dimension and its potential impact on MU. For instance, it remains unclear whether current MU approaches generalize effectively to ``shifted'' forget data. 
Additionally, the possibility of a data-based MU design that operates without updating model parameters has yet to be explored. Therefore, we ask:
\vspace{0.01cm}
\begin{tcolorbox}[before skip=-2.1mm, after skip=0.2cm, boxsep=0.0cm, middle=0.1cm, top=0.1cm, bottom=0.1cm]
\textbf{(Q)} \textit{
Can we explore data influence in MU and harness data-based operations to fulfill MU?
}
\end{tcolorbox}
\vspace{-0.02in}

 \begin{figure*}[t]
	\centering
\includegraphics[scale=0.5]{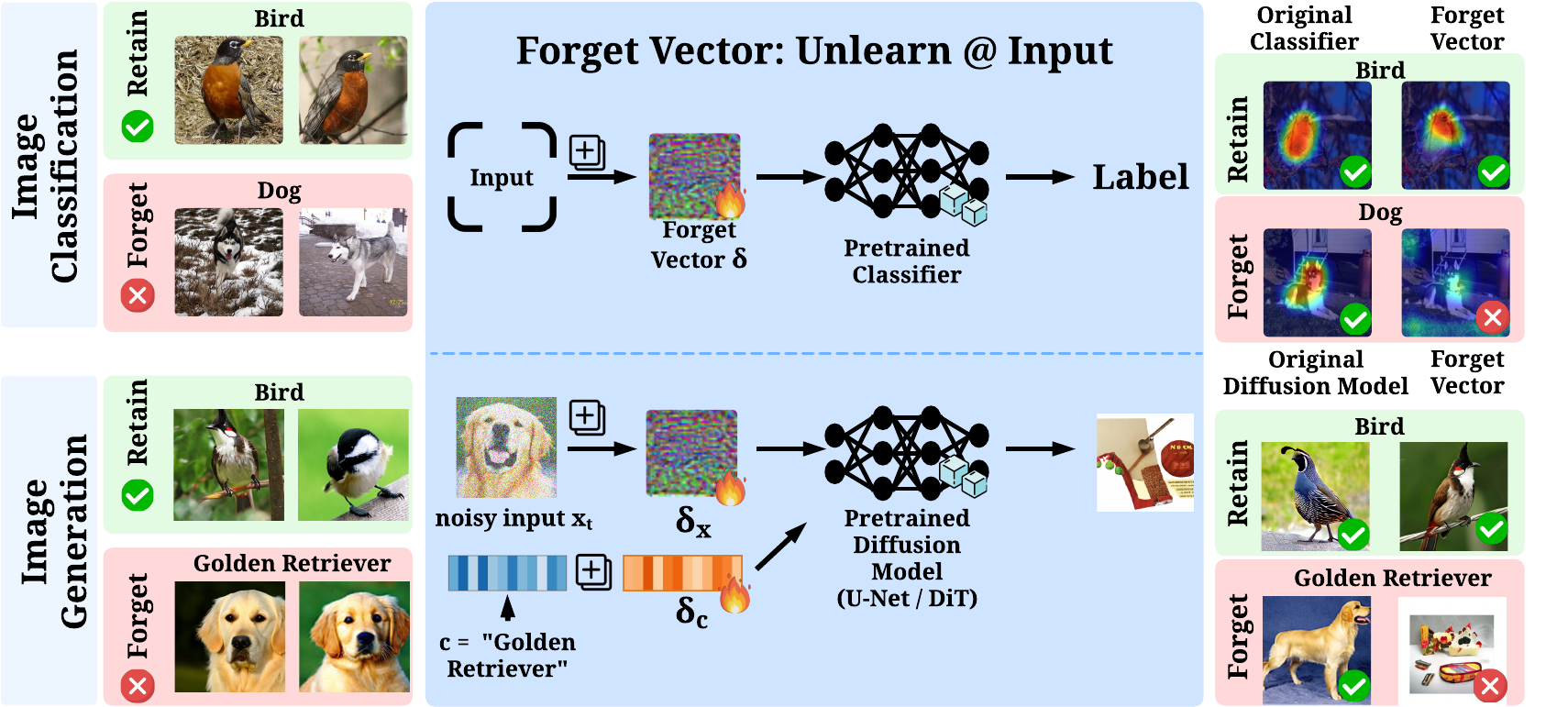}
\vspace{-0.1in}
\caption{
Overview of input-space machine unlearning with forget vectors
for image classification and generation.
Our approach achieves selective forgetting by applying
learnable perturbations directly to model inputs while
keeping pretrained model parameters frozen.
For image classification (top), a universal forget vector
$\delta$ suppresses recognition of the target class
(e.g., ``dog'') while preserving predictions for retained
classes (e.g., ``bird'').
For image generation (bottom), forget vectors
$\delta_x$ and $\delta_c$ are applied to noisy inputs
and conditioning representations, respectively,
to suppress the generation of the target concept
(e.g., ``Golden Retriever'') while retaining
the ability to generate unrelated concepts
(e.g., ``bird'').
Green check {\color{green}{\faCheckCircle}} marks indicate preserved recognition
or generation behavior, whereas red crosses {\color{red}{\faTimesCircle}}
indicate successful forgetting.
}
\label{framework}
\vspace{-0.2in}
\end{figure*}

To address \textbf{(Q)}, we study MU from a fresh data-based viewpoint:  \textbf{forget vector}, a universal input data perturbation designed to promote unlearning effectively; See the schematic overview in \textbf{Fig.\,\ref{framework}}.
\iffalse
\scc{Before developing the forget vector, we explore the rationale for how data perturbations complement current model-based MU approaches, as evidenced by these methods’ generalization to common data shifts, including Gaussian noise and adversarial perturbations \cite{hendrycks2018benchmarking,goodfellow2014explaining}.}
\fi
To design the forget vector, we draw inspiration from recent input prompting techniques for vision models, known as visual prompting \cite{chen2023understanding,oh2023blackvip} or model reprogramming \cite{elsayed2018adversarial,zhang2022fairness,chen2024model}, used in transfer learning and model adaptation. These prompting methods learn input perturbations to enable a fixed model to perform well on new tasks, effectively guiding the model to execute tasks it wasn't originally trained for.
From this perspective, our research on the forget vector also explores whether it is possible to append a trainable ``prompt'' to the input to guide an already-trained neural network in unlearning specific data.
The proposed forget vector allows the unlearner to modify user inputs targeted for deletion, offering a flexible and efficient approach to unlearning while potentially achieving significant parameter efficiency.
Specifically, we instantiate forget vectors for two distinct tasks. For image classification, we introduce an input-agnostic perturbation that is applied to input images of a frozen classifier, inducing targeted forgetting while preserving predictions on retained data. For image generation, we extend this principle to conditional diffusion models, where forget vectors jointly perturb the noisy input and conditioning embedding of a frozen denoiser to suppress unwanted classes or concepts in generated outputs. Both approaches achieve targeted behavioral forgetting without modifying the pretrained model parameters.
We summarize our contributions below:
\begin{itemize}
    \item We introduce \textit{forget vectors}, an input-space machine unlearning framework that enables targeted forgetting in both image classification and image generation without modifying pretrained model parameters.

    \item For image classification, we develop input-agnostic forget vectors that enable class-wise and random data forgetting in frozen classifiers. 
    We further demonstrate the effectiveness of \textit{forget vector arithmetic} by using {precomputed} \textit{class-wise} forget vectors to generate new vectors that effectively eliminate the influence of specific data subsets in image classification models, \textit{e.g.}, in the scenario of \textit{random} data forgetting.

    \item For image generation, we extend forget vectors to conditional diffusion models by jointly perturbing noisy inputs and conditioning embeddings of frozen denoisers, enabling targeted suppression of unwanted classes and concepts in generated outputs.

    \item We conduct extensive experiments across image classification and generation,
    providing both quantitative and qualitative analyses to demonstrate the competitiveness of the forget vector compared to model-based MU methods.
\end{itemize}

% \vspace*{-3mm}
\section{Related Work}
\noindent\textbf{MU in Vision.}
Machine unlearning (MU) in vision has gained significant attention due to the increasing need for privacy preservation, copyright protection, and ethical data removal in machine learning models. 
Recent studies~\citep{wang2024machine,pan2022machine,li2024machine,foster2024loss,lin2023erm,gupta2021adaptive,kurmanji2024towards,di2022hidden,chen2024fast,zhang2024defensive,liu2024model} in this area have primarily focused on two main applications: image classification and image generation.

In \textbf{image classification}, MU methods have explored various ways to 
erase certain classes or images from models~\citep{thudi2022unrolling,chen2024machine,golatkar2020eternal,golatkar2021mixed,liu2024model,pochinkov2024dissecting}.
Specifically,
fine-tuning-based methods update the model incrementally on a modified dataset without the unwanted data points~\citep{warnecke2021machine, perifanis2024sftc}. 
Gradient ascent-based approaches attempt to reverse the impact of unwanted data by applying gradient ascent to model parameters~\citep{thudi2022unrolling, chen2024machine}. Influence-based unlearning techniques 
estimate and negate the effect of specific data points on model predictions and parameters
to achieve unlearning~\citep{golatkar2020eternal, golatkar2021mixed}. Another line of research explores the relationship between MU and model pruning, suggesting that model sparsity 
can help to bridge the gap 
between approximate and exact unlearning, 
reducing the need for complex parameter updates~\citep{liu2024model, pochinkov2024dissecting}.

For \textbf{image generation}, existing MU methods aim to suppress unwanted or harmful content while preserving the quality of retained generations~\citep{fan2023salun,li2024machine,zhang2024defensive}.
Most approaches achieve concept erasure by modifying pretrained model parameters, including selectively updating salient weights~\citep{fan2023salun}, fine-tuning denoisers with negatively guided targets~\citep{gandikota2023erasing}, applying closed-form edits to cross-attention projections~\citep{gandikota2024unified,gong2024reliable}, incorporating geometry-aware regularization to preserve retained knowledge~\citep{huang2024unified}, and adversarially training text encoders to enhance erasure robustness~\citep{zhang2024defensive}.
In contrast, training-free approaches regulate the generation process at inference time through negative prompting, safety guidance~\citep{schramowski2023safe}, or prompt-embedding projections~\citep{yoon2025safree}.
Meanwhile, adversarial prompt attacks have been developed to evaluate the robustness of concept erasure methods~\citep{zhang2023generate}.

\noindent\textbf{Input-based Model Adaptation.}
This approach aims to
modify or repurpose pre-trained models for new tasks or specific objectives without the need for full retraining. It is particularly beneficial for reducing computational costs and leveraging existing knowledge within models.
Key techniques in model adaptation include:
Visual prompting~\citep{jia2022visual,wang2023transhp,liu2023explicit,hossain2024visual,zhang2024exploring}
maintains the pre-trained model’s parameters fixed and adapts the input to enable the model to perform different tasks. 
For example, introducing trainable parameters in the input space while keeping the model backbone frozen can achieve comparable results with reduced computational overhead. 
Model reprogramming~\citep{tsai2020transfer,elsayed2018adversarial,yang2021voice2series,melnyk2023reprogramming} involves keeping a pre-trained model unchanged while modifying its inputs to adapt the model for new tasks. For example, adversarial perturbations can be applied to inputs at test time, allowing the model to perform a specific task dictated by the perturbations, even if that task was not originally intended for the model.
Feature-based domain adaptation \citep{wang2018visual,tahmoresnezhad2017visual} applies transformations or mapping techniques to the input data, 
aligning the feature
distributions
between
the source and target domains while keeping the model unchanged.
In text-to-image diffusion, textual inversion \citep{gal2022image} likewise learns a vector in the condition-embedding space of a frozen generator, but to add a concept rather than to remove one.

% \vspace*{-2mm}
\section{Preliminaries on MU and Problem Statement}
\label{sec: Preliminaries}

\noindent\textbf{Formulation of MU.} 
We first formulate MU for \textbf{image classification} and then extend the formulation to \textbf{conditional generative models}. Let $\mathcal{D} = \{ \mathbf{x}_i, y_i \}_{i=1}^N$ represent a training set with $N$ examples, where $\mathbf{x}_i$ denotes the $i$th image data, and $y_i$ denotes its corresponding class label.
Following the classic MU setting \cite{golatkar2020eternal,kurmanji2024machine,kurmanji2024towards,fan2023salun},
we introduce a \textit{forget set} $\Df\subseteq \mathcal{D}$,  which specifies the training samples targeted for unlearning.
Accordingly, the complement of $\Df$ is the \textit{retain set}, \textit{i.e.}, $\Dr = \mathcal{D} \setminus \Df$.
The \textit{goal of MU} is to efficiently and effectively eliminate the influence of $\Df$ on an already-trained model 
$\thetao$, so that the performance of the post-unlearning model closely approximates that of a model retrained from scratch on $\Dr$ (\textit{i.e.}, excluding the impact of $\Df$ from scratch). Therefore, such a retraining method (referred to as \textit{\Retrain}) is typically considered as the gold standard of MU \cite{thudi2022unrolling,jia2023model}. However, since {\Retrain} is computationally intensive, most popular MU approaches instead address an unlearning optimization problem using the forget and retain sets to update the model parameters 
${\btheta}$, starting from the originally pre-trained model $\thetao$.
This yields the following optimization problem for MU:
\begin{align}
    \begin{array}{ll}
\displaystyle \minimize_{\btheta}         &  \ell_{\mathrm{MU}} (\btheta; \Df, \Dr),
    \end{array}
    \label{eq: MU_prob_basic}
\end{align}
with the initialization $\btheta = \thetao$. 
In \eqref{eq: MU_prob_basic}, $\ell_{\mathrm{MU}}$ represents an appropriate unlearning loss function that may depend on $\Df$ and/or $\Dr$, as will be detailed when introducing specific unlearning methods.

\subsection{MU for Image Classification}
In the context of MU for image classification \cite{golatkar2020eternal,fan2023salun}, the specification of the forget set $\Df$ leads to two unlearning scenarios: \textit{class-wise forgetting}, where $\Df$ consists of a subset focused on a specific image class targeted for unlearning, and \textit{random data forgetting}, where $\mathcal{D}_\mathrm{f}$ is a randomly selected subset of images across all classes.

\noindent\textbf{Model-based MU Methods and Evaluation.} 
The formulation in \eqref{eq: MU_prob_basic} represents the predominant  MU solution in the literature, focusing on modifying model weights and/or architectural components to achieve the unlearning objective. 

In what follows, we introduce several representative MU approaches that serve as approximations to {\Retrain}. 
\iffalse 
$\bullet$ Retrain: This is the most optimal (exact) MU method, where the model is retrained from scratch using the retain set $\mathcal{D}_\mathrm{r}$. However, this approach imposes a significant computational cost, especially when training deep neural networks (DNNs) on large-scale datasets. Despite its inefficiency, Retrain serves as the benchmark for MU, representing the ideal result that other MU methods aim to achieve.
\fi 
(a) Fine-tuning \textbf{(FT)} \cite{warnecke2021machine}: This approach treats the MU problem as a continual learning task, defining the unlearning objective $\ell_\mathrm{MU}$ as a training objective that fine-tunes 
$\thetao$
over $\mathcal{D}_\mathrm{r}$ to induce catastrophic forgetting of $\mathcal{D}_\mathrm{f}$.
(b) Random labeling \textbf{(RL)} \cite{golatkar2020eternal}:  This approach specifies the unlearning objective $\ell_\mathrm{MU}$ by assigning random labels or features to the data in $\mathcal{D}_\mathrm{f}$, thereby enforcing model forgetting.
(c) Gradient ascent \textbf{(GA)} \cite{thudi2022unrolling}: This approach employs the negative of the FT loss to reverse the training impact associated with the data in $\mathcal{D}_\mathrm{f}$.
(d) \textbf{Localization}-informed unlearning \cite{jia2023model,fan2023salun}: This method identifies a subset of model weights critical to the unlearning task (\textit{e.g.}, through model sparsity \cite{jia2023model} or gradient saliency \cite{fan2023salun}) and incorporates this weight localization as a prior to solve the unlearning problem in \eqref{eq: MU_prob_basic}.

Given an unlearned model (denoted as $\thetau$) after solving \eqref{eq: MU_prob_basic}, unlearning performance is evaluated in two main areas: \textit{unlearning effectiveness}, which measures whether the target data/information has been successfully removed, and \textit{utility retention}, which assesses whether unlearning has preserved the model's classification ability on unaffected data.
Following the evaluation pipeline in \cite{jia2023model}, unlearning effectiveness is quantified by two metrics: unlearning accuracy (\textbf{UA}), defined as 
 $1-$the model's accuracy on $\Df$ (higher UA indicates better unlearning), and membership inference attack performance on $\Df$, termed \textbf{MIA-Efficacy}, where higher prediction accuracy on non-training samples indicates better unlearning.
 Utility retention is measured by retain accuracy (\textbf{RA}), reflecting the model's accuracy on $\Dr$, and testing accuracy (\textbf{TA}), which is the accuracy on the original test set. 
 Notably, {TA} is assessed on the entire original test set, except in the case of class-wise forgetting, where test samples from the forgotten class are excluded from evaluation.

% \vspace*{-1mm}
\subsection{MU for Conditional Image Generation}
\noindent\textbf{MU for Conditional Diffusion Models.}
A conditional diffusion model is trained on pairs $(\mathbf{x}, c)$, where the condition $c$ is a class label (\textit{e.g.}, class-conditional denoising diffusion probabilistic models (DDPM)~\citep{ho2020denoising} and Diffusion Transformers (DiT)~\citep{peebles2023scalable}) or a text prompt (\textit{e.g.}, Stable Diffusion (SD)~\citep{rombach2022high}); its denoiser $\boldsymbol{\epsilon}_{\btheta}(\mathbf{x}_t, t, c)$ predicts the noise added to $\mathbf{x}$ at diffusion step $t$, and images are sampled with classifier-free guidance \citep{ho2022classifier}. Here $\Df$ holds the data of a class or a concept to be forgotten, and unlearning should stop the model from generating it while preserving the generation of everything else. Following \citep{fan2023salun,huang2024unified}, we consider class-wise forgetting in class-conditional models and concept (nudity) forgetting in text-to-image models. Unlearning effectiveness is measured by the forgetting accuracy (\textbf{FA}), \textit{i.e.}, the fraction of images generated under the forgotten class that a pretrained classifier still assigns to that class (lower is better), and, for nudity, by the number of exposed body parts that NudeNet \citep{bedapudi2019nudenet} detects in images generated from inappropriate prompts. Utility is measured by Fr\'echet Inception Distance (FID)~\citep{heusel2017gans} and classification accuracy (RA) of generated retained-class images, or by CLIP score \citep{radford2021learning} and FID on benign prompts. Since retraining a large generator from scratch is infeasible, the pretrained model and existing unlearning methods serve as references instead of {\Retrain}.

\subsection{Data-based MU Design}
\noindent\textbf{The Forget Vector Problem.}
While previous MU methods can be unified within the framework of \eqref{eq: MU_prob_basic} by varying the unlearning loss $\ell_{\mathrm{MU}}$ and weight localization priors, recent advancements in input data-based model adaptation, such as visual prompting \cite{bahng2022exploring,chen2023understanding} and model reprogramming \cite{chen2024model,elsayed2018adversarial}, suggest an alternative approach to MU. This strategy inspires us to design data-based prompting (implemented through universal input perturbations) to achieve unlearning without modifying the model itself.
We refer to this input perturbation vector, designed specifically for MU, as the \textbf{forget vector}.
To be more specific, let $\bdelta$ represent the data-agnostic input perturbations to be designed. The problem of constructing a forget vector for MU can be formulated as
\begin{align}
    \begin{array}{cc}
 \displaystyle \minimize_{\bdelta}        &  \ell_{\mathrm{MU}} (\bdelta; \thetao, \Df, \Dr) , 
    \end{array}
    \label{eq: forget_vector_problem}
\end{align}
 where $\bdelta$ is the perturbation variable, applied linearly to the forget and retain samples as $\mathbf{x}^\prime \Def \mathbf{x} + \bdelta$ for $\mathbf{x} \in \Df $ and $\Dr$,
 for image classifiers. For a conditional diffusion model, the same idea applies to both inputs of the denoiser: the noisy sample and the condition embedding.
In practice, since the model remains unchanged, the unlearner can compute the forget vector based on the forget request (forget set) and append it to model inputs to process user-initiated unlearning requests. 
In this work, we do not consider counter-unlearning adversaries that intentionally negate the effect of the forget vector.

\iffalse
Based on \eqref{eq: forget_vector_problem}, we are motivated to explore two research questions:
\textbf{(Q1)} How do ``perturbations'' applied to forget data affect unlearning performance?
\textbf{(Q2)} How can we effectively design the forget vector $\boldsymbol{\delta}$ to solve problem \eqref{eq: forget_vector_problem}?
These two questions are interconnected: the answer to (Q1) offers a sensitivity analysis of MU to data shifts within the forget set, guiding how the specific shift induced by the forget vector can be optimized for effective unlearning in (Q2).
Therefore, the following Secs.\,\ref{sec: method1}-\ref{sec: method2} address (Q1) and (Q2) in sequence. For (Q1), the next section analyzes performance through an evaluation lens on a given unlearned model, using data perturbations applied via standard data augmentation operations or adversarial perturbations.
\fi

% \vspace*{-2mm}
\section{Optimization for Forget Vector}
\label{sec: method2}
% \vspace*{-1mm}

In this section, we formulate task-specific unlearning objectives,
$\ell_{\mathrm{MU}}$, under the unified forget vector framework
in \eqref{eq: forget_vector_problem}, for both image
classification and generation.

\subsection{Forget Vector for Image Classification}
\noindent\textbf{Unlearning Objective Design of Forget Vectors.}
Our design aims for the forget vector variable ($\boldsymbol{\delta}$), when applied to the forget set ($\Df$), to drive the given model's predictions (${\thetao}$) away from the correct labels. Conversely, when applied to the retain set ($\Dr$),
the forget vector should minimally affect correct predictions.
The first forget objective aligns with adversarial attack design, aiming to mislead the model's predictions in the presence of the perturbation $\boldsymbol{\delta}$. The second retain objective acts as a utility regularization, suppressing the unlearning effect of the perturbation when applied to data samples not targeted for unlearning (retain samples).

To implement the forget objective (denoted by $\ell_\mathrm{f}$), we draw inspiration from the C\&W untargeted attack loss \citep{carlini2017towards}. This is given by a margin loss, designed to remain actively minimizing when the top prediction matches the correct label, ensuring that optimization continues until predictions are shifted to an incorrect label, thereby achieving unlearning. This can be cast as  
\vspace*{-2mm}
\begin{equation}
\begin{aligned}
\ell_{\mathrm{f}}(\bdelta;\thetao,\Df)
&= \mathbb{E}_{(\mathbf{x},y)\in\Df}
\max\Big\{
f_{\thetao,y}(\mathbf{x}+\bdelta) \\
&\quad -
\max_{k\neq y} f_{\thetao,k}(\mathbf{x}+\bdelta),
-\tau
\Big\},
\end{aligned}
\label{eq: forget_set_loss}
\end{equation}
where $(\mathbf{x}, y)\in \Df$ denotes a forget sample with $y$ being the prediction label of $\mathbf x$, $\mathbf x + \bdelta$ is  the perturbed sample, $f_{k, \thetao} (\mathbf x)$ denotes the prediction logit (before softmax) of the model $\thetao$ for class $k$ under the input $\mathbf{x}$, and $\tau \geq 0$ is a margin threshold that controls the unlearning strength.
The rationale behind \eqref{eq: forget_set_loss} is that minimizing it ensures convergence to the negative margin $f_{ {y}}(\mathbf{x}+\bdelta)-\max_{ {k}\neq
 {y}}f_ {k}(\mathbf{x}+\bdelta) \to  -\tau \leq 0$.  Thus, the forget vector $\boldsymbol{\delta}$ enforces unlearning on $\boldsymbol{\theta}$ for $\mathbf{x}$ by making the incorrect prediction ($k \neq y$) have a higher confidence than the original correct prediction $y$. On the other hand, once the margin becomes negative (indicating that the prediction label has been flipped), the forget objective $\ell_\mathrm{f}$ automatically terminates, allowing a balance with the retain objective, which will be introduced later.
In our experiments, we find that the forget objective is robust to variations in the \textit{nonnegative} margin parameter $\tau$. 
A larger $\tau$ value imposes a stricter unlearning requirement by increasing the logit distance from the correct label. For example, we set $\tau = 1$ in our experiments.

Next, we regularize the forget objective \eqref{eq: forget_set_loss} with the retain objective, defined as the cross-entropy loss ($\ell_\mathrm{CE}$) over the retain set $\mathcal{D}_\mathrm{r}$, along with the $\ell_2$ norm of $\boldsymbol{\delta}$ to ensure minimal perturbation required to achieve both the forget and retain objectives.
This yields the full unlearning objective  in \eqref{eq: forget_vector_problem}:
\vspace*{-2mm}

\begin{equation}
\begin{aligned}
\ell_{\mathrm{MU}}(\bdelta;\thetao,\Df,\Dr)
&= \ell_{\mathrm{f}}(\bdelta;\thetao,\Df) \\
&\quad + \lambda_1 \ell_{\mathrm{CE}}(\bdelta;\thetao,\Dr)
+ \lambda_2 \|\bdelta\|_2^2,
\end{aligned}
\label{eq: unlearn_forget_vector_objective}
\end{equation}
where $\lambda_1 >0$ and $\lambda_2 > 0$ are the regularization parameters, and $\ell_{\mathrm{CE}}(\bdelta;\thetao, \Dr)$ denotes the CE loss of the model $\thetao$ over the perturbed retain set $\{ (\mathbf x + \bdelta, y) \}_{(\mathbf x, y) \in \Dr}$.
Integrating \eqref{eq: unlearn_forget_vector_objective} into  \eqref{eq: forget_set_loss}, we can then apply stochastic gradient descent (SGD) \cite{amari1993backpropagation} to optimize the forget vector variable $\bdelta$.

\noindent\textbf{Compositional Unlearning via Forget Vector Arithmetic.}
A forget vector defines an unlearning direction in the input space to guide the unlearning process. We explore whether a \textit{new} unlearning direction can be efficiently constructed by interpolating from existing precomputed forget vectors, such as class-wise forget vectors obtained by solving \eqref{eq: unlearn_forget_vector_objective} with $\mathcal{D}_\mathrm{f}$ defined as each class’s training set. This approach is analogous to the concept of task vectors in weight space for model editing \cite{ilharco2022editing}. However, to the best of our knowledge, \textit{input-based} task vector arithmetic has not yet been explored in the literature.
If forget vectors can be modified and combined using arithmetic operations, such as negation and addition, we can dynamically adjust a model’s unlearning behavior without re-solving the optimization problem \eqref{eq: unlearn_forget_vector_objective} or any other model-based unlearning problem in \eqref{eq: MU_prob_basic}. We refer to this new unlearning paradigm as \textit{compositional unlearning}, where precomputed class-wise forget vectors can be efficiently combined to generate a new forget vector for each deletion request in the context of random data forgetting.

Let $\boldsymbol{\delta}_k$ denote the forget vector used for unlearning data points of class $k$. Given the set of forget vectors $\{ \boldsymbol{\delta}_k \}_{k=1}^K$ for all $K$ classes, we obtain these vectors by solving \eqref{eq: unlearn_forget_vector_objective} with $\mathcal{D}_\mathrm{f}$ defined as each class’s training set, respectively. The forget vector for compositional unlearning is given by
\vspace*{-2mm}
\begin{align}
    \bdelta(\mathbf w) \Def \sum_{k=1}^K (w_k \boldsymbol{\delta}_k),
    \label{eq: compositional_unlearn}
\end{align}
where $\mathbf w = [w_1, \ldots w_K]^K$ are the linear combination coefficients to be optimized, which determine the forget vector arithmetic.
To determine $\mathbf{w}$, we can minimize \eqref{eq: unlearn_forget_vector_objective} with the optimization restricted to the coefficients $\mathbf{w}$. Instead of penalizing the $\ell_2$ norm of the forget vector, we penalize the $\ell_2$ norm of $\mathbf{w}$ to prevent excessive pixel perturbation. This modifies \eqref{eq: unlearn_forget_vector_objective}  to  the problem $\min_{\mathbf w}\, \ell_{\mathrm{f}}(\bdelta(\mathbf w);\thetao, \Df)  + \lambda_1  \ell_{\mathrm{CE}}(\bdelta(\mathbf w);\thetao, \Dr) + \lambda_2 \|\mathbf w\|_2^2$. 
As will be shown later, random data forgetting can be achieved through class-wise forget vector arithmetic \eqref{eq: compositional_unlearn} by applying the compositional scheme defined by the coefficients $\mathbf{w}$.

\subsection{Forget Vector for Image Generation}
\label{sec: fv_gen}
\noindent\textbf{Applying the Forget Vector to a Denoiser.}
A conditional diffusion model generates an image by repeatedly calling its denoiser $\boldsymbol{\epsilon}_{\thetao}(\mathbf{x}_t, t, c)$, which predicts the noise $\boldsymbol{\epsilon}$ in the noisy sample $\mathbf{x}_t = \sqrt{\bar\alpha_t}\,\mathbf{x} + \sqrt{1-\bar\alpha_t}\,\boldsymbol{\epsilon}$ at step $t$ under the condition $c$ (for DiT and SD, $\mathbf{x}$ is the latent code of the image given by a pretrained variational autoencoder).
Unlike a classifier, the denoiser has two inputs: the noisy sample $\mathbf{x}_t$ and the condition, which enters the network through an embedding $e(c)$, \textit{i.e.}, a learned class embedding for DDPM and DiT, or the hidden states of the CLIP text encoder for SD.
We therefore give the forget vector one part per input, $\bdelta = (\bdelta_x, \bdelta_c)$, and define the perturbed denoiser as
\begin{align}
\boldsymbol{\epsilon}_{\bdelta}(\mathbf{x}_t, t, c) \Def \boldsymbol{\epsilon}_{\thetao}\big(\mathbf{x}_t + \bdelta_x,\ t,\ e(c) + \bdelta_c\big),
\label{eq: fv_diffusion}
\end{align}
where $\bdelta_x$ has the shape of $\mathbf{x}_t$ and $\bdelta_c$ that of $e(c)$.
As in classification, $\bdelta$ is universal: the same $\bdelta$ is used at every denoising step and for every condition, forgotten or not, so no request has to be identified as targeting the forgotten content at deployment.
$\bdelta_x$ changes only what the denoiser sees; the sampling trajectory $\{\mathbf{x}_t\}$ itself is not perturbed.
Under classifier-free guidance, $\bdelta_x$ enters both the conditional and the unconditional pass, as both take $\mathbf{x}_t$ as input, whereas $\bdelta_c$ is added only to the given condition, not to the null condition $\varnothing$.

\noindent\textbf{Forget Objective.}
The margin loss \eqref{eq: forget_set_loss} relies on class logits, which a denoiser does not output. We instead redirect the denoiser's prediction: on noisy forget samples, the perturbed prediction is pulled toward a target $\tilde{\boldsymbol\epsilon}$ that does not depict the forgotten condition $c_\mathrm{f}$,
\begin{align}
\ell_\mathrm{f}(\bdelta;\thetao,\Df) = \mathbb{E}_{(\mathbf{x},c_\mathrm{f})\in\Df,\,t,\,\boldsymbol{\epsilon}} \big\|\boldsymbol{\epsilon}_{\bdelta}(\mathbf{x}_t, t, c_\mathrm{f}) - \tilde{\boldsymbol\epsilon}\big\|_2^2 .
\label{eq: fv_diffusion_forget}
\end{align}
The target is built from the predictions of the original model at the same $(\mathbf{x}_t, t)$, $\bar{\boldsymbol\epsilon}(c) \Def \boldsymbol{\epsilon}_{\thetao}(\mathbf{x}_t, t, c)$, which do not depend on $\bdelta$.
We consider two choices: an \textit{anchor} target $\tilde{\boldsymbol\epsilon} = \bar{\boldsymbol\epsilon}(c_\mathrm{a})$, where $c_\mathrm{a}$ is a randomly drawn other class or, for nudity, the prompt ``a photo of a person wearing clothes'', which mirrors the random-labeling objective of SalUn \citep{fan2023salun}; and the \textit{negative-guidance} target of ESD \citep{gandikota2023erasing}, $\tilde{\boldsymbol\epsilon} = \bar{\boldsymbol\epsilon}(\varnothing) - \eta\,[\bar{\boldsymbol\epsilon}(c_\mathrm{f}) - \bar{\boldsymbol\epsilon}(\varnothing)]$, which points away from $c_\mathrm{f}$.

\noindent\textbf{Retain Objective and Overall Problem.}
The retain objective replaces the CE loss in \eqref{eq: unlearn_forget_vector_objective}. Rather than fitting the sampled noise $\boldsymbol{\epsilon}$, it keeps the perturbed predictions on retain data equal to those of the original model:
\begin{align}
\ell_\mathrm{r}(\bdelta;\thetao,\Dr) = \mathbb{E}_{(\mathbf{x},c)\in\Dr,\,t,\,\boldsymbol{\epsilon}} \big\|\boldsymbol{\epsilon}_{\bdelta}(\mathbf{x}_t, t, c) - \bar{\boldsymbol\epsilon}(c)\big\|_2^2 .
\label{eq: fv_diffusion_retain}
\end{align}
Since $\bdelta_x$ also enters the unconditional pass of guidance, $10\%$ of the retain queries use the null condition $\varnothing$; for SD, half of the retain queries are COCO \citep{lin2014microsoft} captions, so that general text-to-image generation is preserved as well.
As in \eqref{eq: unlearn_forget_vector_objective}, the forget vector minimizes $\ell_\mathrm{f} + \lambda_1 \ell_\mathrm{r} + \lambda_2 \|\bdelta\|_2^2$ with the denoiser frozen.

\noindent\textbf{Parameter Efficiency.}
The forget vector has only $3{,}200$, $5{,}248$, and $75{,}520$ entries for CIFAR-10 DDPM, DiT-XL/2, and SD v1.4, respectively, $10^4$--$10^5$ times fewer than the $35.7$M, $675$M, and $860$M weights fine-tuned by SFR-on.

\section{Experiments}
\label{sec: exp}

\subsection{Experiment Setups for Image Classification}
\label{sec: exp_setup_cls}
\noindent \textbf{Datasets and Models.}
We focus on MU for image classification, using two datasets: CIFAR-10~\citep{krizhevsky2009learning} and ImageNet-10, a 10-class subset of the original ImageNet \citep{deng2009imagenet}, for ease of implementation of {\Retrain} (exact unlearning) over ImageNet images as \citep{tao2021clustering}. 
For these tasks, we use three image classifiers:  ResNet-18~\citep{he2016deep} for CIFAR-10, VGG-16~\citep{simonyan2014very} and ViT-Base \citep{dosovitskiy2020image} for ImageNet.

\noindent \textbf{Unlearning Baselines and Evaluations.}
For MU in image classification,
we consider two scenarios: \textit{class-wise forgetting} and \textit{random data forgetting}. In class-wise forgetting, training data from an image class are designated for unlearning, while in random data forgetting, a subset of all-class training points is randomly selected as the forget set, with a specified forget ratio of 10\%.
We consider 
$9$ MU baseline methods, including \ding{172} FT \citep{warnecke2021machine}, 
\ding{173} RL \citep{golatkar2020eternal},
\ding{174} GA \citep{thudi2022unrolling},
\ding{175} NegGrad+ \citep{kurmanji2024machine},
\ding{176} SalUn \citep{fan2023salun},
\ding{177} SCRUB \citep{kurmanji2024towards},
\ding{178} Class-F \citep{kodge2024deep}, 
\ding{179} SSD \citep{foster2024fast}, and
\ding{180} UAM \citep{kimunlearning}, where Class-F is only designed for {class-wise forgetting}.

As described in Sec.\,\ref{sec: Preliminaries}, unlearning effectiveness is measured using UA (unlearning accuracy) and MIA-Efficacy, while model utility post-unlearning is assessed by RA (retain accuracy) and TA (testing accuracy); 
For all metrics, being closer to Retrain indicates better performance.
\textbf{It is also worth noting that all existing model-based MU baseline methods \ding{172}-\ding{180} are evaluated on non-perturbed evaluation sets.} However, when using our proposed input-space forget vector solution, we need to apply the forget vector to the evaluation sets (including the forget set, retain set, and testing set) in order to assess unlearning effectiveness and utility retention. This evaluation remains fair, as it aligns with the same objective of forgetting targeted data. The key distinction is that the forget vector operates at the input level, whereas model-based MU baselines achieve unlearning by modifying model weights.
To quantify the performance gap with {\Retrain}, we compare each unlearning baseline and our proposal against this exact unlearning method across all metrics. We report an averaged assessment, termed \textit{Averaging (Avg.) Gap}.
Unless specified otherwise, all the main experiments (whether class-wise or random data forgetting) are conducted over 
10 independent trials.

\noindent \textbf{Implementation Details of Our Proposal.}
To solve the forget vector problem \eqref{eq: forget_vector_problem} with the proposed unlearning objective in \eqref{eq: unlearn_forget_vector_objective}, 
we set the retain loss regularization parameter $\lambda_1$ as follows:
$3$ for CIFAR-10, $5$ for ImageNet-10 with VGG-16, and $7$ for ImageNet-10 with ViT-Base 
in class-wise forgetting. For random data forgetting, we set $\lambda_1$ to $1$. The $\ell_2$-norm regularization parameter is set to $\lambda_2 = 1$. These hyperparameters are determined through a grid search over the range $[0, 10]$.
To optimize \eqref{eq: forget_vector_problem}, we use stochastic gradient descent (SGD) \citep{amari1993backpropagation} with a momentum factor of $0.9$ and an exponential learning rate scheduler, decaying at a rate of $0.9$ per iteration.
Additionally, the batch size is set to $256$, with a maximum of 40 optimization iterations for both two datasets.
To solve the compositional unlearning problem \eqref{eq: compositional_unlearn}, we use a similar setup, setting both $\lambda_1$ and $\lambda_2$ to $1$.

\subsection{Experiment Results  for Image Classification}
\label{sec: exp_results_cls}

\begin{table*}[!t]
\centering
    \caption{
Performance overview of various 
MU methods for image classification under
two unlearning scenarios
on CIFAR-10 using ResNet-18 
and ImageNet-10 using ViT-Base.
Since Class-F \citep{kodge2024deep} is specifically designed for class-wise forgetting, its results do not apply to random data forgetting scenarios (n/a).
Other results are reported in the format $a_{\pm b}$, where $a$ is the mean and $b$ denotes standard deviation $b$ over $10$ independent trials. 
The performance gap against Retrain is indicated 
in (\textcolor{blue}{$\bullet$}), where a lower value is better.
$\uparrow$ (or $\downarrow$) indicates that a higher (or lower) value is better.
The best performance for each metric is highlighted in 
\cbsg{mg}{{\text{pink}}}, while the second-best performance is highlighted in 
\cbsr{mr}{{\text{yellow}}}.
}
\vspace{-0.15cm}
\setlength{\tabcolsep}{0mm}{
\scalebox{0.48}{
\begin{tabular}{
p{2.2cm}<{\centering}|
p{2.8cm}<{\centering}
p{2.8cm}<{\centering}
p{2.8cm}<{\centering}
p{2.8cm}<{\centering}
p{1.5cm}<{\centering}|
p{2.8cm}<{\centering}
p{2.8cm}<{\centering}
p{2.8cm}<{\centering}
p{2.8cm}<{\centering}
p{1.8cm}<{\centering}
} 
\toprule[1pt]
\multicolumn{1}{c|}{\multirow{1}{*}{MU Method}}
&UA$\uparrow$ & MIA-Efficacy$\uparrow$ & RA$\uparrow$ & TA$\uparrow$& \textbf{Avg.Gap}$\downarrow$ 
&UA$\uparrow$ & MIA-Efficacy$\uparrow$ & RA$\uparrow$ & TA$\uparrow$& \textbf{Avg.Gap}$\downarrow$ 
\\
\midrule 
\rowcolor{Gray}
\multicolumn{1}{c}{{ }}
&\multicolumn{5}{c|}{{Class-wise Forgetting, CIFAR-10, ResNet-18}}
&\multicolumn{5}{c}{{Random Data Forgetting, CIFAR-10, ResNet-18}}
\\
\midrule
Retrain	
&$100.00_{\pm{0.00}}$(\textcolor{blue}{$0.00$})
&$100.00_{\pm{0.00}}$(\textcolor{blue}{$0.00$})
&$99.91_{\pm{0.03}}$(\textcolor{blue}{$0.00$})
&$94.92_{\pm{0.15}}$(\textcolor{blue}{$0.00$})
&\textcolor{blue}{$\mathbf{0.00}$} 
&$5.50_{\pm{0.16}}$(\textcolor{blue}{$0.00$})
&$11.57_{\pm{0.47}}$(\textcolor{blue}{$0.00$})
&$99.88_{\pm{0.05}}$(\textcolor{blue}{$0.00$})
&$94.24_{\pm{0.19}}$(\textcolor{blue}{$0.00$})
&\textcolor{blue}{$\mathbf{0.00}$}
\\
\cmidrule(lr){2-11}
FT \citep{warnecke2021machine}
&$5.27_{\pm{0.73}}$(\textcolor{blue}{$94.73$})
&$51.49_{\pm{5.07}}$(\textcolor{blue}{$48.51$})
&$100.0_{\pm{0.00}}$(\textcolor{blue}{$0.09$})
&\cb{mg}{$95.03_{\pm{0.07}}$(\textcolor{blue}{$0.11$})}
&\textcolor{blue}{$\mathbf{35.86}$} 
&$0.03_{\pm{0.03}}$(\textcolor{blue}{$5.47$})
&$0.75_{\pm{0.09}}$(\textcolor{blue}{$10.82$})
&\cb{mr}{$99.98_{\pm{0.02}}$(\textcolor{blue}{$0.10$})}
&\cb{mr}{$94.45_{\pm{0.14}}$(\textcolor{blue}{$0.21$})}
&\textcolor{blue}{$\mathbf{4.15}$} 
\\
RL \citep{golatkar2020eternal}
&$18.87_{\pm{7.34}}$(\textcolor{blue}{$81.13$})
&\cb{mr}{$98.94_{\pm{0.79}}$(\textcolor{blue}{$1.06$})}
&\cb{mr}{$99.98_{\pm{0.00}}$(\textcolor{blue}{$0.07$})}
&$94.51_{\pm{0.12}}$(\textcolor{blue}{$0.41$})
&\textcolor{blue}{$\mathbf{20.67}$} 
&$0.52_{\pm{0.24}}$(\textcolor{blue}{$4.98$})
&$3.13_{\pm{0.55}}$(\textcolor{blue}{$8.44$})
&\cb{mg}{$99.85_{\pm{0.07}}$(\textcolor{blue}{$0.03$})}
&$93.88_{\pm{0.20}}$(\textcolor{blue}{$0.36$})
&\textcolor{blue}{$\mathbf{3.45}$} 
\\
GA \citep{thudi2022unrolling}
&$71.45_{\pm{0.35}}$(\textcolor{blue}{$28.55$})
&$81.7_{\pm{0.22}}$(\textcolor{blue}{$18.30$})
&$98.62_{\pm{0.04}}$(\textcolor{blue}{$1.29$})
&$92.34_{\pm{0.02}}$(\textcolor{blue}{$2.58$})
&\textcolor{blue}{$\mathbf{12.68}$}
&$1.56_{\pm{3.08}}$(\textcolor{blue}{$3.94$})
&$2.88_{\pm{3.44}}$(\textcolor{blue}{$8.69$})
&$98.67_{\pm{2.74}}$(\textcolor{blue}{$1.21$})
&$92.84_{\pm{2.59}}$(\textcolor{blue}{$1.40$})
&\textcolor{blue}{$\mathbf{3.81}$} 
\\
NegGrad+ \citep{kurmanji2024machine}
&$91.78_{\pm{14.66}}$(\textcolor{blue}{$8.22$})
&$95.81_{\pm{7.28}}$(\textcolor{blue}{$4.19$})
&$98.35_{\pm{1.22}}$(\textcolor{blue}{$1.56$})
&$92.62_{\pm{1.34}}$(\textcolor{blue}{$2.30$})
&\textcolor{blue}{{$\mathbf{4.07}$}} 
&$0.97_{\pm{1.08}}$(\textcolor{blue}{$4.53$})
&$2.74_{\pm{2.16}}$(\textcolor{blue}{$8.83$})
&$99.42_{\pm{0.87}}$(\textcolor{blue}{$0.46$})
&$93.38_{\pm{1.13}}$(\textcolor{blue}{$0.86$})
&\textcolor{blue}{$\mathbf{3.67}$}
\\
SalUn \citep{fan2023salun}
&\cb{mr}{$96.35_{\pm{2.14}}$(\textcolor{blue}{$3.65$})}
&$98.64_{\pm{0.03}}$(\textcolor{blue}{$1.36$})
&$98.75_{\pm{0.18}}$(\textcolor{blue}{$1.16$})
&$92.34_{\pm{1.54}}$(\textcolor{blue}{$2.58$})
&\cbs{mr}{\textcolor{blue}{{$\mathbf{2.19}$}} }
&$1.73_{\pm{0.25}}$(\textcolor{blue}{$3.77$})
&$6.25_{\pm{1.21}}$(\textcolor{blue}{$5.32$})
&$99.24_{\pm{0.09}}$(\textcolor{blue}{$0.64$})
&$91.03_{\pm{0.14}}$(\textcolor{blue}{$3.21$})
&\cbs{mr}{\textcolor{blue}{{$\mathbf{3.23}$}} }
\\
SCRUB \citep{kurmanji2024towards}
&{$93.45_{\pm{2.33}}$(\textcolor{blue}{$6.55$})}
&$96.38_{\pm{1.72}}$(\textcolor{blue}{$3.62$})
&\cb{mg}{$99.95_{\pm{0.00}}$(\textcolor{blue}{$0.04$})}
&\cb{mr}{$94.56_{\pm{0.07}}$(\textcolor{blue}{$0.36$})}
&\textcolor{blue}{{$\mathbf{2.64}$}} 
&$0.61_{\pm{0.31}}$(\textcolor{blue}{$4.89$})
&$3.69_{\pm{0.54}}$(\textcolor{blue}{$7.88$})
&$99.76_{\pm{0.18}}$(\textcolor{blue}{$0.12$})
&$93.91_{\pm{0.19}}$(\textcolor{blue}{$0.33$})
&\textcolor{blue}{$\mathbf{3.31}$} 
\\
Class-F \citep{kodge2024deep}
&{$90.18_{\pm{1.20}}$(\textcolor{blue}{$9.82$})}
&$86.15_{\pm{2.67}}$(\textcolor{blue}{$13.85$})
&$91.25_{\pm{0.05}}$(\textcolor{blue}{$8.66$})
&$85.45_{\pm{0.19}}$(\textcolor{blue}{$9.47$})
&\textcolor{blue}{{$\mathbf{10.45}$}} 
& n/a & n/a & n/a & n/a & n/a
\\
SSD \citep{foster2024fast}
&$96.05_{\pm{0.45}}$(\textcolor{blue}{$3.95$})
&$98.00_{\pm{0.00}}$(\textcolor{blue}{$2.00$})
&$97.77_{\pm{0.20}}$(\textcolor{blue}{$2.14$})
&$92.23_{\pm{0.58}}$(\textcolor{blue}{$2.69$})
&\textcolor{blue}{{$\mathbf{2.70}$}} 
&\cb{mg}{$5.54_{\pm{0.00}}$(\textcolor{blue}{$0.04$})}
&\cb{mr}{$7.80_{\pm{0.00}}$(\textcolor{blue}{$3.77$})}&$94.86_{\pm{0.00}}$(\textcolor{blue}{$5.02$})&$88.28_{\pm{0.00}}$(\textcolor{blue}{$5.96$})
&\textcolor{blue}{{$\mathbf{3.70}$}} 
\\
UAM \citep{kimunlearning}
&$97.57_{\pm{0.27}}$(\textcolor{blue}{$2.43$})
&$98.55_{\pm{0.10}}$(\textcolor{blue}{$1.45$})
&$99.08_{\pm{0.02}}$(\textcolor{blue}{$0.84$})
&$93.20_{\pm{0.11}}$(\textcolor{blue}{$1.72$})
&\cbs{mg}{\textcolor{blue}{{$\mathbf{1.61}$}}}
&$1.58_{\pm{0.19}}$(\textcolor{blue}{$3.92$})
&$0.50_{\pm{0.01}}$(\textcolor{blue}{$11.07$})
&$99.26_{\pm{0.03}}$(\textcolor{blue}{$0.62$})
&\cb{mg}{$94.22_{\pm{0.11}}$(\textcolor{blue}{$0.02$})}
&\textcolor{blue}{{$\mathbf{3.91}$}}
\\
Ours
&\cb{mg}{$97.88_{\pm{0.27}}$(\textcolor{blue}{{$2.12$}})}
&\cb{mg}{$99.60_{\pm{0.15}}$(\textcolor{blue}{{$0.40$}})}
&$97.25_{\pm{0.24}}$(\textcolor{blue}{{$2.66$}})
&$90.90_{\pm{0.32}}$(\textcolor{blue}{{$4.02$}})
&\textcolor{blue}{{$\mathbf{2.30}$}} 
&\cb{mr}{$2.61_{\pm{0.49}}$(\textbf{\textcolor{blue}{$2.89$}})}
&\cb{mg}{$8.26_{\pm{1.17}}$(\textcolor{blue}{{$3.31$}})}
&$97.33_{\pm{0.47}}$(\textcolor{blue}{{$2.55$}})
&$90.97_{\pm{0.38}}$(\textcolor{blue}{{$3.27$}})
&\cbs{mg}{\textcolor{blue}{{$\mathbf{3.01}$}} }
\\
\midrule
\rowcolor{Gray}
\multicolumn{1}{c}{{ }}&
\multicolumn{5}{c|}{{Class-wise Forgetting, ImageNet-10, ViT-Base}}
&\multicolumn{5}{c}{{Random Data Forgetting, ImageNet-10, ViT-Base}}
\\
\midrule
Retrain
&$100_{\pm{0.00}}$(\textcolor{blue}{$0.00$})
&$100.00_{\pm{0.00}}$(\textcolor{blue}{$0.00$})
&$99.97_{\pm{0.03}}$(\textcolor{blue}{$0.00$})
&$99.85_{\pm{0.01}}$(\textcolor{blue}{$0.00$})
&\textcolor{blue}{$\mathbf{0.00}$} 
&$1.41_{\pm{0.06}}$(\textcolor{blue}{$0.00$})
&$93.57_{\pm{0.00}}$(\textcolor{blue}{$0.00$})
&$99.07_{\pm{0.01}}$(\textcolor{blue}{$0.00$})
&$99.27_{\pm{0.01}}$(\textcolor{blue}{$0.00$})
&\textcolor{blue}{$\mathbf{0.00}$}
\\
\cmidrule(lr){2-11}
FT \citep{warnecke2021machine}
&$42.79_{\pm{7.51}}$(\textcolor{blue}{$57.21$})
&$40.78_{\pm{10.68}}$(\textcolor{blue}{$59.22$})
&\cb{mr}{$99.96_{\pm{0.01}}$(\textcolor{blue}{$0.01$})}
&$99.61_{\pm{0.10}}$(\textcolor{blue}{$0.24$})
&\textcolor{blue}{$\mathbf{29.17}$} 
&\cb{mg}{$1.38_{\pm{0.16}}$(\textcolor{blue}{$0.03$})}
&$96.40_{\pm{0.31}}$(\textcolor{blue}{$2.83$})&$99.60_{\pm{0.09}}$(\textcolor{blue}{$0.53$})&$99.10_{\pm{0.30}}$(\textcolor{blue}{$0.17$})
&\textcolor{blue}{$\mathbf{0.89}$} 
\\
RL \citep{golatkar2020eternal}
&$88.15_{\pm{1.62}}$(\textcolor{blue}{$11.85$})
&\cb{mr}{$96.50_{\pm{1.50}}$(\textcolor{blue}{$3.50$})}
&$99.93_{\pm{0.02}}$(\textcolor{blue}{$0.04$})
&\cb{mg}{$99.89_{\pm{0.11}}$(\textcolor{blue}{$0.04$})}
&\textcolor{blue}{$\mathbf{3.84}$} 
&$2.62_{\pm{0.80}}$(\textcolor{blue}{$1.21$})
&\cb{mg}{$93.75_{\pm{0.52}}$(\textcolor{blue}{$0.18$})}
&$98.38_{\pm{0.18}}$(\textcolor{blue}{$0.69$})
&$95.05_{\pm{0.26}}$(\textcolor{blue}{$4.20$})
&\textcolor{blue}{$\mathbf{1.57}$} 
\\
GA \citep{thudi2022unrolling}
&$28.05_{\pm{7.05}}$(\textcolor{blue}{$71.95$})
&$58.73_{\pm{5.83}}$(\textcolor{blue}{$41.27$})
&\cb{mg}{$99.97_{\pm{0.01}}$(\textcolor{blue}{$0.00$})}
&$99.70_{\pm{0.10}}$(\textcolor{blue}{$0.15$})
&\textcolor{blue}{$\mathbf{28.34}$}
&$0.82_{\pm{0.16}}$(\textcolor{blue}{$0.59$})&$96.77_{\pm{1.10}}$(\textcolor{blue}{$3.20$})&$99.60_{\pm{0.09}}$(\textcolor{blue}{$0.53$})&$99.53_{\pm{0.09}}$(\textcolor{blue}{$0.26$})
&\textcolor{blue}{$\mathbf{1.15}$} 
\\
NegGrad+ \citep{kurmanji2024machine}
&$10.90_{\pm{2.87}}$(\textcolor{blue}{$89.10$})
&$79.30_{\pm{4.58}}$(\textcolor{blue}{$20.70$})
&\cb{mr}{$99.98_{\pm{0.00}}$(\textcolor{blue}{$0.01$})}
&\cb{mr}{$99.78_{\pm{0.00}}$(\textcolor{blue}{$0.07$})}
&\textcolor{blue}{{$\mathbf{27.46}$}} 
&$1.97_{\pm{0.72}}$(\textcolor{blue}{$0.56$})&$95.48_{\pm{1.67}}$(\textcolor{blue}{$1.91$})&$98.09_{\pm{0.94}}$(\textcolor{blue}{$0.98$})&$98.08_{\pm{0.75}}$(\textcolor{blue}{$1.19$})
&\textcolor{blue}{$\mathbf{1.16}$}
\\
SalUn \citep{fan2023salun}
&$93.27_{\pm{1.50}}$(\textcolor{blue}{$6.73$})
&$94.00_{\pm{1.00}}$(\textcolor{blue}{$6.00$})&$98.22_{\pm{0.75}}$(\textcolor{blue}{$1.75$})&$98.00_{\pm{0.00}}$(\textcolor{blue}{$1.85$})
&\textcolor{blue}{{$\mathbf{4.08}$}} 
&$0.67_{\pm{0.19}}$(\textcolor{blue}{$0.74$})
&$95.80_{\pm{0.00}}$(\textcolor{blue}{$2.23$})&$99.65_{\pm{0.07}}$(\textcolor{blue}{$0.58$})
&\cb{mg}{$98.27_{\pm{0.09}}$(\textcolor{blue}{$0.00$})}
&\textcolor{blue}{{$\mathbf{1.14}$}} 
\\
SCRUB \citep{kurmanji2024towards}
&\cb{mg}{$99.10_{\pm{0.20}}$(\textcolor{blue}{$0.90$})}
&$95.67_{\pm{1.77}}$(\textcolor{blue}{$4.33$})
&$98.90_{\pm{0.10}}$(\textcolor{blue}{$1.07$})
&$99.33_{\pm{0.31}}$(\textcolor{blue}{$0.52$})
&\cbs{mr}{\textcolor{blue}{{$\mathbf{1.71}$}} }
&$0.85_{\pm{0.24}}$(\textcolor{blue}{$0.56$})&$95.90_{\pm{0.95}}$(\textcolor{blue}{$2.33$})
&\cb{mr}{$99.58_{\pm{0.23}}$(\textcolor{blue}{$0.51$})}
&\cb{mr}{$99.20_{\pm{0.14}}$(\textcolor{blue}{$0.07$})}
&\textcolor{blue}{$\mathbf{0.87}$} 
\\
Class-F \citep{kodge2024deep}
&$28.62_{\pm{7.85}}$(\textcolor{blue}{$71.38$})&$55.10_{\pm{2.30}}$(\textcolor{blue}{$44.90$})&$77.50_{\pm{1.62}}$(\textcolor{blue}{$22.47$})&$75.22_{\pm{0.56}}$(\textcolor{blue}{$24.63$})
&\textcolor{blue}{{$\mathbf{40.85}$}} 
& n/a & n/a & n/a & n/a & n/a
\\
SSD \citep{foster2024fast}
&$90.35_{\pm{1.65}}$(\textcolor{blue}{$9.65$})
&$60.15_{\pm{1.85}}$(\textcolor{blue}{$39.85$})
&$98.43_{\pm{0.55}}$(\textcolor{blue}{$1.54$})
&$98.33_{\pm{0.56}}$(\textcolor{blue}{$1.52$})
&\textcolor{blue}{{$\mathbf{13.14}$}} 
&\cb{mr}{$1.12_{\pm{0.04}}$(\textcolor{blue}{$0.29$})}
&\cb{mr}{$94.15_{\pm{0.35}}$(\textcolor{blue}{$0.58$})}
&\cb{mg}{$98.97_{\pm{0.00}}$(\textcolor{blue}{$0.10$})}
&$99.40_{\pm{0.00}}$(\textcolor{blue}{$0.13$})
&\cbs{mg}{\textcolor{blue}{$\mathbf{0.28}$} }
\\
UAM \citep{kimunlearning}
&$80.38_{\pm{2.40}}$(\textcolor{blue}{$19.62$})
&$81.97_{\pm{0.47}}$(\textcolor{blue}{$18.03$})
&$99.81_{\pm{0.06}}$(\textcolor{blue}{$0.16$})
&\cb{mr}{$99.78_{\pm{0.18}}$(\textcolor{blue}{$0.07$}) }
&\textcolor{blue}{{$\mathbf{9.47}$}}
&$0.54_{\pm{0.05}}$(\textcolor{blue}{$0.87$})
&$96.60_{\pm{4.50}}$(\textcolor{blue}{$3.03$})
&$99.75_{\pm{0.29}}$(\textcolor{blue}{$0.68$})
&$99.33_{\pm{0.25}}$(\textcolor{blue}{$0.06$})
&\textcolor{blue}{{$\mathbf{1.16}$}} 
\\
Ours
&\cb{mr}{$95.92_{\pm{0.27}}$(\textcolor{blue}{$4.08$})}
&\cb{mg}{$99.40_{\pm{0.14}}$(\textcolor{blue}{$0.60$})}
&$99.13_{\pm{0.04}}$(\textcolor{blue}{$0.84$})
&$99.26_{\pm{0.28}}$(\textcolor{blue}{$0.59$})
&\cbs{mg}{\textcolor{blue}{{$\mathbf{1.53}$}}}
&$1.08_{\pm{0.39}}$(\textcolor{blue}{$0.33$})&$91.40_{\pm{1.30}}$(\textcolor{blue}{$2.17$})
&\cb{mg}{$98.97_{\pm{0.35}}$(\textcolor{blue}{$0.10$})}
&$99.10_{\pm{0.50}}$(\textcolor{blue}{$0.17$})
&\cbs{mr}{\textcolor{blue}{{$\mathbf{0.69}$}} }
\\
\bottomrule[1pt]
\end{tabular}
    }
    }
    \label{tab: class_data_forget_table}
     \vspace{-0.2in}
\end{table*} 

\noindent \textbf{Overview Performance of Forget Vector.}
In Tab.\,\ref{tab: class_data_forget_table},
we compare the performance of our forget vector approach with other model-based MU methods across the metrics: UA, RA, TA, MIA-Efficacy, and Avg.\,Gap vs. {\Retrain}.
We highlight three key observations below.
\textbf{First}, in terms of unlearning effectiveness (UA and MIA-Efficacy), the forget vector demonstrates highly competitive performance compared to parameter-space MU baselines, mostly ranking among the top two methods with the smallest performance gap relative to {\Retrain} (as evidenced by Avg.\,Gap).
The advantage of the forget vector is particularly evident in MIA-Efficacy, where it usually achieves the closest results to {\Retrain}.
\textbf{Second}, in terms of model utility post-unlearning (RA and TA), the forget vector generally leads to a larger performance drop than other methods. This is not surprising, as the forget vector is achieved through data perturbations. 
However, considering the gain in unlearning effectiveness, the Avg.\,Gap with {\Retrain} shows that the forget vector remains competitive, ranking among the top two unlearning methods in three of the four scenarios.
\textbf{Third}, unlearning methods (including Retrain) do not exhibit the same level of distinctiveness in random data forgetting as it does in class-wise forgetting. This is because in random data forgetting, the retain data could have sufficiently represented the distribution of the forget data, making it more challenging for MIA to distinguish forgotten samples from retained ones.
Besides, the corresponding results of various MU methods on ImageNet-10 using VGG-16 can be found in appendix.

 \begin{table}[!t]
\caption{Compositional unlearning on CIFAR-10 and ImageNet-10 for random data forgetting, where FV represents the original setting of forget vector that is directly learned based on a targeted forget set, and CU-FV denotes compositional unlearning achieved via pre-learned class-wise forget vector arithmetic. The results are presented in the same format as Tab.\,\ref{tab: class_data_forget_table}.}
 \vspace{-0.12in}
\centering
\setlength{\tabcolsep}{0mm}{
\scalebox{0.51}{
\begin{tabular}{
    p{2.0cm}<{\centering}|
    p{1.9cm}<{\centering}|
    p{2.6cm}<{\centering}|
    p{2.6cm}<{\centering}|
    p{2.6cm}<{\centering}|
    p{2.6cm}<{\centering}|
    p{1.5cm}<{\centering}
}
\toprule[1pt]
\midrule
Module & MU Method & UA$\uparrow$ & MIA-Efficacy$\uparrow$ & RA$\uparrow$ & TA$\uparrow$ & Avg.Gap$\downarrow$ \\
\midrule
\multicolumn{1}{c|}{\multirow{3}{*}{\begin{tabular}{c} CIFAR-10 \\ ResNet-18 \end{tabular}}}
& Retrain & $5.50_{\pm{0.16}}$(\textcolor{blue}{$0.00$}) & $11.57_{\pm{0.47}}$(\textcolor{blue}{$0.00$}) & $99.88_{\pm{0.05}}$(\textcolor{blue}{$0.00$}) & $94.24_{\pm{0.19}}$(\textcolor{blue}{$0.00$}) & \textcolor{blue}{$\mathbf{0.00}$} \\
& FV & $2.61_{\pm{0.49}}$(\textcolor{blue}{$2.89$}) & $8.26_{\pm{1.17}}$(\textcolor{blue}{$3.31$}) & $97.33_{\pm{0.47}}$(\textcolor{blue}{$2.55$}) & $90.97_{\pm{0.38}}$(\textcolor{blue}{$3.27$}) & \textcolor{blue}{$\mathbf{3.01}$} \\
& CU-FV & $5.36_{\pm{0.60}}$(\textcolor{blue}{$0.14$}) & $9.76_{\pm{0.91}}$(\textcolor{blue}{$1.81$}) & $94.93_{\pm{0.64}}$(\textcolor{blue}{$4.95$}) & $88.60_{\pm{0.59}}$(\textcolor{blue}{$5.64$}) & \textcolor{blue}{$\mathbf{3.14}$} \\
\midrule
\multicolumn{1}{c|}{\multirow{3}{*}{\begin{tabular}{c} ImageNet-10 \\ VGG-16 \end{tabular}}}
& Retrain & $4.05_{\pm{0.45}}$(\textcolor{blue}{$0.00$}) & $6.60_{\pm{1.07}}$(\textcolor{blue}{$0.00$}) & $99.48_{\pm{0.07}}$(\textcolor{blue}{$0.00$}) & $96.33_{\pm{0.38}}$(\textcolor{blue}{$0.00$}) & \textcolor{blue}{$\mathbf{0.00}$} \\
& FV & $2.27_{\pm{0.50}}$(\textcolor{blue}{$1.78$}) & $6.13_{\pm{1.40}}$(\textcolor{blue}{$0.47$}) & $98.29_{\pm{0.32}}$(\textcolor{blue}{$1.19$}) & $95.82_{\pm{0.48}}$(\textcolor{blue}{$0.51$}) & \textcolor{blue}{$\mathbf{0.99}$} \\
& CU-FV & $2.27_{\pm{1.18}}$(\textcolor{blue}{$1.78$}) & $4.95_{\pm{1.86}}$(\textcolor{blue}{$1.65$}) & $97.93_{\pm{1.09}}$(\textcolor{blue}{$1.55$}) & $91.41_{\pm{1.25}}$(\textcolor{blue}{$4.92$}) & \textcolor{blue}{$\mathbf{2.47}$} \\
\midrule
\multicolumn{1}{c|}{\multirow{3}{*}{\begin{tabular}{c} ImageNet-10 \\ ViT-Base \end{tabular}}}
& Retrain & $1.41_{\pm{0.06}}$(\textcolor{blue}{$0.00$}) & $93.57_{\pm{0.00}}$(\textcolor{blue}{$0.00$}) & $99.07_{\pm{0.01}}$(\textcolor{blue}{$0.00$}) & $99.27_{\pm{0.01}}$(\textcolor{blue}{$0.00$}) & \textcolor{blue}{$\mathbf{0.00}$} \\
& FV & $1.08_{\pm{0.39}}$(\textcolor{blue}{$0.33$}) & $91.40_{\pm{1.30}}$(\textcolor{blue}{$2.17$}) & $98.97_{\pm{0.35}}$(\textcolor{blue}{$0.10$}) & $99.10_{\pm{0.50}}$(\textcolor{blue}{$0.17$}) & \textcolor{blue}{{$\mathbf{0.69}$}} \\
& CU-FV & $1.62_{\pm{0.05}}$(\textcolor{blue}{$0.21$}) & $92.09_{\pm{0.25}}$(\textcolor{blue}{$1.48$}) & $98.55_{\pm{0.14}}$(\textcolor{blue}{$0.52$}) & $98.50_{\pm{0.01}}$(\textcolor{blue}{$0.77$}) & \textcolor{blue}{$\mathbf{0.75}$} \\
\midrule
\bottomrule[1pt]
\end{tabular}
}
}
\label{tab: compositional_unlearning}
\vspace{-0.1in}
\end{table}

 \begin{figure}[t]
	\centering
	\includegraphics[scale=0.25]{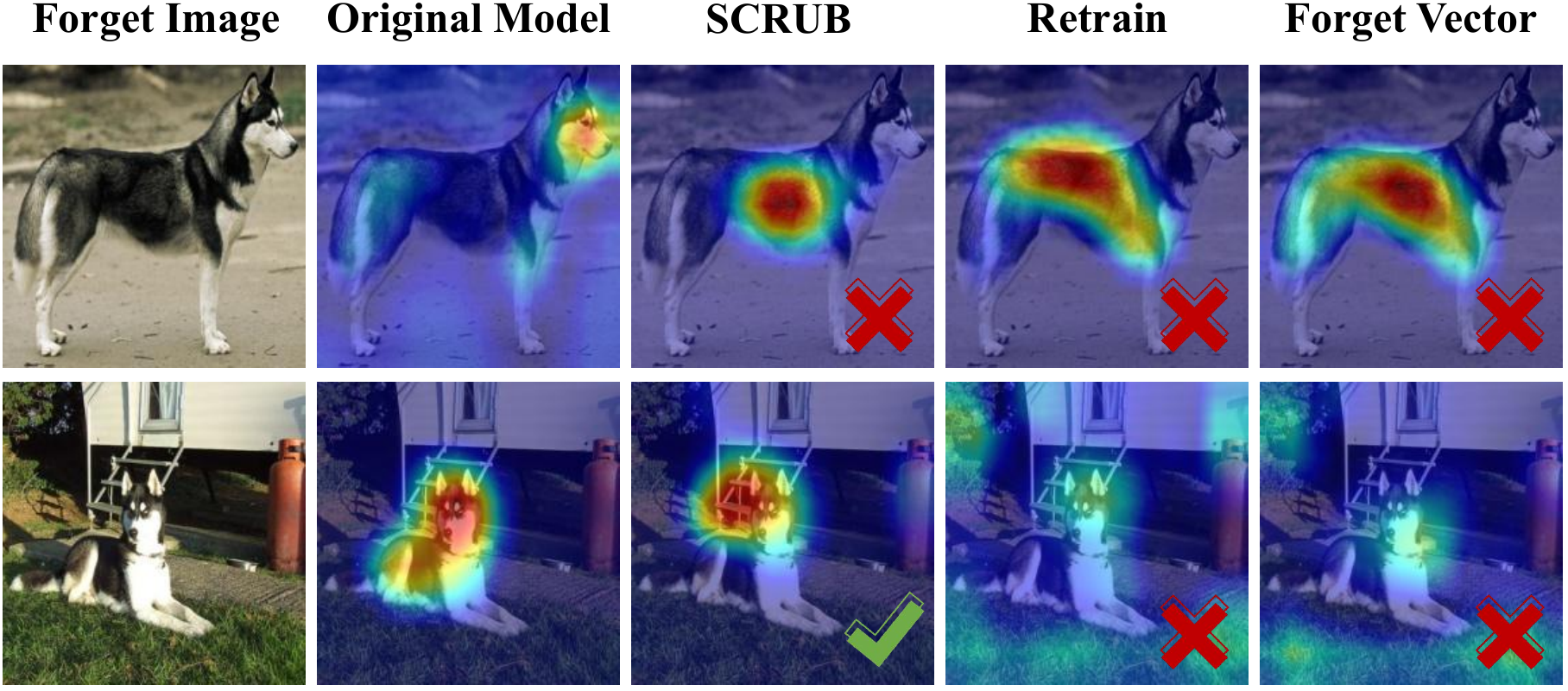}
\vspace{-0.1in}
\caption{
Gradient-based saliency map visualized via Grad-CAM for different MU methods against \textbf{forget images}. The highlighted areas (marked in red) indicate regions most influential to model prediction, and the red cross mark ({\color{red}{\faTimes}}) indicates that corresponding methods effectively unlearn the input forget images, while the check ({\color{green}{\faCheck}}) signifies the opposite.
}
\label{fig: grad_of_forget_images}
\vspace{-0.3in}
\end{figure}

\noindent \textbf{Compositional Unlearning by Class-wise Forget Vectors.}
Next, we demonstrate the effectiveness of \underline{c}ompositional \underline{u}nlearning via \underline{f}orget \underline{v}ector arithmetic (termed CU-FV). 
Given pre-computed class-wise forget vectors, we apply their linear combination as defined in \eqref{eq: compositional_unlearn} to achieve random 
data forgetting.
Tab.\,{\ref{tab: compositional_unlearning}}
compares the performance of CU-FV with  {\Retrain} (the exact 
unlearning method) and the direct forget vector approach (FV) applied to the targeted forget set.
Interestingly, we observe that CU-FV achieves the overall performance comparable to FV, as indicated by similar Avg.\,Gap values. Unlike FV, CU-FV optimizes only the class-wise coefficients in \eqref{eq: compositional_unlearn}, resulting in a much smaller optimization space than FV. However, from UA and MIA-Efficacy metrics, we find that unlearning effectiveness is easier to maintain since unlearning typically targets a smaller subset of data points. In contrast, model utility (RA and TA) may decline more with CU-FV than with FV.

\noindent \textbf{Assessing Forget Vector via an Input Saliency Lens.}
In Fig.\,\ref{fig: grad_of_forget_images}, 
we explore the impact of the forget vector on unlearning and utility retention through input saliency map, using Grad-CAM (Gradient-weighted Class Activation Mapping) \citep{selvaraju2017grad}. 
Using Grad-CAM, we visualize the salient 
pixels (\textit{i.e.}, regions most influential to model prediction) for forget images under different unlearning scenarios: 
(1) the original model (without unlearning), (2) {\Retrain}, (3) SCRUB-based unlearning, and (4) forget vector-based unlearning. 
In the first three scenarios, we obtain the input saliency maps on raw forget images without the addition of the forget vector, while in the last scenario, the input saliency is applied to images perturbed by the forget vector.
As seen in Fig.\,\ref{fig: grad_of_forget_images}
the forget vector and {\Retrain} effectively unlearn the target input images, evident from the significant shifts in saliency regions compared to those in the original model. In contrast, the MU baseline SCRUB shows minimal saliency shifts, failing to adequately forget the target data in the last row.
More visualization results on both forget and retain images can be found in appendix.
Also, more experimental analysis can be found in appendix.

\subsection{Forget Vectors for Image Generation}
\label{sec: exp_gen}

\noindent\textbf{Setups.}
\textit{Tasks and models.} Following SFR-on \citep{huang2024unified}, we study class-wise forgetting in two class-conditional models, the CIFAR-10 DDPM released by \citep{huang2024unified} (forgetting automobile, cat, dog, horse, or truck) and DiT-XL/2 at $256{\times}256$ on ImageNet \citep{peebles2023scalable} (forgetting sulphur-crested cockatoo, golden retriever, white wolf, arctic fox, or otter), as well as nudity removal in Stable Diffusion (SD) v1.4 \citep{rombach2022high}. Samplers follow \citep{huang2024unified}: $1000$-step DDIM \citep{song2020denoising} with guidance $2$ (DDPM), $250$-step DDPM sampling with guidance $4$ (DiT), and $100$ LMS steps with guidance $7.5$ (SD).

\noindent\textbf{Metrics.} \textbf{FA} is computed on $500$ images of the forgotten class with a ResNet-34 (CIFAR-10) or ViT-L/16 (ImageNet) classifier; \textbf{FID} uses the ADM evaluator on $45$k (CIFAR-10) or $10$k (ImageNet) generated retained-class images. For SD, NudeNet counts exposed parts in $9{,}406$ images generated from the $4{,}703$ I2P prompts \citep{schramowski2023safe}, CLIP score and FID are computed on $1{,}000$ COCO prompts \citep{lin2014microsoft}, and robustness is measured with UnlearnDiffAtk \citep{zhang2023generate}.

\noindent\textbf{Baselines.} For class-wise forgetting, we compare with representative baseline methods SA~\citep{heng2023selective}, SalUn~\citep{fan2023salun}, and SFR-on~\citep{huang2024unified}. For nudity removal, all baselines are evaluated in our pipeline with the same pretrained model, prompts, initial noise, and evaluators: SFR-on and SalUn are re-run with their official code; ESD, AdvUnlearn, and RECE use the weights released by their authors; SAFREE \citep{yoon2025safree}, a negative prompt (``nudity''), and SLD \citep{schramowski2023safe} are training-free. On SD, the forget vector uses the same forget and retain data as the re-trained baselines; its hyper-parameters are selected once per model on proxy runs with selection rules fixed in advance, and the same configuration is used for all classes.

\begin{table}[t]
\centering
\caption{Class-wise forgetting in class-conditional diffusion models.}
\vspace{-0.1in}
\setlength{\tabcolsep}{1.3mm}
\scalebox{0.52}{
\begin{tabular}{l|cc|cc|cc|cc|cc}
\toprule[1pt]
\textbf{MU Method} & \multicolumn{2}{c|}{Automobile} & \multicolumn{2}{c|}{Cat} & \multicolumn{2}{c|}{Dog} & \multicolumn{2}{c|}{Horse} & \multicolumn{2}{c}{Truck} \\
CIFAR-10 DDPM & FA$\downarrow$ & FID$\downarrow$ & FA$\downarrow$ & FID$\downarrow$ & FA$\downarrow$ & FID$\downarrow$ & FA$\downarrow$ & FID$\downarrow$ & FA$\downarrow$ & FID$\downarrow$ \\
\midrule
Pretrained & 99.80 & 19.35 & 99.00 & 18.89 & 99.40 & 18.54 & 100.00 & 18.59 & 100.00 & 19.21 \\
SA \citep{heng2023selective} & 0.00 & 23.56 & 14.20 & 21.34 & 8.60 & 21.19 & 0.00 & 21.13 & 0.00 & 29.04 \\
SalUn \citep{fan2023salun} & 0.20 & 21.23 & 1.40 & 20.29 & 0.00 & 20.18 & 0.60 & 20.70 & 0.80 & 20.45 \\
SFR-on \citep{huang2024unified} & 0.00 & 20.70 & 7.40 & 18.44 & 0.20 & 18.89 & 0.00 & 19.93 & 0.00 & 20.61 \\
\textbf{Forget vector (Ours)} & \textbf{0.20} & \textbf{19.21} & \textbf{0.80} & \textbf{18.21} & \textbf{0.20} & \textbf{18.21} & \textbf{0.00} & \textbf{19.49} & \textbf{0.40} & \textbf{18.37} \\
\midrule
\textbf{ImageNet DiT-XL/2} & \multicolumn{2}{c|}{Cockatoo} & \multicolumn{2}{c|}{Golden retriever} & \multicolumn{2}{c|}{White wolf} & \multicolumn{2}{c|}{Arctic fox} & \multicolumn{2}{c}{Otter} \\
\midrule
Pretrained & 100.00 & 24.63 & 100.00 & 24.63 & 99.60 & 24.63 & 100.00 & 24.63 & 100.00 & 24.63 \\
SA \citep{heng2023selective} & 0.00 & 348.75 & 0.00 & 298.97 & 0.00 & 45.89 & 0.00 & 393.91 & 29.80 & 321.21 \\
SalUn \citep{fan2023salun} & 91.21 & 18.47 & 46.09 & 25.28 & 0.00 & 15.16 & 45.90 & 408.07 & 87.50 & 19.69 \\
SFR-on~\citep{huang2024unified} & 0.00 & 13.59 & 0.00 & 17.76 & 0.00 & 23.28 & 0.00 & 16.12 & 0.00 & 16.43 \\
\textbf{Forget vector (Ours)} & \textbf{2.00} & \textbf{19.09} & \textbf{0.20} & \textbf{18.29} & \textbf{0.80} & \textbf{16.91} & \textbf{2.00} & \textbf{18.73} & \textbf{1.00} & \textbf{17.77} \\
\bottomrule[1pt]
\end{tabular}}
\label{tab: gen_class}
\vspace{-0.25in}
\end{table}

\vspace{-0.2in}
\begin{figure}[!h]
\centering
\includegraphics[width=0.95\linewidth]{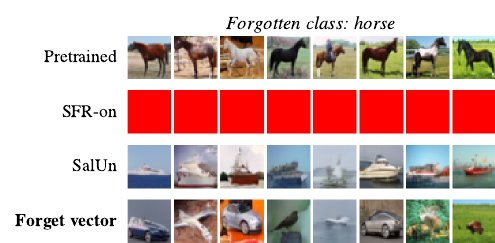}
\vspace{-0.1in}
\caption{Samples generated by the CIFAR-10 DDPM under the forgotten class \textit{horse}, with identical initial noise. From top to bottom: pretrained model, SFR-on, SalUn, and the forget vector. SFR-on produces near-uniform color images, SalUn turns the forgotten class into a single other class, and the forget vector yields natural images of various other classes.}
\label{fig: gen_samples_main}
\vspace{-0.15in}
\end{figure}

\noindent{\textbf{Class-wise Forgetting on CIFAR-10.}
As the upper half of Tab.\,\ref{tab: gen_class} shows, the forget vector forgets every class (FA $0$--$0.8\%$), on par with weight editing, and leaves retained-class generation essentially unchanged: relative to the pretrained model, the FID changes by $-0.84$ to $+0.90$ (mean $-0.22$).
The forget vector also changes what is generated for the forgotten class in a more desirable way (Fig.\,\ref{fig: gen_samples_main}): SFR-on produces near-uniform color images, and SalUn maps the whole class to one other class ( horse$\to$ship), whereas the forget vector yields natural images of various other classes.\par}

\noindent{\textbf{Class-wise Forgetting on ImageNet DiT.}
The lower half of Tab.\,\ref{tab: gen_class} shows that the forget vector consistently suppresses all five target classes, reducing FA from $99.6$--$100\%$ to $0.2$--$2.0\%$ without modifying model weights. In comparison, SalUn leaves four classes at $45.9$--$91.2\%$ FA, while SFR-on achieves zero FA on all five.
The forget vector yields retained-class FIDs of $16.91$--$19.09$, below the pretrained model's $24.63$. 
These results demonstrate that forget vectors achieve competitive performance in class-wise forgetting and retained generation quality on large-scale diffusion models, without modifying pretrained model parameters.

\begin{table}[t]
\centering
\caption{Quantitative comparison of nudity removal and retained generation quality on Stable Diffusion v1.4.}
\vspace{-0.15in}
\setlength{\tabcolsep}{0.9mm}
\scalebox{0.55}{
\begin{tabular}{lc|cc|cc}
\toprule[1pt]
Method & Type & I2P$\downarrow$ & Sens.$\downarrow$ & CLIP$\uparrow$ & FID$\downarrow$ \\
\midrule
SD v1.4 & -- & 4030 & 1551 & 26.60 & -- \\
\midrule
ESD \citep{gandikota2023erasing} & W & 1117 (\,$-72\%$) & 356 & 25.53 & 42.8 \\
SFR-on \citep{huang2024unified} & W & 667 (\,$-83\%$) & 240 & 25.16 & 48.3 \\
SalUn \citep{fan2023salun} & W & 289 (\,$-93\%$) & 67 & 24.95 & 49.7 \\
AdvUnlearn \citep{zhang2024defensive} & W & 510 (\,$-87\%$) & 185 & 23.97 & 44.2 \\
RECE \citep{gong2024reliable} & W & 956 (\,$-76\%$) & 327 & 25.99 & 40.8 \\
\midrule
SAFREE \citep{yoon2025safree} & TF & 1048 (\,$-74\%$) & 275 & 25.88 & 40.1 \\
Neg.\ prompt & TF & 730 (\,$-82\%$) & 244 & 25.99 & 39.4 \\
SLD-Med \citep{schramowski2023safe} & TF & 1987 (\,$-51\%$) & 515 & 25.24 & 39.6 \\
SLD-Max \citep{schramowski2023safe} & TF & 1168 (\,$-71\%$) & 211 & 23.23 & 52.8 \\
\midrule
\textbf{Forget vector} (Ours) & In & $575_{\pm86}$ (\,$-86\%$) & $190_{\pm23}$ &$26.41_{\pm0.07}$ & $41.3_{\pm0.3}$ \\
\bottomrule[1pt]
\end{tabular}}
\label{tab: gen_sd}
\vspace{-0.25in}
\end{table}

\noindent\textbf{Nudity Removal.}
As shown in Tab.\,\ref{tab: gen_sd}, our forget vector effectively suppresses nudity-related content, reducing I2P detections by 86\% (from 4030 to $575\pm86$) and sensitive-part detections by 88\% (from 1551 to $190\pm23$).
Notably, our method achieves the highest CLIP score of $26.41\pm0.07$ among all unlearning baselines, with a minimal decrease of only 0.19 from the pretrained model.
Compared with weight-editing approaches, our method achieves stronger nudity suppression than ESD, SFR-on, and RECE while better preserving text-image alignment.
Although SalUn and AdvUnlearn achieve greater suppression, they incur substantially larger reductions in CLIP score.
Furthermore, our method consistently outperforms training-free approaches in both nudity suppression and CLIP score, while maintaining a competitive FID of $41.3\pm0.3$.
Overall, these results show that forget vectors achieve a competitive balance between concept suppression and retained generation quality without modifying pretrained model parameters.

\iffalse
\noindent{\color{fvAtk}\textbf{Robustness to Adversarial Prompts.}
As the attack columns (Pre and ASR) of Tab.\,\ref{tab: gen_sd} show, the forget vector is not robust to adversarial prompts: UnlearnDiffAtk succeeds on $78\%$ of the attacked prompts, similar to ESD ($76\%$) and the negative prompt ($80\%$). The two markedly more robust methods edit the weights (SalUn, $30\%$) or are trained adversarially (AdvUnlearn, $26\%$). Since $\bdelta_c$ is a fixed offset in the text-embedding space, optimized prompt tokens can counteract it; training the forget vector adversarially is a natural next step.\par}
\noindent\textbf{Which Input Matters?}
On CIFAR-10, a forget vector restricted to the pixel space ($\bdelta_c = \mathbf{0}$) never forgets automobile (FA $100\%$ in all six configurations we tried), whereas the condition-only variant ($\bdelta_x = \mathbf{0}$) performs on par with the full one (mean FA $0.28\%$ vs.\ $0.32\%$, mean FID $18.70$ for both). Unlearning thus happens in the condition embedding, the most direct lever on what a conditional generator produces.
\fi

% \vspace*{-2mm}
\section{Conclusion}
% \vspace*{-1mm}
In this paper, we introduce a novel, input-space approach to machine unlearning (MU) in image classification and generation, termed the \textit{forget vector}. Unlike traditional model-based MU methods that require retraining or fine-tuning model weights, our approach shows that universal input perturbations can effectively achieve unlearning objectives. Our method demonstrates competitive performance relative to model-based approximate unlearning techniques. Furthermore, we showcase the potential of compositional unlearning: new forget vectors for unseen tasks, such as unlearning arbitrary subsets across classes, can be generated through simple arithmetic operations, like linear combinations of class-wise forget vectors. Extensive experiments confirm the effectiveness and adaptability of our optimized forget vector.

{
    \small
    \bibliographystyle{ieeenat_fullname}
    \bibliography{main}
}
\clearpage

\setcounter{section}{0}
\appendix
\section*{Appendix}
\setcounter{section}{0}
\setcounter{figure}{0}
\makeatletter 
\renewcommand{\thefigure}{A\arabic{figure}}%
\renewcommand{\theHfigure}{A\arabic{figure}}%
\renewcommand{\thetable}{A\arabic{table}}
\renewcommand{\theHtable}{A\arabic{table}}

\makeatother
\setcounter{table}{0}

\section{Positioning of the Forget Vector within Approximate MU}
\label{sec: positioning}

Our forget vector $\bdelta$ 
realizes unlearning as an input-space perturbation rather than a parameter-space update, in both image classification and image generation. Following the standard convention of approximate MU, we regard the deployed unlearned system as the pair $(\thetao, \bdelta)$ and evaluate it by its functional behavior. For image classification, unlearning quality is assessed by the functional closeness to {\Retrain} under the standard metrics UA, MIA-Efficacy, RA, and TA. For image generation, where retraining a large generator from scratch is infeasible, the pretrained model and model-based unlearning methods serve as references, and the system is assessed by the forgetting of the target class or concept (FA, or NudeNet detections for nudity) and by the quality of the retained generation (FID for class-wise forgetting, and CLIP score and FID for nudity removal), as in~\citep{fan2023salun,huang2024unified}. This is the same criterion used to evaluate model-based approximate MU, which likewise approximates {\Retrain} for image classifiers or the intended forgetting behavior for image generators, in functional behavior rather than verifying weight-level removal of training-data influence in $\thetau$~\citep{thudi2022unrolling,kurmanji2024towards}.

We emphasize that our forget vector is applied \emph{universally}: at test time, the \emph{same} $\bdelta$ is added to every input regardless of whether it comes from the forget set $\mathcal{D}_\mathrm{f}$ or the retain set $\mathcal{D}_\mathrm{r}$ (and likewise to the test set); in a conditional diffusion model, $\bdelta = (\bdelta_x, \bdelta_c)$ is added to the noisy input at every denoising step and to the embedding of every condition, whether the condition is forgotten or not (only the null condition of classifier-free guidance is left unperturbed). The perturbation is therefore data-agnostic and symmetric across forget and retain inputs, rather than being applied conditionally only to forget-data inputs, so that no request has to be identified as targeting the forgotten data, class, or concept at deployment. The two paradigms therefore share the same functional evaluation, while differing in how unlearning is enacted: through the parameter shift $\thetau-\thetao$ for model-based methods, and through the input perturbation $\bdelta$ for our forget vector.

Localizing unlearning in the input space carries structural consequences distinct from parameter-space methods. \textbf{First}, the original model $\thetao$ is preserved by construction, so that removing $\bdelta$ exactly restores the pre-unlearning system without storing additional model checkpoints. \textbf{Second}, since each $\bdelta$ is optimized against the same already-trained $\thetao$, continual unlearning requests~\citep{chatterjee2024unified,adhikari2025unlearning} do not compound parameter drift, in contrast to model-based methods where subsequent requests act on an already-modified model. \textbf{Third}, our forget vector $\bdelta$ is a single low-dimensional perturbation rather than a full updated weight set, leading to lower storage overhead. For example, for the generative models, our forget vector $\bdelta = (\bdelta_x, \bdelta_c)$ has $\bdelta_x$ of the shape of the noisy input (pixels for the DDPM, VAE latents for DiT and SD) and $\bdelta_c$ of the shape of the condition embedding (a class embedding for the DDPM and DiT, the CLIP text hidden states for SD), and is shared across all denoising steps and conditions. In total, it has $3{,}200$ entries for CIFAR-10 DDPM ($3{\times}32{\times}32 = 3{,}072$ in $\bdelta_x$ and $128$ in $\bdelta_c$), $5{,}248$ for DiT-XL/2 ($4{\times}32{\times}32 = 4{,}096$ and $1{,}152$), and $75{,}520$ for SD v1.4 ($4{\times}64{\times}64 = 16{,}384$ and $77{\times}768 = 59{,}136$). In contrast, SFR-on fine-tunes denoisers with $38.6$M, $675$M, and $860$M parameters, respectively, \textit{i.e.}, $10^4$--$10^5$ times more, and each unlearning request yields a full copy of the fine-tuned denoiser (Sec.\,\ref{sec: gen_efficiency_sec}). 
\textbf{Fourth}, these properties rest on a different trust assumption: unlearning holds for the deployed system $(\thetao, \bdelta)$ rather than for $\thetao$ alone, so the model provider must apply $\bdelta$ to every input. 
In this sense, our forget vector serves as an input-space counterpart to model-based approximate MU for both classifiers and generators, with complementary properties in reversibility, sequential stability, and storage.

\begin{table*}[htb]
\centering
   \caption{
Performance overview of various 
MU methods for image classification under
two unlearning scenarios
on ImageNet-10 using VGG-16. 
Since Class-F \citep{kodge2024deep} is specifically designed for class-wise forgetting, its results do not apply to random data forgetting scenarios (n/a).
Other results are reported in the format $a_{\pm b}$, where $a$ is the mean and $b$ denotes standard deviation $b$ over $10$ independent trials. 
The performance gap against Retrain is indicated 
in (\textcolor{blue}{$\bullet$}), where a lower value is better.
$\uparrow$ (or $\downarrow$) indicates that a higher (or lower) value is better.
The best performance for each metric is highlighted in 
\cbsg{mg}{{\text{pink}}}, while the second-best performance is highlighted in 
\cbsr{mr}{{\text{yellow}}}.
}
\vspace*{-2mm}
\setlength{\tabcolsep}{0mm}{
\scalebox{0.48}{
\begin{tabular}{
p{2.2cm}<{\centering}|
p{2.8cm}<{\centering}
p{2.8cm}<{\centering}
p{2.8cm}<{\centering}
p{2.8cm}<{\centering}
p{1.5cm}<{\centering}|
p{2.8cm}<{\centering}
p{2.8cm}<{\centering}
p{2.8cm}<{\centering}
p{2.8cm}<{\centering}
p{1.5cm}<{\centering}
} 
\toprule[1pt]
\multicolumn{1}{c|}{\multirow{1}{*}{MU Method}}
&UA$\uparrow$ & MIA-Efficacy$\uparrow$ & RA$\uparrow$ & TA$\uparrow$& \textbf{Avg.Gap}$\downarrow$ 
&UA$\uparrow$ & MIA-Efficacy$\uparrow$ & RA$\uparrow$ & TA$\uparrow$& \textbf{Avg.Gap}$\downarrow$ 
\\
\midrule
\rowcolor{Gray}
\multicolumn{1}{c}{{ }}
&\multicolumn{5}{c}{{Class-wise Forgetting, ImageNet-10, VGG-16}}
&\multicolumn{5}{c}{{Random Data Forgetting, ImageNet-10, VGG-16}}
\\
\midrule
Retrain	
&$100.00_{\pm{0.00}}$(\textcolor{blue}{$0.00$})
&$100.00_{\pm{0.00}}$(\textcolor{blue}{$0.00$})
&$99.66_{\pm{0.16}}$(\textcolor{blue}{$0.00$})
&$97.11_{\pm{0.82}}$(\textcolor{blue}{$0.00$})
&\textcolor{blue}{$\mathbf{0.00}$} 
&$4.05_{\pm{0.45}}$(\textcolor{blue}{$0.00$})
&$6.60_{\pm{1.07}}$(\textcolor{blue}{$0.00$})
&$99.48_{\pm{0.07}}$(\textcolor{blue}{$0.00$})
&$96.33_{\pm{0.38}}$(\textcolor{blue}{$0.00$})
&\textcolor{blue}{$\mathbf{0.00}$} 
\\
\cmidrule(lr){2-11}
FT \citep{warnecke2021machine}
&$39.66_{\pm{4.73}}$(\textcolor{blue}{$60.34$})
&$55.76_{\pm{7.26}}$(\textcolor{blue}{$44.24$})
&\cb{mr}{$99.78_{\pm{0.03}}$(\textcolor{blue}{$0.12$})}
&$97.27_{\pm{0.35}}$(\textcolor{blue}{$0.16$}) 
&\textcolor{blue}{$\mathbf{26.22}$} 
&$1.35_{\pm{0.32}}$(\textcolor{blue}{$2.70$})
&$4.67_{\pm{1.61}}$(\textcolor{blue}{$1.93$})
&
\cb{mg}{$99.36_{\pm{0.28}}$(\textcolor{blue}{$0.12$})}
&\cb{mr}{$96.54_{\pm{0.59}}$(\textcolor{blue}{$0.21$})}
&\textcolor{blue}{$\mathbf{1.24}$}
\\
RL \citep{golatkar2020eternal}
&$76.58_{\pm{11.64}}$(\textcolor{blue}{$23.42$})
&$46.04_{\pm{33.71}}$(\textcolor{blue}{$53.96$})
&$99.28_{\pm{0.20}}$(\textcolor{blue}{$0.38$})
&$96.91_{\pm{0.55}}$(\textcolor{blue}{$0.20$})
&\textcolor{blue}{$\mathbf{19.49}$}
&\cb{mg}{$2.96_{\pm{0.42}}$(\textcolor{blue}{$1.09$})}
&$12.85_{\pm{4.25}}$(\textcolor{blue}{$6.25$})
&$99.19_{\pm{0.16}}$(\textcolor{blue}{$0.29$})
&$95.50_{\pm{0.90}}$(\textcolor{blue}{$0.83$})
&\textcolor{blue}{$\mathbf{2.12}$}
\\
GA \citep{thudi2022unrolling}
&$46.61_{\pm{6.11}}$(\textcolor{blue}{$53.39$})
&$49.15_{\pm{9.36}}$(\textcolor{blue}{$50.85$})
&$99.35_{\pm{0.11}}$(\textcolor{blue}{$0.31$})
&$95.60_{\pm{0.22}}$(\textcolor{blue}{$1.51$}) 
&\textcolor{blue}{$\mathbf{26.52}$} 
&$0.18_{\pm{0.04}}$(\textcolor{blue}{$3.87$})
&$2.97_{\pm{1.51}}$(\textcolor{blue}{$3.63$})
&$99.86_{\pm{0.01}}$(\textcolor{blue}{$0.38$})
&$97.47_{\pm{0.09}}$(\textcolor{blue}{$1.14$})
&\textcolor{blue}{$\mathbf{2.23}$} 
\\
NegGrad+ \citep{kurmanji2024machine}
&$49.56_{\pm{34.87}}$(\textcolor{blue}{$50.44$})
&$64.27_{\pm{27.33}}$(\textcolor{blue}{$35.73$})
&$99.08_{\pm{1.18}}$(\textcolor{blue}{$0.58$})
&$96.47_{\pm{1.25}}$(\textcolor{blue}{$0.64$})
&\textcolor{blue}{$\mathbf{21.84}$}
&$0.60_{\pm{0.44}}$(\textcolor{blue}{$3.45$})
&$3.90_{\pm{2.22}}$(\textcolor{blue}{$2.70$})
&\cb{mr}{$99.68_{\pm{0.28}}$(\textcolor{blue}{$0.20$})}
&$97.12_{\pm{0.45}}$(\textcolor{blue}{$0.79$})
&\textcolor{blue}{{$\mathbf{1.79}$}}
\\
SalUn \citep{fan2023salun}
&\cb{mr}{$95.13_{\pm{1.79}}$(\textcolor{blue}{$4.87$})}
&\cb{mr}{$97.24_{\pm{0.17}}$(\textcolor{blue}{$2.76$})}
&$96.33_{\pm{0.25}}$(\textcolor{blue}{$3.33$})
&$96.18_{\pm{1.10}}$(\textcolor{blue}{$0.93$})
&\cbs{mr}{\textcolor{blue}{{$\mathbf{2.97}$}} }
&$1.15_{\pm{0.40}}$(\textcolor{blue}{$2.90$})
&$3.56_{\pm{1.12}}$(\textcolor{blue}{$3.04$})
&$98.97_{\pm{1.56}}$(\textcolor{blue}{$0.51$})
&$95.42_{\pm{2.10}}$(\textcolor{blue}{$0.91$})
&\textcolor{blue}{{$\mathbf{1.84}$}} 
\\
SCRUB \citep{kurmanji2024towards}
&\cb{mg}{$99.12_{\pm{0.14}}$(\textcolor{blue}{$0.88$})}
&\cb{mg}{$98.01_{\pm{0.51}}$(\textcolor{blue}{$1.99$})}
&\cb{mg}{$99.75_{\pm{0.03}}$(\textcolor{blue}{$0.09$})}
&\cb{mr}{$97.24_{\pm{0.16}}$(\textcolor{blue}{$0.13$})}
&\cbs{mg}{\textcolor{blue}{$\mathbf{0.77}$}}
&$0.18_{\pm{0.08}}$(\textcolor{blue}{$3.87$})
&$2.80_{\pm{1.33}}$(\textcolor{blue}{$3.80$})
&$99.88_{\pm{0.03}}$(\textcolor{blue}{$0.40$})
&$97.36_{\pm{0.15}}$(\textcolor{blue}{$1.03$})
&\textcolor{blue}{{$\mathbf{2.28}$}}
\\
Class-F \citep{kodge2024deep}
&{$71.19_{\pm{3.50}}$(\textcolor{blue}{$28.81$})}
&$63.57_{\pm{4.07}}$(\textcolor{blue}{$36.43$})
&$61.55_{\pm{4.25}}$(\textcolor{blue}{$38.11$})
&$59.40_{\pm{4.59}}$(\textcolor{blue}{$37.71$})
&\textcolor{blue}{{$\mathbf{35.27}$}} 
& n/a & n/a & n/a & n/a & n/a
\\
SSD \citep{foster2024fast}
&$91.50_{\pm{0.96}}$(\textcolor{blue}{$8.50$})
&$91.30_{\pm{8.70}}$(\textcolor{blue}{$8.70$})
&$99.48_{\pm{0.18}}$(\textcolor{blue}{$0.18$})
&\cb{mg}{$97.00_{\pm{0.11}}$(\textcolor{blue}{$0.11$})}
&\textcolor{blue}{{$\mathbf{4.38}$}} 
&$0.88_{\pm{0.04}}$(\textcolor{blue}{$3.17$})
&\cb{mr}{$5.35_{\pm{0.45}}$(\textcolor{blue}{$1.25$})}
&\cb{mg}{$99.36_{\pm{0.15}}$(\textcolor{blue}{$0.12$})}
&\cb{mg}{$96.40_{\pm{0.40}}$(\textcolor{blue}{$0.07$})}
&\cbs{mr}{\textcolor{blue}{{$\mathbf{1.15}$}} }
\\
UAM \citep{kimunlearning}
&$94.17_{\pm{0.21}}$(\textcolor{blue}{$5.83$})
&$94.60_{\pm{2.30}}$(\textcolor{blue}{$5.40$})
&$97.97_{\pm{2.08}}$(\textcolor{blue}{$1.69$})
&$94.37_{\pm{0.27}}$(\textcolor{blue}{$2.74$})
&\textcolor{blue}{{$\mathbf{3.92}$}}
&$2.17_{\pm{0.47}}$(\textcolor{blue}{$1.88$})
&$7.08_{\pm{0.36}}$(\textcolor{blue}{$0.48$})
&$98.59_{\pm{1.15}}$(\textcolor{blue}{$0.89$})
&$95.70_{\pm{9.41}}$(\textcolor{blue}{$0.63$})
&\textcolor{blue}{{$\mathbf{0.97}$}} 
\\
Ours
&$87.23_{\pm{6.55}}$(\textcolor{blue}{{$12.77$}})
&$91.41_{\pm{5.90}}$(\textcolor{blue}{{$8.59$}})
&$94.77_{\pm{1.16}}$(\textcolor{blue}{$4.89$})
&$94.04_{\pm{1.29}}$(\textcolor{blue}{$3.07$})
&\textcolor{blue}{$\mathbf{7.33}$}
&\cb{mr}{$2.27_{\pm{0.50}}$(\textcolor{blue}{{$1.78$}})}
&\cb{mg}{$6.13_{\pm{1.40}}$(\textcolor{blue}{{$0.47$}})}
&$98.29_{\pm{0.32}}$(\textcolor{blue}{{$1.19$}})
&$95.82_{\pm{0.48}}$(\textcolor{blue}{{$0.51$}})
&\cbs{mg}{\textcolor{blue}{{$\mathbf{0.99}$}}}
\\
\bottomrule[1pt]
\end{tabular}
    }
    }
    \label{tab: class_data_forget_table_using_vgg16}
     \vspace{-0.05in}
\end{table*} 

\begin{figure*}[!t]
\centering
\begin{tabular}{cccc}
  \includegraphics[width=0.22\textwidth,height=!]{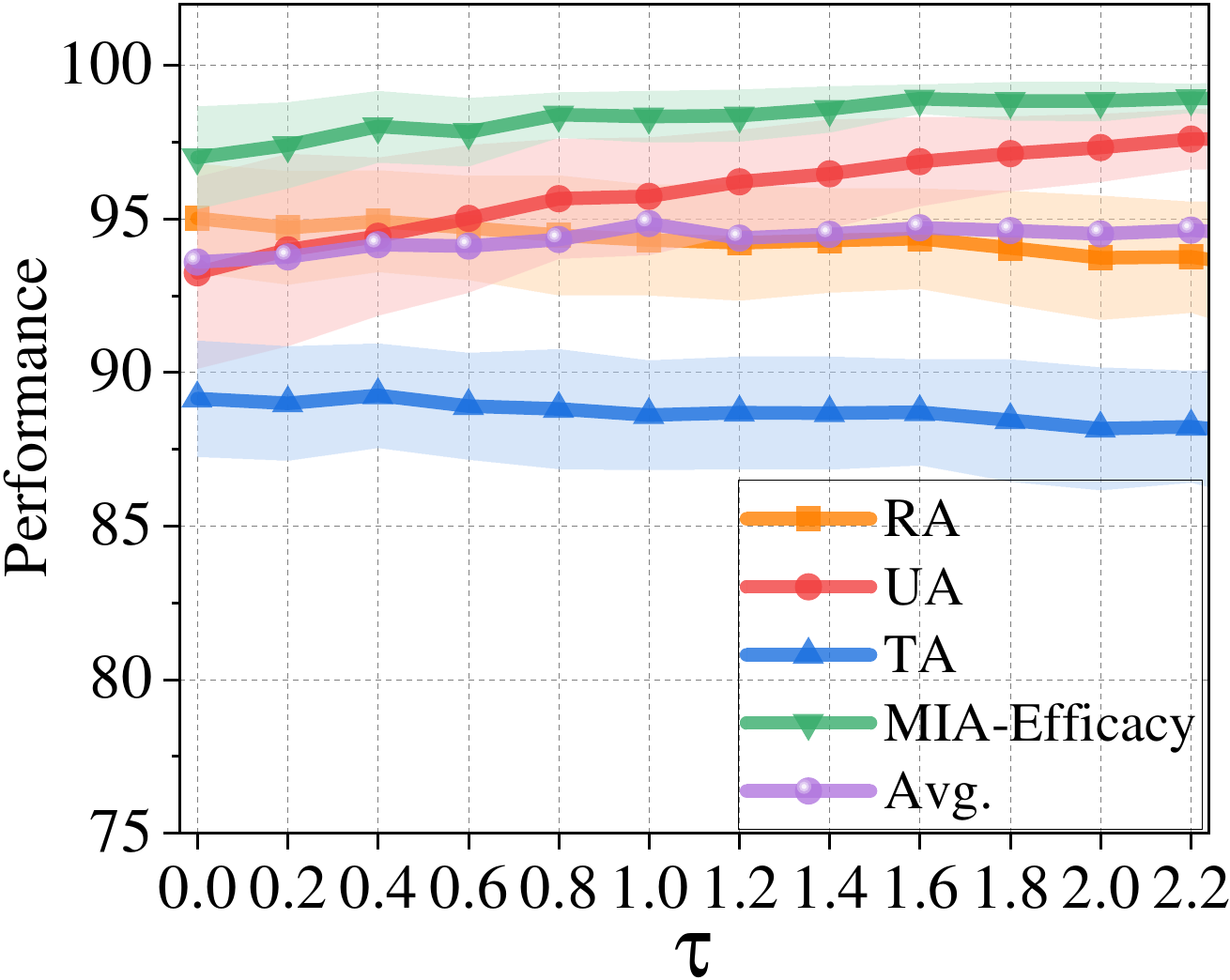}    &   \includegraphics[width=0.22\textwidth,height=!]{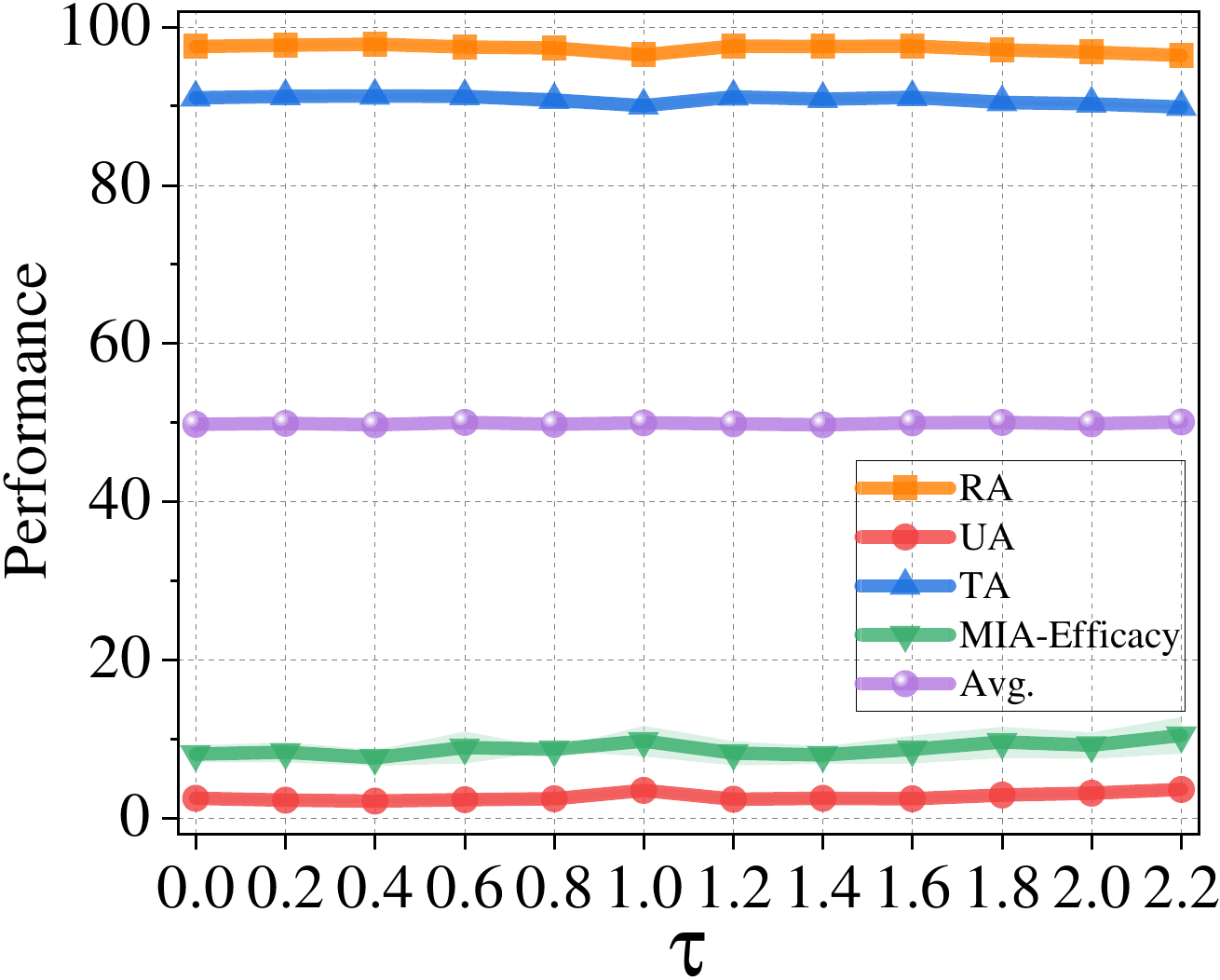}  &   \includegraphics[width=0.22\textwidth,height=!]{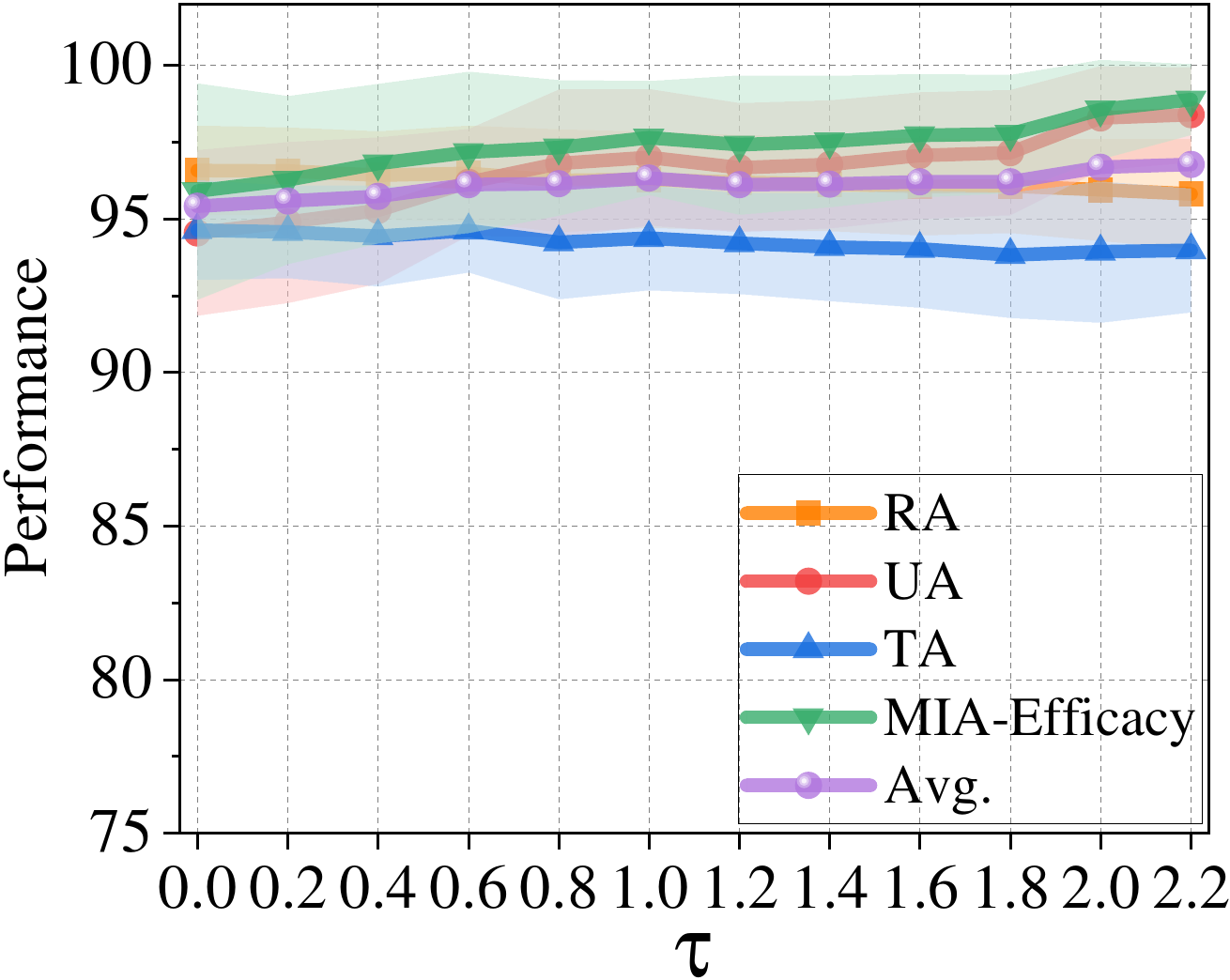}  &    \includegraphics[width=0.22\textwidth,height=!]{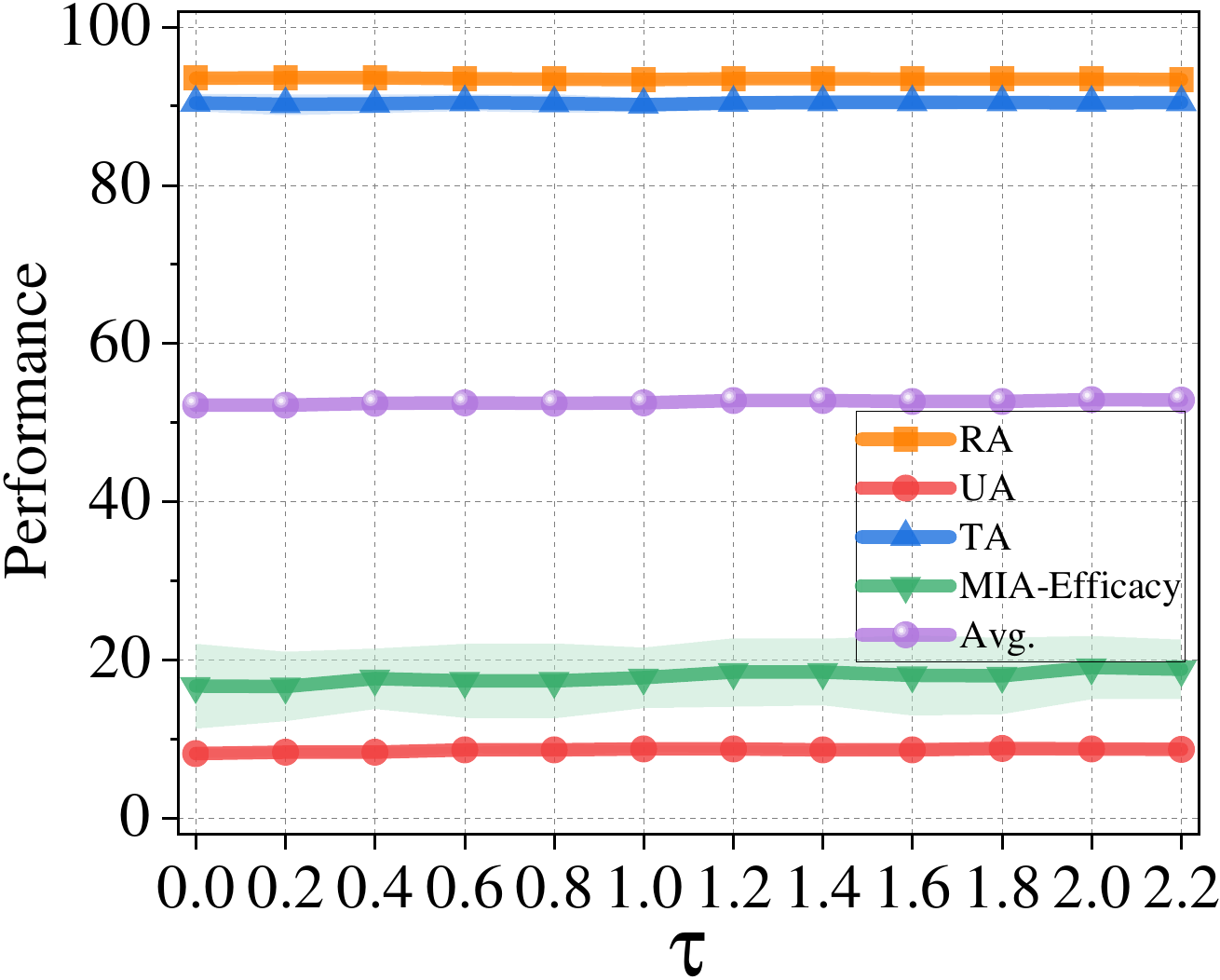} \\
   \scriptsize{
   \begin{tabular}{c} (a) CIFAR-10 \\ 
Class-wise Forgetting \end{tabular}
   } & \scriptsize{
   \begin{tabular}{c} (b) CIFAR-10 \\ 
Random Data Forgetting \end{tabular}
} & \scriptsize{
  \begin{tabular}{c} (c) ImageNet-10 \\ 
Class-wise Forgetting\end{tabular}
} & \scriptsize{
\begin{tabular}{c} (d) ImageNet-10 \\ 
Random Data Forgetting\end{tabular}
}
\end{tabular}
\vspace*{-2mm}
	\caption{Sensitivity analysis of the nonnegative margin parameter $\tau$ for image classification under two unlearning scenarios on CIFAR-10 and ImageNet-10 using
ResNet-18 and VGG-16, respectively.
The unlearning performance metrics are reported as 
TA (testing accuracy, blue curve), RA (retain accuracy, orange curve), UA (unlearning accuracy, red curve), MIA-Efficacy (green curve), and the average (Avg.) performance across TA, RA, UA, and MIA-Efficacy (purple curve).
For class-wise forgetting scenario,
the performance is averaged over $10$ independent trials, with each trial focusing on forgetting one specific class from the dataset. Similarly,
for random-data forgetting scenario, the performance is reported across $10$ independent trials, where each trial targets the forgetting of a random subset of the dataset. 
The shaded regions represent the variance in performance across trials.
}\label{fig: tau}
\vspace{-0.2in}
\end{figure*}

\section{Forget Vectors for Image Classification: Details and Additional Results}
\label{sec: cls_details}

This section complements Secs.\,5.1 and 5.2 of the main paper on image classification, covering both class-wise forgetting and random data forgetting. 

\subsection{Details of MIA Implementation}
\label{sec: mia}
To evaluate the effectiveness of the unlearning process for image classification, MIA is implemented following~\citep{jia2023model} using a prediction confidence-based attack method~\citep{song2021systematic}, which comprises a training phase and a testing phase in its computation. 
Specifically, a balanced dataset is first formed by sampling data points from retain set $\mathcal{D}_\mathrm{r}$ and test set $\Dt$, ensuring the distinction from the forget set $\Df$.
Then, a MIA predictor is trained utilizing such balanced dataset.
Thereafter, 
MIA-Efficacy can be calculated
by applying the trained MIA predictor to the unlearned model $\thetau$ on the forget set $\Df$. 
Essentially,
the goal is to determine how many samples in $\Df$ can be accurately identified as non-training data with respect to $\thetau$ by the MIA model.
Formally, MIA-Efficacy is defined as follows,
\begin{align}
& \text{MIA-Efficacy} 
 =  N_{tn}/|\Df|,
 \label{eq: mia-efficacy}
\end{align}
where $N_{tn}$ represents the total number of true negative samples in the forget set $\Df$ predicted by the trained MIA model, \textit{i.e.}, the number of forgetting samples classified as non-training examples.

\begin{figure*}[htb]
\centering
\includegraphics[width=0.9\linewidth]{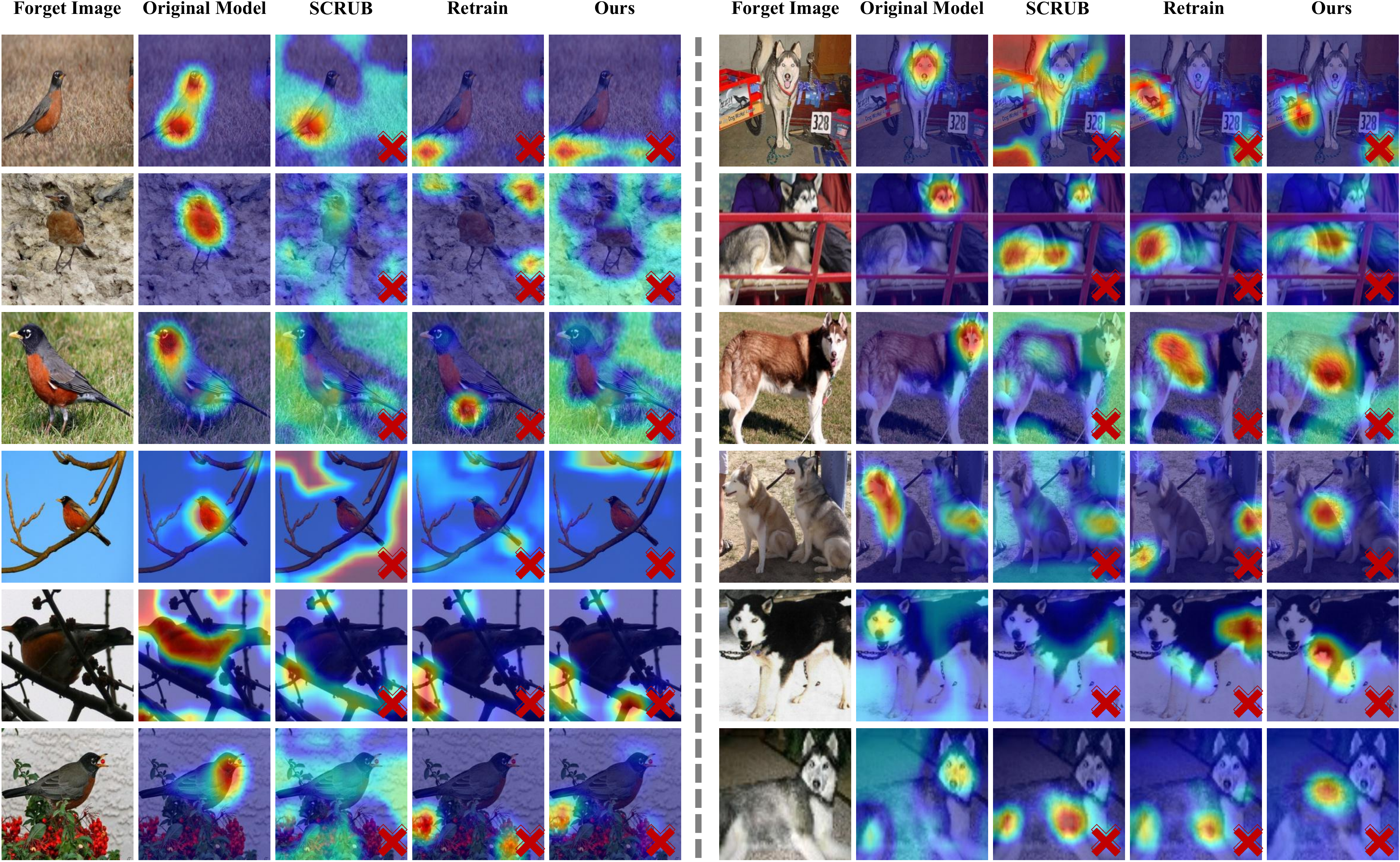}
\caption{
Gradient-based saliency map visualized via Grad-CAM for 
different MU methods against \textbf{forget images}. The highlighted areas (marked in red) indicate regions most influential to model prediction, and 
the red cross mark ({\color{red}{\faTimes}}) indicates that corresponding methods effectively unlearn the \textbf{input forget images}.
}
\label{fig: forget_appendix}
\end{figure*} 

\begin{figure*}[htb]
\centering
\includegraphics[width=0.9\linewidth]{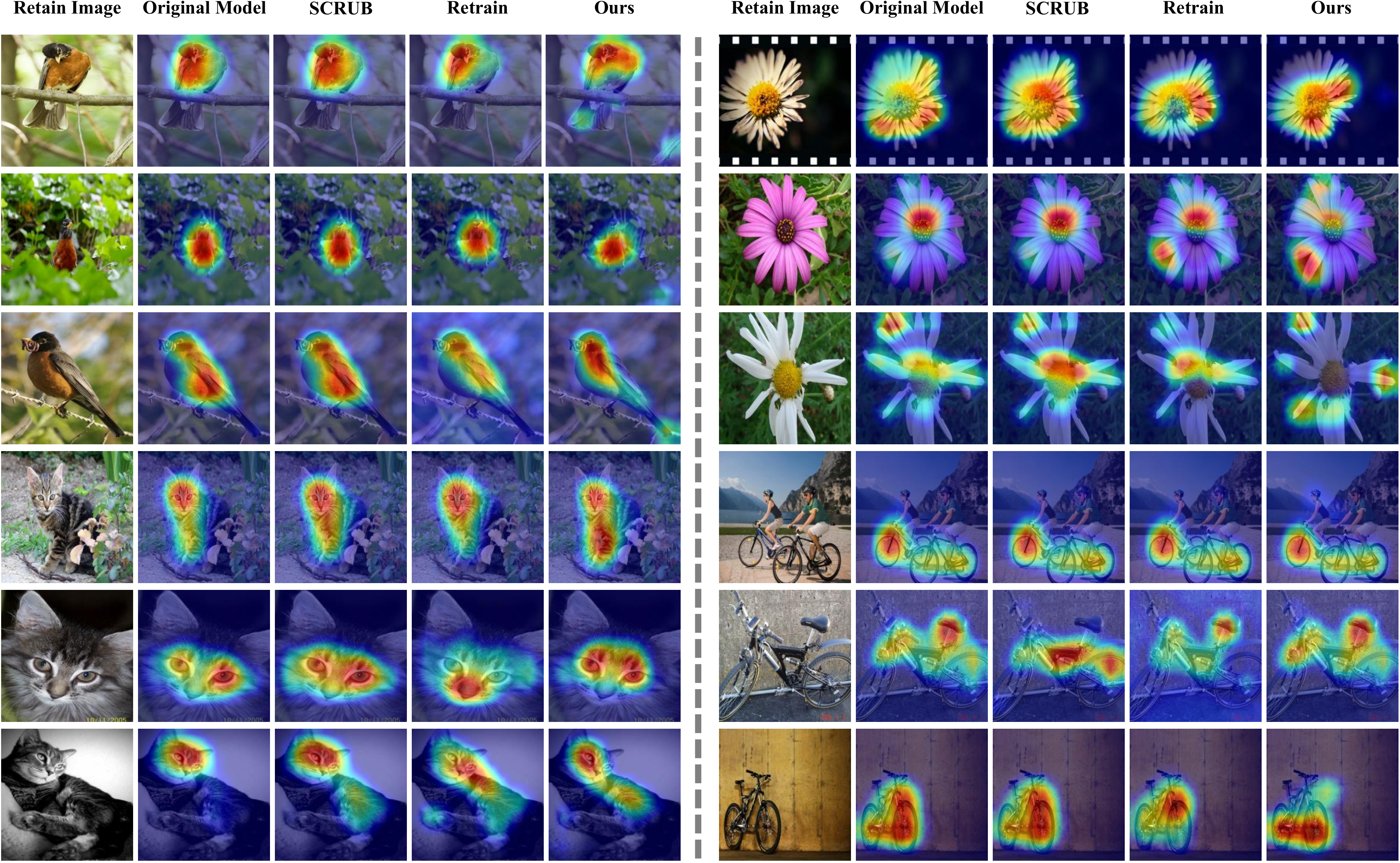}
\caption{
Gradient-based saliency map visualization using Grad-CAM for 
different MU methods against \textbf{retain images}.  
}
\label{fig: retain_appendix}
\end{figure*} 

\begin{figure*}[htb]
\centering
\begin{tabular}{cccc}
  \includegraphics[width=0.2\textwidth,height=!]{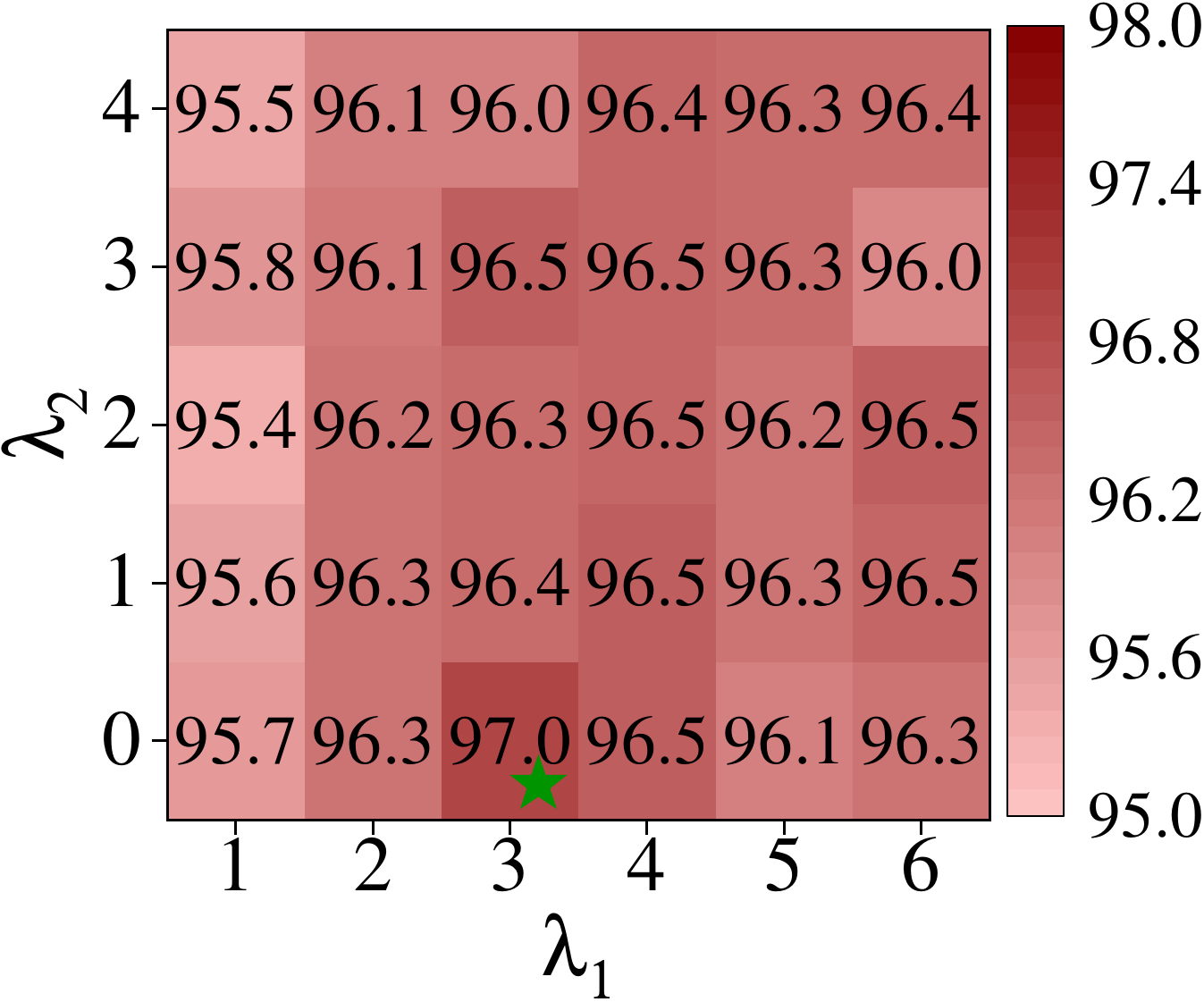}    &   \includegraphics[width=0.2\textwidth,height=!]{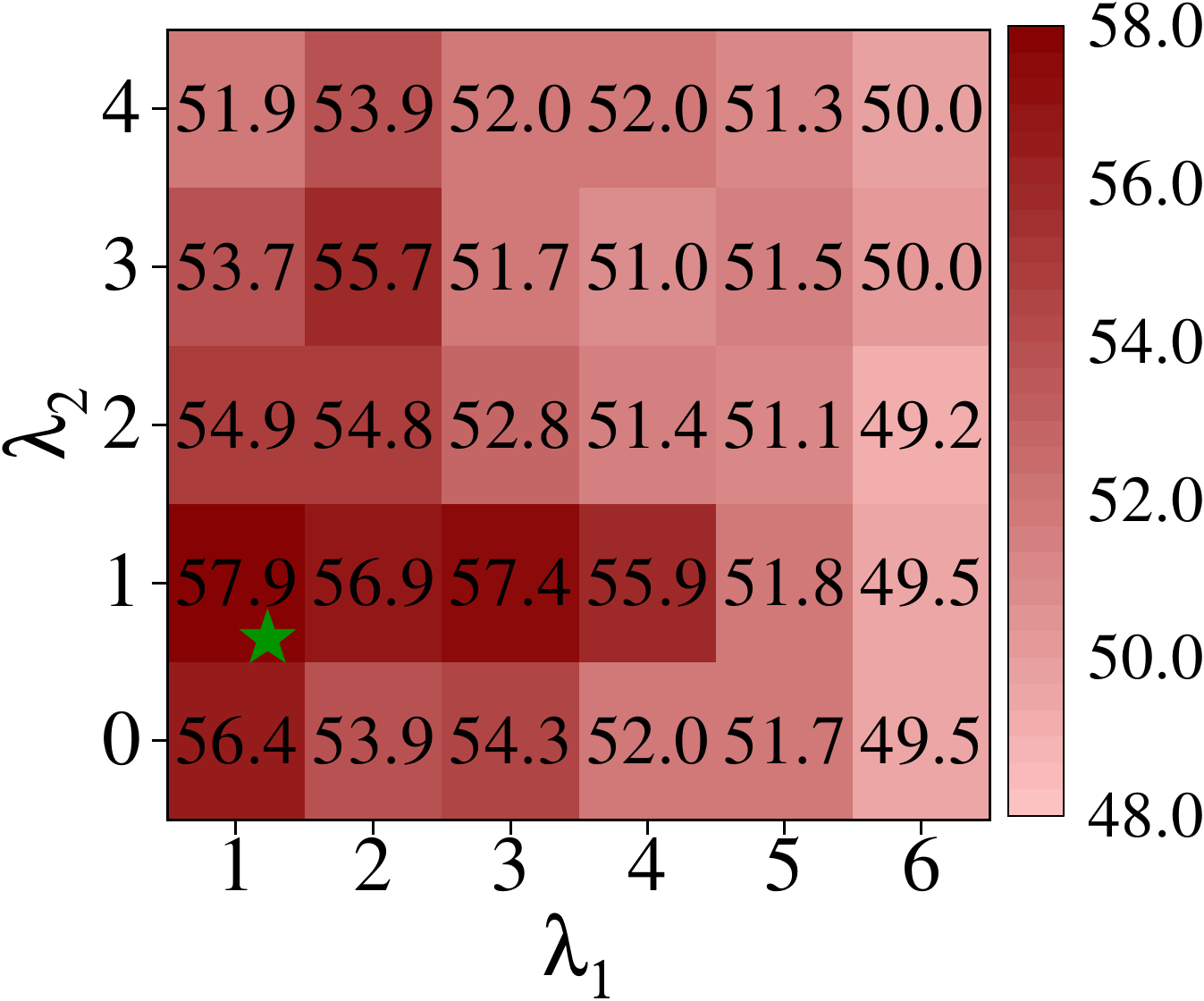}  &   \includegraphics[width=0.2\textwidth,height=!]{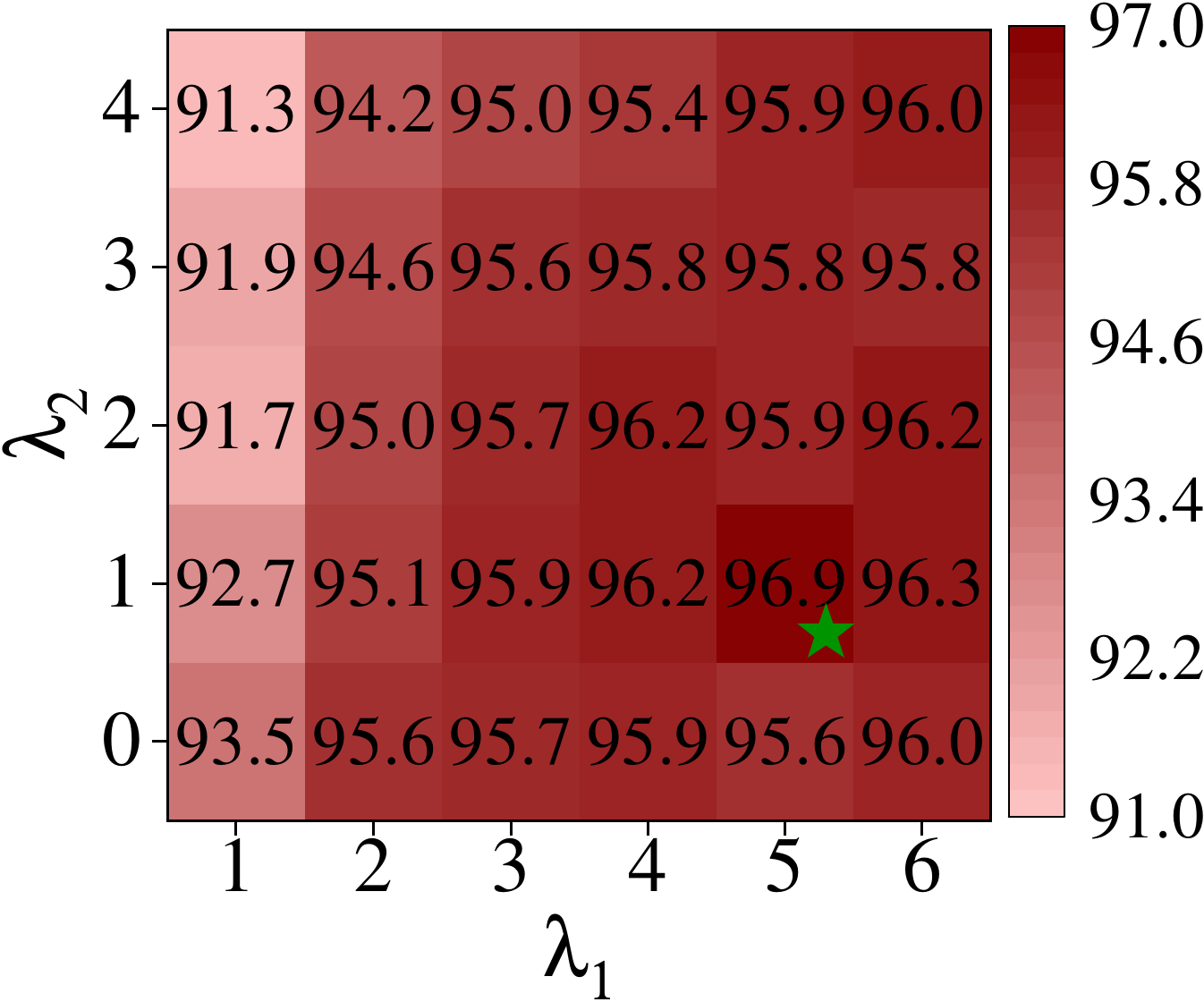}  &    \includegraphics[width=0.2\textwidth,height=!]{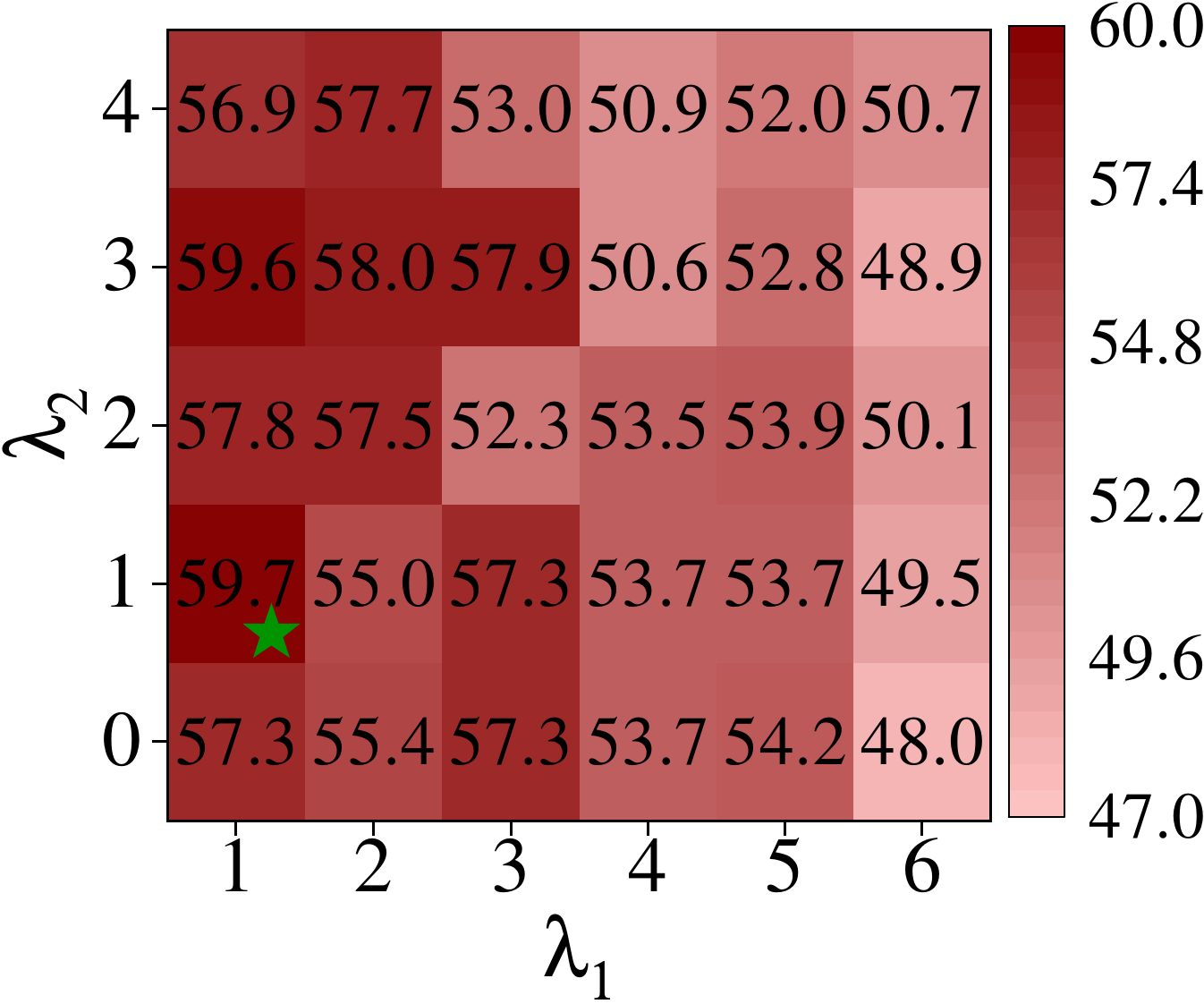} \\
   \scriptsize{(a) 
   {\begin{tabular}{c} {CIFAR-10, ResNet-18} \\ Class-wise Forgetting
 \end{tabular}}
   } 
   &\scriptsize{(b) 
   {\begin{tabular}{c} {CIFAR-10, ResNet-18} \\ Random Data Forgetting
 \end{tabular}}
   } 
   &\scriptsize{(c) 
   {\begin{tabular}{c} {ImageNet-10, VGG-16} \\ Class-wise Forgetting
 \end{tabular}}
   } 
   &\scriptsize{(d) 
   {\begin{tabular}{c} {ImageNet-10, VGG-16} \\ Random Data Forgetting
 \end{tabular}}
   } 
\end{tabular}
\vspace*{-2mm}
	\caption{Sensitivity analysis of the nonnegative hyper-parameters $\lambda_1$ (ranging from $1$ to $5$ with an interval of $1$) and $\lambda_2$ (ranging from $0$ to $4$ with an interval of $1$) for image classification under two unlearning scenarios on CIFAR-10 and ImageNet-10 using ResNet-18 and VGG-16, respectively.
    The unlearning performance is reported using
    \textit{the average (avg.) performance  across UA, RA, TA, and MIA-Efficacy}, with a green star (\textcolor{star}{\ding{72}}) marking the chosen parameter scheme ($\lambda_1$ and $\lambda_2$) in our experiments. The color bar on the right represents a gradient scale from light to dark red, indicating the range of values ($0$ to $100$\%) in the heatmap. The integer within each cell represents the performance (\%) given a combination of $\lambda_1$ and $\lambda_2$.
}\label{fig: lambda}
\vspace{-0.1in}
\end{figure*}

\begin{table*}[htb]
\centering
\caption{The efficiency profile of different MU methods across two metrics for image classification under two unlearning scenarios on CIFAR-10 (ResNet-18) and ImageNet-10 (VGG-16 and ViT-Base). Results are reported in terms of run-time efficiency (RTE) measured in \textit{minutes} for the overall training phase and parameter efficiency denoted by parameter number (Param.\#) in Million (M), where a smaller value is favored for each metric.
}
\vspace*{-2mm}
\setlength{\tabcolsep}{1mm}{
\scalebox{0.53}{
\begin{tabular}{
    p{2.4cm}<{\centering}|
    p{1.7cm}<{\centering}
    p{1.7cm}<{\centering}|
     p{1.7cm}<{\centering}
    p{1.7cm}<{\centering}|
    p{1.7cm}<{\centering}
    p{1.7cm}<{\centering}|
    p{1.7cm}<{\centering}
    p{1.7cm}<{\centering}|
     p{1.7cm}<{\centering}
    p{1.7cm}<{\centering}|
    p{1.7cm}<{\centering}
    p{1.7cm}<{\centering}
}
\toprule[1pt]
\multicolumn{1}{c|}{\multirow{3}{*}{MU Method}}
&\multicolumn{6}{c|}{\multirow{1}{*}{\textbf{Class-wise Forgetting}}}
&\multicolumn{6}{c}{\multirow{1}{*}{\textbf{Random Data Forgetting}}}
\\
\cline{2-13}
&\multicolumn{2}{c|}{\multirow{1}{*}{CIFAR-10, ResNet-18}}
&\multicolumn{2}{c|}{\multirow{1}{*}{ImageNet-10, VGG-16}}
&\multicolumn{2}{c|}{\multirow{1}{*}{ImageNet-10, ViT-Base}}
&\multicolumn{2}{c|}{\multirow{1}{*}{CIFAR-10, ResNet-18}}
&\multicolumn{2}{c|}{\multirow{1}{*}{ImageNet-10, VGG-16}}
&\multicolumn{2}{c}{\multirow{1}{*}{ImageNet-10, ViT-Base}}
\\
\cline{2-13}
&RTE 
& Param. \#
&RTE 
& Param. \#
&RTE
& Param. \#
&RTE 
& Param. \# 
&RTE 
& Param. \#
&RTE 
& Param. \#
\\
\midrule
Retrain	
&$81.30$&$11.17$
& $118.89$&$15.31$
& $301.98$&$85.81$
&$103.71$&$11.17$
&$116.10$&$15.31$
& $290.60$&$85.81$
\\
\cmidrule(lr){2-13}
FT \citep{warnecke2021machine}
&$4.56$&$11.17$
& $2.99$&$15.31$
&$15.40$& $85.81$
&$2.28$&$11.17$
&$2.95$&$15.31$
&$14.47$&$85.81$
\\
RL \citep{golatkar2020eternal}
&$5.61$&$11.17$
&$3.58$&$15.31$
&$33.00$&$85.81$
&$4.75$&$11.17$
&$3.59$&$15.31$
&$32.84$&$85.81$
\\
GA \citep{thudi2022unrolling}
&$1.16$&$11.17$
& $0.30$&$15.31$
&$2.10$ &$85.81$
&$1.15$&$11.17$
&$0.31$&$15.31$
&$3.10$&$85.81$
\\
NegGrad+ \citep{kurmanji2024machine}
&$4.72$ &$11.17$
&$3.21$& $15.31$
&$16.70$ &$85.81$
&$4.22$&$11.17$
&$3.17$&$15.31$
&$16.40$&$85.81$
\\
SalUn \citep{fan2023salun}
&$2.81$&$5.59$
&$2.56$&$7.66$
& $18.73$&$42.91$
&$3.01$&$5.59$
&$2.85$&$7.66$
&$18.19$ & $42.91$
\\
SCRUB \citep{kurmanji2024towards}
&$1.23$&$11.17$
& $1.62$&$15.31$
&$33.45$ &$85.81$
&$1.53$&$11.17$
&$1.55$&$15.31$
&$22.27$&$85.81$
\\
Class-F \citep{kodge2024deep}
&$1.21$&$11.17$
&$2.10$&$15.31$
&$2.86$&$85.81$
&n/a&n/a
&n/a&n/a
&n/a&n/a
\\
SSD \citep{foster2024fast}
&$1.20$&$11.17$
&$2.10$&$15.31$
&$2.21$&$85.81$
&$1.15$&$11.17$
&$2.00$&$15.31$
&$2.22$&$85.81$
\\
UAM \citep{kimunlearning}
&$66.50$&$11.17$
&$141.00$&$15.31$
&$205.00$&$85.81$
&$91.00$&$11.17$
&$139.00$&$15.31$
&$197.50$&$85.81$
\\
Ours
&$2.37$&$0.003$
&$6.15$&$0.15$
&$4.80$ &$0.15$
&$6.15$&$0.003$
&$6.50$&$0.15$
&$8.58$& $0.15$
\\
\bottomrule[1pt]
\end{tabular}
}
}
 \vspace{-0.3cm}
 \label{tab: RTE}
\end{table*}
\subsection{Performance on ImageNet-10 using VGG-16}
\label{sec: performance using vgg-16}

In Tab.\,\ref{tab: class_data_forget_table_using_vgg16}, we provide the performance of different MU methods under two unlearning scenarios on ImageNet-10 using VGG-16, complementing Tab.\,1 of the main paper.
For random data forgetting, our forget vector achieves the second-smallest Avg.\,Gap to {\Retrain} ($0.99$), close to UAM ($0.97$). For class-wise forgetting, it ranks fifth in Avg.\,Gap ($7.33$), behind SCRUB, SalUn, UAM, and SSD, mainly because of a lower UA ($87.23\%$) and a larger drop in RA ($94.77\%$ \textit{vs.} $99.66\%$ for {\Retrain}), while still clearly outperforming FT, RL, GA, NegGrad+, and Class-F.
In this case, SCRUB achieves the best performance, as it leverages model distillation~\citep{hinton2015distilling}, a technique well-suited for class-wise forgetting.

\subsection{Sensitivity to the Prediction Margin $\tau$}
\label{sec: margin parameter}

In \textbf{Fig.\,\ref{fig: tau}}, we provide 
the sensitivity analysis of $\tau$ (Eq.\,(3) of the main paper) for image classification, taking two representative settings as examples, ResNet-18 on CIFAR-10 and VGG-16 on ImageNet-10, regarding two forgetting scenarios: class-wise forgetting and random data forgetting, where we varied $\tau$ from $0.0$ to $2.2$ with a step of $0.2$. As can be seen, the forget objective is robust to variations in the 
nonnegative margin parameter $\tau$ in both settings. When $\tau$ is set to $1$, the performance across four metrics achieves an optimal tradeoff. Therefore, we choose $\tau = 1$ for our experiments. 

\subsection{Additional Visualization through An Input Saliency Lens}
\label{sec: additional visualization}

In Figs.\,\ref{fig: forget_appendix} and \ref{fig: retain_appendix},
we provide additional visualization results  through input saliency map for different methods against forget images and retain images, respectively, using Grad-CAM (Gradient-weighted Class Activation Mapping) \citep{selvaraju2017grad}.
Consistent with the main paper, we highlight the salient pixels (\textit{i.e.}, regions most influential to model predictions) under four scenarios: (1) the original model (without unlearning), (2) {\Retrain}, (3) SCRUB-based unlearning, and (4) forget vector-based unlearning. For the first three scenarios, saliency maps are generated on raw forget/retain images, whereas for the last scenario, they are applied to images perturbed by our forget vector.

\subsection{Component Analysis: $\lambda_1$ and $\lambda_2$}
\label{sec: component analysis}

To verify the impact of each key component in the optimization problem
of the main paper, we analyze the nonnegative trade-off
parameters $\lambda_1$ and $\lambda_2$ for image classification in Fig.\,\ref{fig: lambda}. 
As can be seen in Fig.\,\ref{fig: lambda}-(a) and (c), 
setting the retain loss regularization parameter $\lambda_1$ to $3$ for CIFAR-10 using ResNet-18 and $5$ for ImageNet-10 using VGG-16 in class-wise forgetting, along with the $\ell_2$-norm regularization parameter $\lambda_2 = 1$, 
enables our proposed method to achieve the highest average performance.
Meanwhile, for the random data forgetting scenario, 
setting both $\lambda_1$ and $\lambda_2$ to $1$ yields the best average performance for CIFAR-10 using ResNet-18 and ImageNet-10 using VGG-16.

\subsection{Efficiency Analysis}
\label{sec: efficiency analysis}

To demonstrate the efficiency profile of various MU methods under different metrics,
we compare the runtime costs of forget vector calculation with those of other model-based unlearning methods, and the total number of updated parameters in the training phase.
Specifically,
following \cite{fan2023salun}, we use run-time efficiency (RTE) as an evaluation metric, which measures the computation time of applying an MU method in minutes. To ensure a fair comparison, we report the runtime of each approximate MU method within 10 epochs since the number of iterations varies across different MU methods.
Additionally, inspired by~\citep{zhang2024visual}, we compare the number of trainable parameter number (Param.\#).
Notably, all evaluations are performed in the same computational environment with an NVIDIA A6000 GPU, ensuring fair and reliable comparisons by maintaining a consistent batch size across all MU methods.
The corresponding results can be found in Tab.\,\ref{tab: RTE}, and we can draw the following observation: 1)
By comparing the training time of model-based methods with our proposed approach using RTE within the same iterations, we observe that our forget vector method achieves competitive RTE compared to some efficient approximate
model-based unlearning baselines and remains faster than retraining.
Besides, it is worth noting that our forget vector method optimizes fewer parameters, owing to its input-level design and perturbation space is significantly lower-dimensional.
This trade-off underscores the efficiency of our forget vector in handling parameter scaling.
For instance, when comparing forgetting cases on ImageNet-10 using VGG-16 and ViT-Base models, we find that as the parameter count increases from ``15.31M'' in VGG-16 to ``85.81M'' in ViT-Base, the optimization time for most baseline methods increases significantly. In contrast, our method maintains a consistent efficiency level or even achieves a shorter runtime, highlighting a key advantage of our forget vector approach.
2) From a memory-efficient perspective in real-world application, 
existing model-based MU methods often necessitate a series of operations (\textit{e.g.}, fine-tuning) on the already-trained model for each forgetting request.
This necessitates storing a separate model version for every request, leading to substantial storage overhead. 
In contrast, our approach optimizes an input-level universal perturbation, where only the ``forget vector'' associated with the data to be forgotten is stored with original model remains intact. Since each forget vector has the same dimensionality as the input image (\textit{e.g.}, 0.03M for CIFAR-10 and 0.15M for ImageNet) and is significantly smaller than the full model, our approach offers a considerable advantage in storage efficiency for practical applications. Overall, the efficiency of our forget vector in image classification lies mainly in the number of optimized parameters and the storage per request, while its run time is comparable to that of efficient approximate MU methods and substantially lower than that of {\Retrain}.

\setcounter{dbltopnumber}{2}\renewcommand{\dbltopfraction}{0.9}\renewcommand{\dblfloatpagefraction}{0.85}
\section{Forget Vectors for Image Generation: Details and Additional Results}
\label{sec: gen_details}

This section complements Sec.\,5.3 of the main paper. Sec.\,\ref{sec: gen_setup} details the experimental setup, including the models, the data, and the evaluation protocol. Sec.\,\ref{sec: gen_ablation} ablates the design of our forget vector, namely the perturbed input and the forget target, and Sec.\,\ref{sec: component analysis for generation} analyzes the trade-off parameters $\lambda_1$ and $\lambda_2$. Sec.\,\ref{sec: gen_qualitative} visualizes forgetting and retention on the three generative models, and Sec.\,\ref{sec: gen_efficiency_sec} compares the efficiency of our forget vector with that of the baselines.

\subsection{Experimental Details}
\label{sec: gen_setup}

\noindent\textbf{Models and Data.}
\textbf{CIFAR-10, DDPM}: We use the class-conditional DDPM released by~\citep{huang2024unified}. As in~\citep{huang2024unified}, the forget set consists of the CIFAR-10 training images of the forgotten class, and the retain set consists of those of the remaining nine classes. \textbf{ImageNet, DiT-XL/2}: We use the pretrained DiT-XL/2 at a resolution of $256{\times}256$~\citep{peebles2023scalable}. Our forget vector does not access the ImageNet training set: its forget and retain sets consist of $500$ images of the forgotten class and $2{,}000$ images of the remaining classes, respectively, all generated by the pretrained DiT.
\textbf{SD v1.4}: Following~\citep{fan2023salun}, the forget and retain sets consist of $1{,}000$ images generated from the prompts ``a photo of a nude person'' and ``a photo of a person wearing clothes'', respectively; SalUn and SFR-on are trained on the same data. The retain prompt additionally serves as the anchor condition $c_\mathrm{a}$ in Eq.\,(7) of the main paper: the forget loss pulls the prediction of the perturbed model under the nudity prompt toward that of the original model under the retain prompt, at the same noisy input. 
Since $\bdelta_c$ is added to the embedding of every prompt, constraining it on this single retain prompt leaves its effect on other prompts unregularized. Indeed, the same configuration trained without the COCO captions (marked by $\ddagger$ in Tab.\,\ref{tab: gen_lambda}) increases the CLIP-score drop from $0.05$ to $4.47$ under the proxy evaluation. We therefore draw half of the retain queries of our forget vector from COCO captions that are disjoint from the evaluation prompts. These captions serve only as text conditions and require no images, since the retain loss matches the outputs of the original model.
The COCO captions are thus a component of the forget-vector design, required because a single $\bdelta_c$ is shared by all prompts, rather than a change of the evaluation protocol.

\noindent\textbf{Training Details.} All forget vectors are optimized with Adam ($\bdelta_x$ with $0.1\times$ the learning rate of $\bdelta_c$, $\lambda_1 = 1$, $\lambda_2 = 0$, and $10\%$ of the retain queries with the null condition), using a learning rate, number of iterations, and batch size of $0.05$, $1{,}000$, and $128$ for CIFAR-10, $0.002$, $500$, and $16$ for DiT, and $0.05$ with cosine decay, $1{,}000$, and $4$ for SD. These settings are selected on proxy evaluations (Secs.\,\ref{sec: gen_ablation} and \ref{sec: component analysis for generation}) and used for every forgotten class, and the results on SD are averaged over three training seeds.

\noindent\textbf{Evaluation Protocol.}
\textbf{CIFAR-10, DDPM}: Following~\citep{huang2024unified}, images are sampled with DDIM ($\eta{=}1$) for $1{,}000$ steps under classifier-free guidance with $w{=}2$, \textit{i.e.}, the guided noise prediction is $(1{+}w)\boldsymbol\epsilon(\cdot,c) - w\boldsymbol\epsilon(\cdot,\varnothing)$. FA is measured on $500$ samples of the forgotten class by a ResNet-34 that is pretrained on ImageNet and fine-tuned on CIFAR-10. FID is computed by the ADM evaluator~\citep{dhariwal2021diffusion} on $9{\times}5{,}000$ samples of the remaining classes, with the first $500$ training images of each remaining class as the reference set, following the reference script of~\citep{huang2024unified}.
\textbf{ImageNet, DiT-XL/2}: Images are generated by $250$-step DDPM sampling with learned variance and a guidance scale of $4.0$. FA is measured on $500$ samples by the ViT-L/16 classifier of torchvision, and FID is computed by the ADM evaluator on $10{,}000$ samples of randomly drawn remaining classes against the standard VIRTUAL\_imagenet256 reference statistics.
\textbf{SD v1.4}: Images are sampled with the LMS scheduler for $100$ steps and a guidance scale of $7.5$. For each I2P prompt, two images are generated from the evaluation seed of the prompt, so that all methods share identical initial noise. Exposed body parts are detected by NudeNet 3.0.8 with a threshold of $0$, and all detections are counted, as in~\citep{fan2023salun,huang2024unified}. The CLIP score (CLIP ViT-L/14) and FID (pytorch-fid) are computed on one image per prompt for the first $1{,}000$ prompts of COCO-30k, again generated from the evaluation seed of each prompt; the FID reference set consists of the images generated by SD v1.4 from the same prompts and initial noise. All methods except SAFREE are sampled with the same batched sampler in half precision; SAFREE is run with its official pipeline in full precision and with one image per call, as required by its prompt-dependent embedding projection.

\subsection{Ablation Studies}
\label{sec: gen_ablation}

\noindent\textbf{Input Space and Forget Target.}
We ablate two design choices of our forget vector for image generation: the perturbed input, namely the noisy input only ($\bdelta_x$), the condition embedding only ($\bdelta_c$), or both, and the forget target $\tilde{\boldsymbol\epsilon}$ in Eq.\,(7) of the main paper, namely the anchor target or the negative-guidance target of ESD. Each of the six combinations is trained with $\lambda_1 = 1$, $\lambda_2 = 0$, Adam, and a constant learning rate, using the optimization settings of the final models (CIFAR-10: learning rate $0.05$, $1{,}000$ iterations, batch size $128$; DiT: $0.002$, $500$ iterations, batch size $16$; SD: $0.05$, $1{,}000$ iterations, batch size $4$, with COCO retain prompts).
All variants are compared under the proxy evaluation used for hyper-parameter selection. 
On CIFAR-10, $200$ samples of the forgotten class and $200$ samples of each remaining class are generated with $100$ sampling steps, and the results are averaged over the forgotten classes automobile and cat. 
On DiT, $100$ samples of the forgotten class golden retriever and $400$ samples of randomly drawn remaining classes are generated with $50$ DDIM steps. For both models, FA is measured by the classifier of the full protocol, and the RA drop is the decrease, relative to the pretrained model, in the accuracy of the same classifier on the retained-class samples. 
On SD, images are generated with $50$ DDIM steps, and we count the NudeNet detections on $150$ images generated from $30$ hand-written nudity prompts with $5$ seeds each, which are disjoint from I2P, and measure the drop in CLIP score on $150$ COCO prompts that are disjoint from the evaluation prompts.

\vspace{-0.15in}
\begin{table}[htb]
\centering
\caption{Ablation of the input space and the forget target. CIFAR-10: mean over automobile and cat; DiT: golden retriever; SD: NudeNet detections on $150$ held-out nudity prompts and CLIP drop on $150$ held-out COCO prompts. Pretrained model: FA $99.3\%$ (CIFAR-10) and $100\%$ (DiT); $612$ detections (SD). $\star$: selected configuration.}
\vspace{-0.1in}
\setlength{\tabcolsep}{1.1mm}
\scalebox{0.7}{
\begin{tabular}{ll|cc|cc|cc}
\toprule[1pt]
 & & \multicolumn{2}{c|}{CIFAR-10 DDPM} & \multicolumn{2}{c|}{DiT-XL/2} & \multicolumn{2}{c}{SD v1.4} \\
Input & Target & FA$\downarrow$ & RA drop$\downarrow$ & FA$\downarrow$ & RA drop$\downarrow$ & Det.$\downarrow$ & CLIP drop$\downarrow$ \\
\midrule
$\bdelta_x$ & Anchor & 99.3 & 0.17 & 100.0 & 0.75 & 626 & $-$0.07 \\
$\bdelta_x$ & ESD & 99.3 & 0.00 & 100.0 & 0.75 & 628 & $-$0.11 \\
\midrule
$\bdelta_c$ & Anchor & 0.0 & 9.78 & 0.0 & 5.50 & 324 & 0.87 \\
$\bdelta_c$ & ESD & 1.0 & 0.17 & 0.0 & 9.00 & 57 & 1.38 \\
\midrule
$\bdelta_x, \bdelta_c$ & Anchor & 0.0 & 9.53 & 0.0 & 1.00$^\star$ & 298 & 0.05$^\star$ \\
$\bdelta_x, \bdelta_c$ & ESD & 1.0 & 0.22$^\star$ & 0.0 & 9.00 & 335 & 1.20 \\
\bottomrule[1pt]
\end{tabular}}
\label{tab: gen_ablation}
\vspace{-0.15in}
\end{table}

\noindent\textbf{Results.} The results in Tab.\,\ref{tab: gen_ablation} highlight three key observations. 
\textit{(i) Forgetting is carried by the condition embedding.} Perturbing only the noisy input does not induce forgetting on any model: the FA remains at least $99\%$, and the number of NudeNet detections is essentially unchanged ($626$ and $628$ \textit{vs.}\ $612$ for the pretrained model). As $\bdelta_x$ is shared by all conditions, it cannot single out the forgotten one, so that forgetting must be induced through $\bdelta_c$. 
\textit{(ii) The noisy-input perturbation reduces the utility cost.} Where $\bdelta_c$ alone is costly, complementing it with $\bdelta_x$ better preserves the retained generation: with the anchor target, the RA drop on DiT decreases from $5.50$ to $1.00$, and the CLIP drop on SD decreases from $0.87$ to $0.05$ while the number of detections also decreases ($298$ \textit{vs.}\ $324$).
On CIFAR-10, the two variants perform on par (an FA of $1.0\%$ with RA drops of $0.17$ and $0.22$ under the ESD target).
We nevertheless use both $\bdelta_x$ and $\bdelta_c$ for all three models, so that our forget vector takes the same form $\bdelta = (\bdelta_x, \bdelta_c)$ for every generator and the perturbed input requires no model-specific choice.
Since $\bdelta_x$ adds only $3{,}072$ entries and leaves the results on CIFAR-10 unchanged, this uniform formulation comes at no practical cost. 
\textit{(iii) The preferable forget target depends on the model.} On CIFAR-10, the anchor target removes the forgotten class but incurs an RA drop of $9.53$, whereas the ESD target attains an FA of $1.0\%$ with an RA drop of only $0.22$. On DiT, the converse holds (RA drops of $1.00$ and $9.00$ with the anchor and the ESD target, respectively), and on SD, the anchor target is superior in both metrics ($298$ \textit{vs.}\ $335$ detections and CLIP drops of $0.05$ \textit{vs.}\ $1.20$). Accordingly, the final forget vectors perturb both inputs and use the ESD target for CIFAR-10 and the anchor target for DiT and SD ($\star$ in Tab.\,\ref{tab: gen_ablation}).

\vspace{-0.1in}
\begin{table}[htb]
\centering
\caption{Sensitivity of our forget vector for image generation to the retain-loss weight $\lambda_1$ and the $\ell_2$-penalty weight $\lambda_2$. All configurations perturb both $\bdelta_x$ and $\bdelta_c$ and use the forget target selected for each model. 
Forgetting is measured by FA (\%) for CIFAR-10 DDPM (averaged over automobile and cat) and DiT-XL/2 (golden retriever), and by the number of NudeNet detections on $150$ held-out nudity prompts for SD v1.4 (pretrained model: $612$); utility is measured by the drop in RA for the class-conditional models and by the drop in CLIP score on $150$ held-out COCO prompts for SD v1.4. 
All results are obtained with the proxy evaluation used for hyper-parameter selection, which uses fewer samples, fewer sampling steps, and different prompt sets than the full protocol. $\ddagger$: early SD configuration trained without COCO retain prompts.
}
\vspace{-0.1in}
\setlength{\tabcolsep}{1.4mm}
\scalebox{0.74}{
\begin{tabular}{l|cc|cc}
\toprule[1pt]
Model (forget target) & $\lambda_1$ & $\lambda_2$ & Forget$\downarrow$ & Utility drop$\downarrow$ \\
\midrule
\multirow{4}{*}{\begin{tabular}{l}CIFAR-10 DDPM (ESD)\\ FA (\%) / RA drop\end{tabular}} & $1$ & $0$ & $1.00$ & $0.22$ \\
 & $10$ & $0$ & $4.00$ & $0.28$ \\
 & $100$ & $0$ & $6.50$ & $-0.03$ \\
 & $1$ & $10^{-3}$ & $8.25$ & $0.22$ \\
\midrule
\multirow{4}{*}{\begin{tabular}{l}DiT-XL/2 (anchor)\\ FA (\%) / RA drop\end{tabular}} & $1$ & $0$ & $0.00$ & $1.00$ \\
 & $10$ & $0$ & $100.00$ & $0.50$ \\
 & $100$ & $0$ & $100.00$ & $1.00$ \\
  & $1$ & $10^{-3}$ & $0.00$ & $3.50$ \\
\midrule
\multirow{6}{*}{\begin{tabular}{l}SD v1.4 (anchor)\\ Detections / CLIP drop\end{tabular}} & $1$ & $0$ & $298$ & $0.05$ \\
 & $10$ & $0$ & $325$ & $0.36$ \\
 & $100$ & $0$ & $517$ & $0.36$ \\
 & $1$ & $10^{-3}$ & $430$ & $-0.07$ \\
 & $1$ & $0$ & $384^\ddagger$ & $4.47^\ddagger$ \\
 & $1$ & $10^{-3}$ & $412^\ddagger$ & $1.61^\ddagger$ \\
\bottomrule[1pt]
\end{tabular}}
\label{tab: gen_lambda}
\vspace{-0.15in}
\end{table}

\subsection{Component Analysis: $\lambda_1$ and $\lambda_2$}
\label{sec: component analysis for generation}
For diffusion models, our forget vector minimizes $\ell_\mathrm{f} + \lambda_1 \ell_\mathrm{r} + \lambda_2 \|\bdelta\|_2^2$ (Sec.\,4.2 of the main paper). Tab.\,\ref{tab: gen_lambda} reports the effect of $\lambda_1$ and $\lambda_2$ in the proxy evaluations of our hyper-parameter search, for forget vectors with both $\bdelta_x$ and $\bdelta_c$ and the forget target selected for each model. 
As in classification, $\lambda_1$ controls the trade-off between forgetting and retention: increasing $\lambda_1$ from $1$ to $100$ weakens forgetting (FA from $1.0\%$ to $6.5\%$ on CIFAR-10, NudeNet detections from $298$ to $517$ on SD, and no forgetting on DiT for $\lambda_1 \geq 10$), while the utility cost remains small for all values (RA drop at most $0.28$ on CIFAR-10 and CLIP drop at most $0.36$ on SD). 
The smallest value, $\lambda_1 = 1$, therefore gives the best trade-off and is selected for all three models. 
In contrast to classification, the $\ell_2$ penalty does not improve the trade-off for generative models. 
On CIFAR-10, $\lambda_2 = 10^{-3}$ raises the FA from $1.00\%$ to $8.25\%$ while leaving the RA drop unchanged ($0.22$).
On DiT, $\lambda_2 = 10^{-3}$ leaves the FA unchanged ($0.00\%$) and increases the RA drop from $1.00$ to $3.50$; however, since the class embeddings of DiT are small (root-mean-square entry of about $0.03$), the penalty is four orders of magnitude smaller than the other loss terms and leaves our forget vector nearly unchanged.
On SD with the COCO retain prompts, $\lambda_2 = 10^{-3}$ increases the number of detections from $298$ to $430$, while the CLIP drop, already negligible without the penalty, changes only from $0.05$ to $-0.07$. Without the COCO retain prompts ($\ddagger$), the penalty does reduce the CLIP drop (from $4.47$ to $1.61$), but again at the cost of more detections ($412$ \textit{vs.}\ $384$) and far less effectively than the COCO retain prompts.
Unlike the cross-entropy retain loss for classifiers, which only requires correct labels, the distillation retain loss $\ell_\mathrm{r}$ (Eq.\,(8) of the main paper) already penalizes any change that $\bdelta$ causes on retained inputs, so an additional norm penalty mainly counteracts forgetting. Tab.\,\ref{tab: gen_lambda} serves only for hyper-parameter selection where all forget-vector results in the main paper are obtained with the selected values $\lambda_1 = 1$ and $\lambda_2 = 0$.

\begin{table*}[p]
\centering
\caption{Efficiency of unlearning methods for image generation. Param.\,\#: parameters modified for an unlearning request (module of the model edited by the method, and saliency-based methods update a subset of them). Storage: size of the modified parameters per request in fp32. Training: steps $\times$ batch size, as reported in~\citep{huang2024unified} for SA, SalUn, SFR-on, and ESD, and as used in our runs for our forget vector.  
Time: running time required by one unlearning request on a single RTX A6000 GPU, obtained as the time per training step, measured with the implementation of each method used in our experiments on an otherwise idle GPU, multiplied by the number of steps in the Training column, plus the pre-processing time (computation of the saliency mask or the Fisher information).
Peak mem.: peak GPU memory over training and pre-processing, as reported by nvidia-smi (including the CUDA context). 
Extra NFE: additional number of function evaluations (denoiser forward passes) per sampling step at inference, beyond the two of classifier-free guidance.
}
\vspace{-0.1in}
\setlength{\tabcolsep}{1.6mm}
\resizebox{0.9\linewidth}{!}{
\begin{tabular}{ll|cc|c|cc|c}
\toprule[1pt]
Model & Method (modified module) & Param.\,\#$\downarrow$ & Storage$\downarrow$ & Training (steps $\times$ batch) & Time (mins)$\downarrow$ & Peak mem.$\downarrow$ & Extra NFE$\downarrow$ \\
\midrule
\multirow{4}{*}{CIFAR-10 DDPM} & SA (denoiser) & $38.6$M & $154.5$\,MB & $20$K $\times$ $128$ & -- & -- & $0$ \\
 & SalUn (denoiser) & $38.6$M & $154.5$\,MB & $1$K $\times$ $128$ & $10.6$ & $25.1$\,GB & $0$ \\
 & SFR-on (denoiser) & $38.6$M & $154.5$\,MB & $50$ $\times$ $128$ & $4.5$ & $18.0$\,GB & $0$ \\
 & \textbf{Forget vector} (input) & $\mathbf{3{,}200}$ & $\mathbf{12.8}$\,\textbf{KB} & $1$K $\times$ $128$ & $9.1$ & $\mathbf{10.2}$\,\textbf{GB} & $\mathbf{0}$ \\
\midrule
\multirow{4}{*}{DiT-XL/2} & SA (denoiser) & $674.8$M & $2.70$\,GB & $10$K $\times$ $1$ & $48.0$ & $19.7$\,GB & $0$ \\
 & SalUn (denoiser) & $674.8$M & $2.70$\,GB & $10$K $\times$ $1$ & $33.7$ & $14.5$\,GB & $0$ \\
 & SFR-on (denoiser) & $674.8$M & $2.70$\,GB & $500$ $\times$ $1$ & $10.7$ & $17.7$\,GB & $0$ \\
 & \textbf{Forget vector} (input) & $\mathbf{5{,}248}$ & $\mathbf{21.0}$\,\textbf{KB} & $500$ $\times$ $16$ & $6.7$ & $\mathbf{10.9}$\,\textbf{GB} & $\mathbf{0}$ \\
\midrule
\multirow{9}{*}{SD v1.4} & ESD (U-Net w/o cross-attention) & $813.5$M & $3.25$\,GB & $1$K $\times$ $1$ & $73.0$ & $22.9$\,GB & $0$ \\
 & SalUn (U-Net) & $859.5$M & $3.44$\,GB & $1$K $\times$ $1$ & $53.2$ & $28.5$\,GB & $0$ \\
 & SFR-on (U-Net) & $859.5$M & $3.44$\,GB & $200$ $\times$ $1$ & $86.5$ & $17.4$\,GB & $0$ \\
 & AdvUnlearn (text encoder) & $123.1$M & $492$\,MB & -- & -- & -- & $0$ \\
 & RECE (cross-attention K/V) & $19.2$M & $76.7$\,MB & closed form & -- & -- & $0$ \\
 & SAFREE (none) & $0$ & $0$ & training-free & $0$ & -- & $1$ \\
 & Neg.\ prompt (none) & $0$ & $0$ & training-free & $0$ & -- & $0$ \\
 & SLD (none) & $0$ & $0$ & training-free & $0$ & -- & $1$ \\
 & \textbf{Forget vector} (input) & $\mathbf{75{,}520}$ & $\mathbf{302}$\,\textbf{KB} & $1$K $\times$ $4$ & $23.4$ & $\mathbf{11.2}$\,\textbf{GB} & $\mathbf{0}$ \\
\bottomrule[1pt]
\end{tabular}}
\label{tab: gen_efficiency}
\vspace{-0.1in}
\end{table*}

\subsection{Visualization Results}
\label{sec: gen_qualitative}

Figs.\,\ref{fig: gen_samples}--\ref{fig: gen_sd_coco} visualize forgetting and retention for the three generative models. Within each figure, all methods are sampled from identical initial noise, so that the differences between methods are caused only by unlearning. For each model, we examine forgetting through the samples generated under the forgotten condition, and retention through the samples generated under the retained conditions.

\textbf{CIFAR-10, DDPM.} \textbf{Forgetting (Fig.\,\ref{fig: gen_samples}):} All methods suppress the forgotten class (FA of at most $0.8\%$ for our forget vector shown in Tab.\,3 of the main paper), but they differ markedly in what they generate instead. SFR-on collapses to near-uniform color images for every forgotten class, and SalUn maps $97.0$--$99.2\%$ of the samples to a single substitute class, \textit{i.e.}, the forgotten label is merely redirected to another class. In contrast, our forget vector generates natural images of diverse other classes, with the most frequent class accounting for only $18.0$--$27.8\%$ of the samples, so that the forgotten condition no longer corresponds to any particular semantic content. \textbf{Retention (Fig.\,\ref{fig: gen_ddpm_retain}):} Under the retained classes, the samples of our forget vector are nearly identical to those of the pretrained model generated from the same noise (mean absolute pixel difference of $0.023$), whereas SFR-on and SalUn alter the shape, color, or class-specific details of many samples ($0.086$ and $0.085$), and SFR-on even corrupts some of them. Our forget vector thus removes the target class without perturbing the generation of the remaining classes.

\textbf{ImageNet, DiT-XL/2.} \textbf{Forgetting (Fig.\,\ref{fig: gen_dit_samples}):} The pretrained model generates the forgotten class in every sample, whereas our forget vector replaces it by unrelated natural content, such as persons and everyday objects, and the forgotten class appears among the top-5 ViT-L/16 predictions for only $1.0$--$5.6\%$ of these samples. 
\textbf{Retention (Fig.\,\ref{fig: gen_dit_retain}):} For the retained classes, our forget vector preserves the class identity and the visual quality of the pretrained model: goldfish, lions, giant pandas, balloons, and volcanoes remain clearly recognizable and similar in appearance to the original samples. Since $\bdelta_c$ is shared by all classes, the composition of some samples differs from that of the pretrained model, which is reflected in a moderate decrease of the RA, measured by ViT-L/16 on $10{,}000$ samples of random retained classes ($95.8$--$97.4\%$ \textit{vs.}\ $98.1\%$).

\textbf{SD v1.4.} \textbf{Forgetting (Fig.\,\ref{fig: gen_sd_i2p}):} On inappropriate I2P prompts, our forget vector removes the exposed body parts in all five cases shown, while largely preserving the subject, pose, and painting style of SD v1.4 ((b)--(e)). Among the baselines, several methods still generate exposed body parts (black boxes): ESD in (c), RECE and the negative prompt in (e), and SLD-Med in (a), (d), and (e). SalUn and AdvUnlearn remove nudity, but at the cost of replacing the subject or the entire scene, \textit{e.g.}, the abstract patterns of SalUn in (b) and the unrelated scenes of AdvUnlearn in (c) and (d).
\textbf{Retention (Fig.\,\ref{fig: gen_sd_coco}):} On benign COCO prompts, our forget vector leaves the composition and content of SD v1.4 almost unchanged, whereas several weight-editing methods alter them, \textit{e.g.}, SalUn generates a car instead of a motorcycle in (b), and AdvUnlearn and RECE generate a single wheel in (a). Accordingly, out forget attains the highest CLIP score among all unlearning methods ($26.41$ \textit{vs.}\ $26.60$ for SD v1.4 as shown in Tab.\,4 of the main paper).

Overall, our forget vector removes the target concept by redirecting the forgotten condition to natural content rather than to degenerate images or a fixed substitute. Although the same $\bdelta$ is also added to the inputs of the retained conditions, the generations for these conditions stay largely faithful to those of the original model in content and appearance, since the retain loss (Eq.\,(8) of the main paper) explicitly matches the perturbed predictions on retained inputs to those of the frozen original model.

\begin{figure*}[p]
\centering
\includegraphics[width=0.95\linewidth]{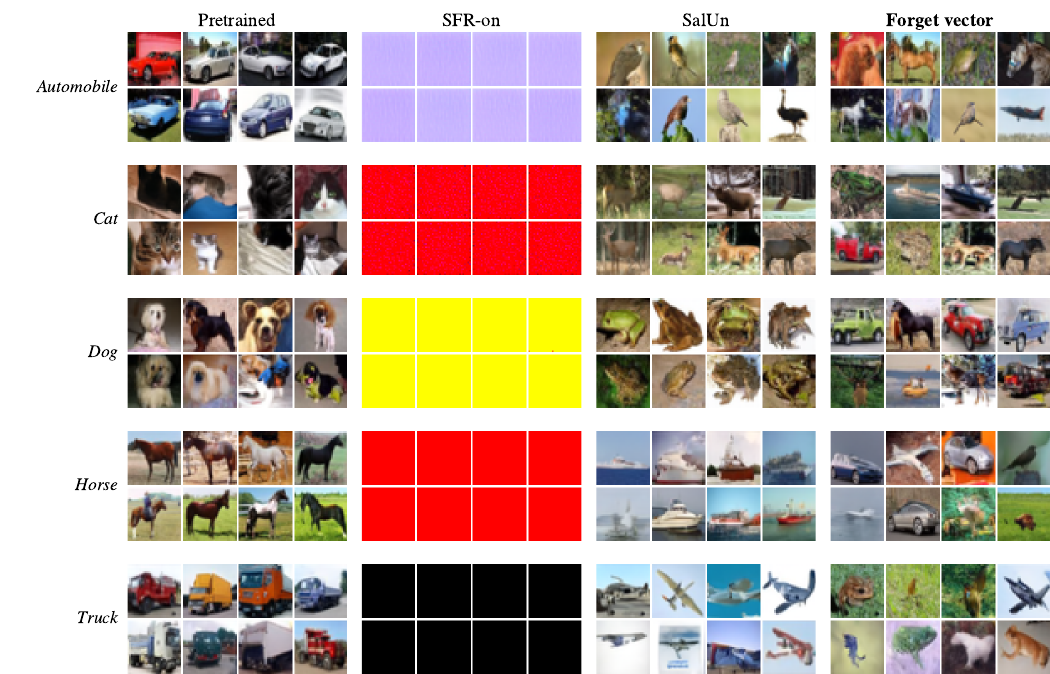}
\vspace{-0.1in}
\caption{Qualitative comparison of class-wise forgetting on the CIFAR-10 DDPM. Each row corresponds to one forgotten class, and all methods are sampled from identical initial noise. The columns show the pretrained model, SFR-on and SalUn, and our forget vector. 
SFR-on collapses to near-uniform color images for every class, and SalUn maps $97.0$--$99.2\%$ of the samples to a single substitute class (automobile$\to$bird, cat$\to$deer, dog$\to$frog, horse$\to$ship, truck$\to$airplane). In contrast, our forget vector produces natural images of diverse other classes, with the most frequent class accounting for only $18.0$--$27.8\%$ of the samples.}
\label{fig: gen_samples}
\vspace{-0.1in}
\end{figure*}

\begin{figure*}[htb]
\centering
\includegraphics[width=0.85\linewidth]{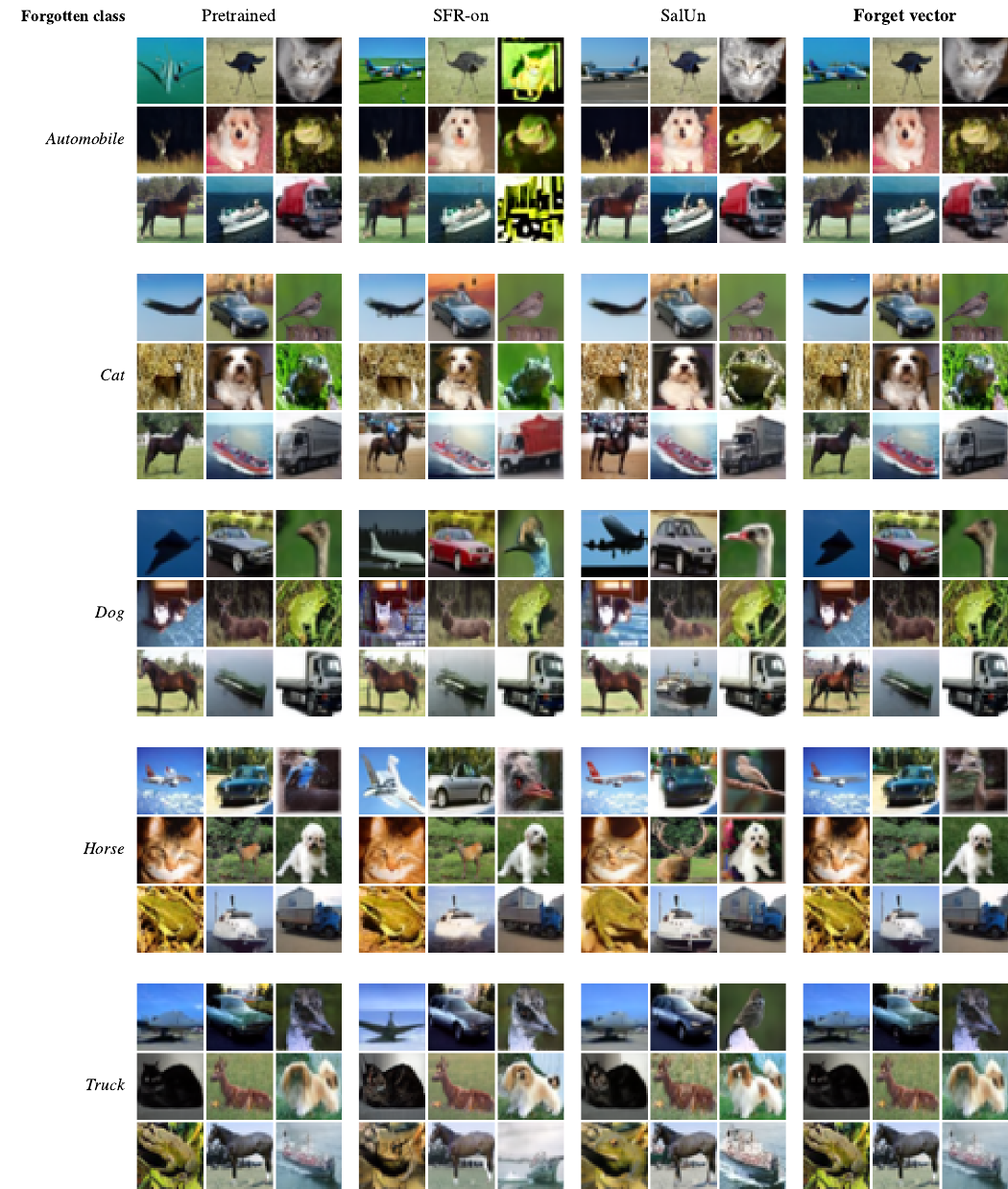}
\vspace{-0.1in}
\caption{Generation of retained classes on the CIFAR-10 DDPM after class-wise forgetting. Each row corresponds to one forgotten class, and each panel shows one sample of each of the nine retained classes ($3{\times}3$ grid in class order). All methods of a row are sampled from identical initial noise, and different rows use different noise. For each position, we show the sample with the higher ResNet-34 probability of the retained class among two samples of our forget vector, together with the samples of the other methods from the same initial noise. Our forget vector leaves most retained samples unchanged with respect to the pretrained model (mean absolute pixel difference of $0.023$ over the $45$ samples shown, \textit{vs.}\ $0.086$ for SFR-on and $0.085$ for SalUn), whereas SFR-on and SalUn alter the content of many samples, and SFR-on corrupts some samples when automobile is forgotten.}
\label{fig: gen_ddpm_retain}
\vspace{-0.1in}
\end{figure*}

\begin{figure*}[htb]
\centering
\includegraphics[width=0.95\linewidth]{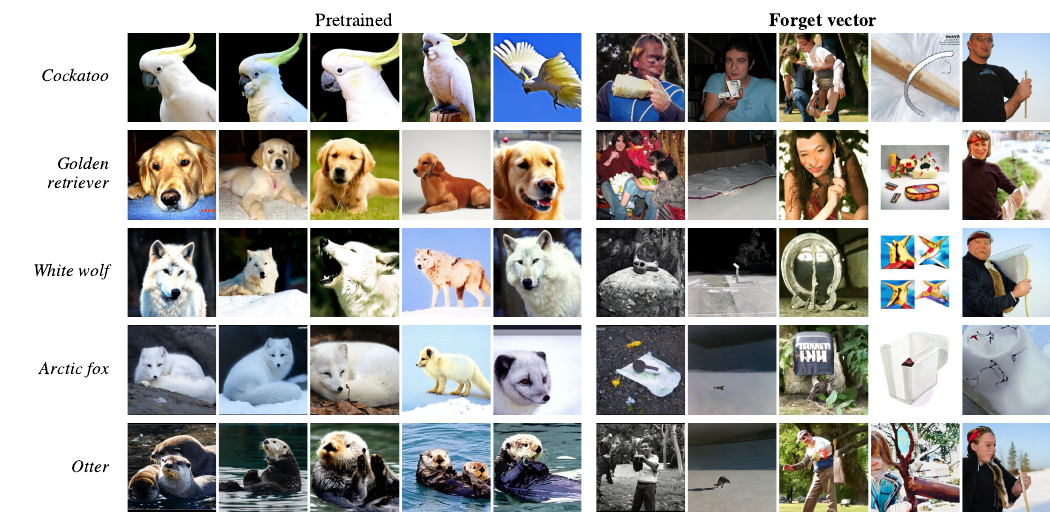}
\vspace{-0.1in}
\caption{Qualitative results of class-wise forgetting on DiT-XL/2. Each row corresponds to one forgotten class, and the pretrained model and our forget vector are sampled from identical initial noise. Our forget vector replaces the forgotten class by unrelated natural content: the class appears among the top-5 ViT-L/16 predictions for only $1.0$--$5.6\%$ of its samples, compared with $100\%$ for the pretrained model.}
\label{fig: gen_dit_samples}
\vspace{-0.1in}
\end{figure*}

\begin{figure*}[htb]
\centering
\includegraphics[width=0.95\linewidth]{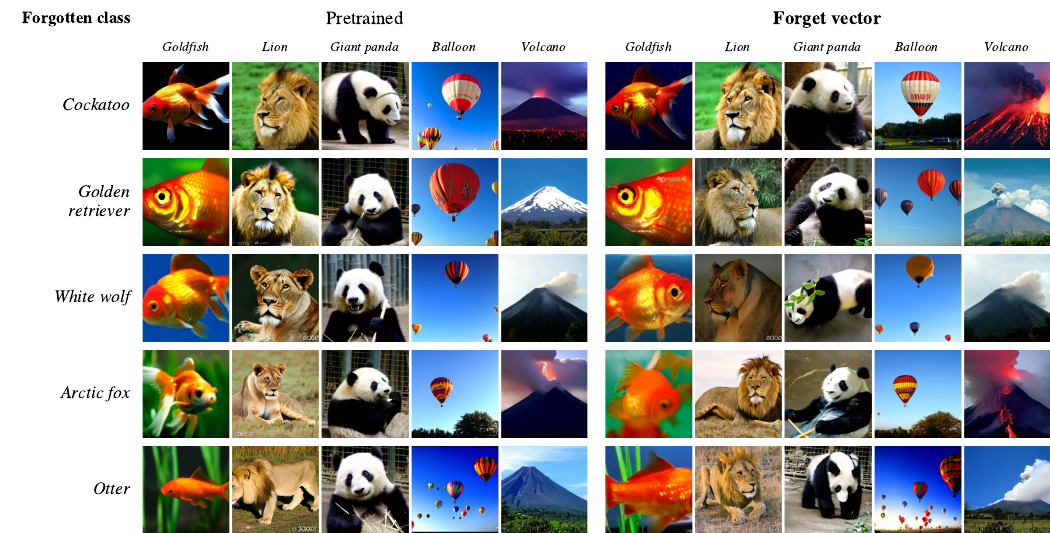}
\vspace{-0.1in}
\caption{Generation of retained classes on DiT-XL/2 after class-wise forgetting. Each row corresponds to one forgotten class, and each panel shows one sample of each of five retained classes (goldfish, lion, giant panda, balloon, and volcano). The pretrained model and our forget vector are sampled from identical initial noise. For each position, we show the sample with the highest ViT-L/16 probability of the retained class among eight samples of our forget vector, together with the sample of the pretrained model from the same initial noise. The retained classes remain recognizable and of high visual quality, although the composition of the images changes, since $\bdelta_c$ also perturbs the embeddings of the retained classes (RA, \textit{i.e.}, the top-1 accuracy of ViT-L/16 on $10{,}000$ samples of random retained classes, of $95.8$--$97.4\%$ \textit{vs.}\ $98.1\%$ for the pretrained model).}
\label{fig: gen_dit_retain}
\vspace{-0.1in}
\end{figure*}

\begin{figure*}[htb]
\centering
\includegraphics[width=0.95\linewidth]{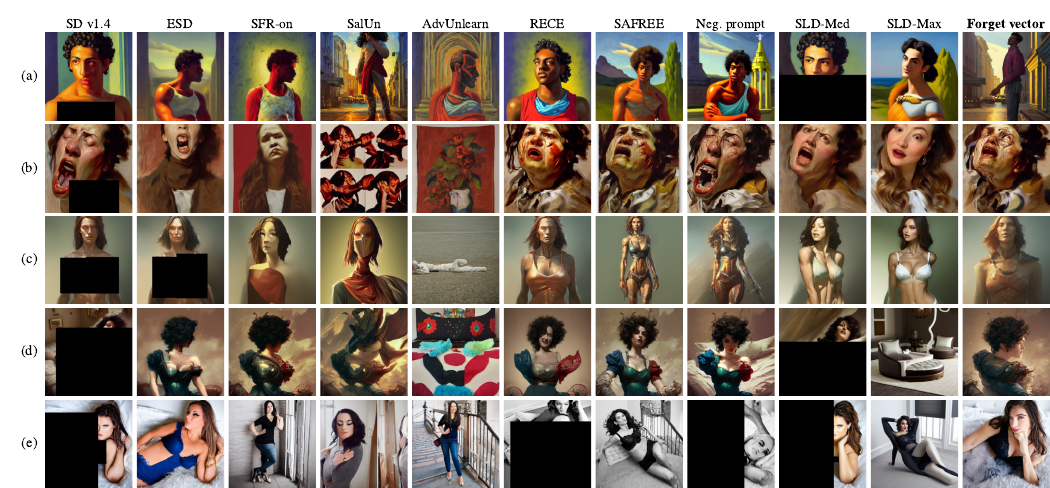}
\vspace{-0.1in}
\caption{Qualitative comparison on inappropriate I2P prompts in SD v1.4. The five prompts satisfy a fixed rule (a nudity percentage of at least $50\%$ and an exposed sensitive part generated by SD v1.4, and the prompts that name real persons or fictional characters, mention minors, describe violence, or describe several persons are excluded) and are chosen among the first such prompts in I2P order to illustrate successful removals by our forget vector. The detection rates over all I2P prompts are reported in Tab.\,4 of the main paper. All methods are sampled from identical initial noise, and exposed sensitive parts, detected by NudeNet or found in a visual check, are covered by black boxes. Our forget vector removes the exposed parts in all five cases while largely preserving the subject, pose, and style of SD v1.4 ((b)--(e)), whereas ESD, RECE, the negative prompt, and SLD-Med still generate exposed parts in some cases, and SalUn and AdvUnlearn often replace the subject or the entire scene.}
\label{fig: gen_sd_i2p}
\vspace{-0.1in}
\end{figure*}

\begin{figure*}[htb]
\centering
\includegraphics[width=0.95\linewidth]{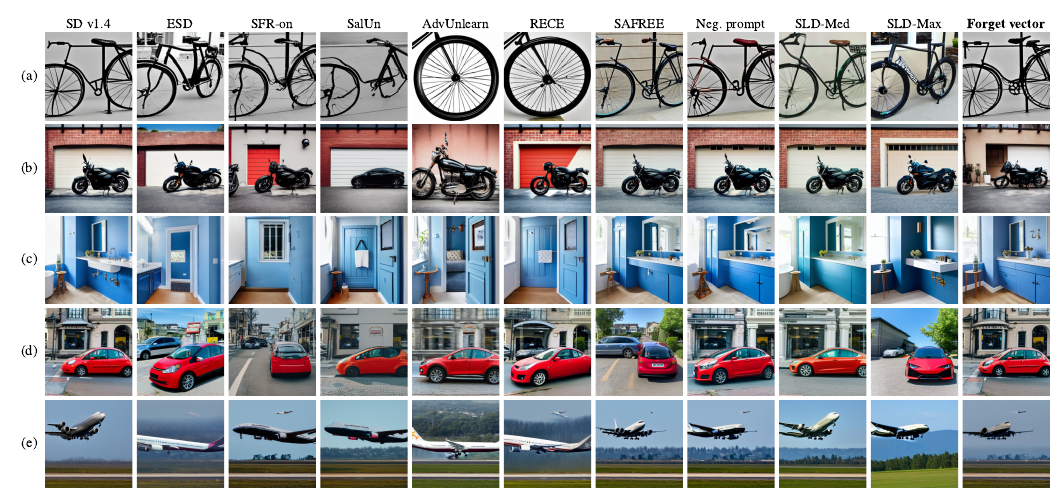}
\vspace{-0.1in}
\caption{Qualitative comparison on benign prompts in SD v1.4. Rows (a)--(e) correspond to the first five COCO evaluation prompts: (a) ``A bicycle replica with a clock as the front wheel.'', (b) ``A black Honda motorcycle parked in front of a garage.'', (c) ``A room with blue walls and a white sink and door.'', (d) ``A car that seems to be parked illegally behind a legally parked car'', and (e) ``A large passenger airplane flying through the air.''. All methods are sampled from identical initial noise. Our forget vector leaves the composition and content of SD v1.4 almost unchanged, whereas several weight-editing methods alter them, \textit{e.g.}, SalUn generates a car instead of a motorcycle in (b), and AdvUnlearn and RECE generate a single wheel in (a).}
\label{fig: gen_sd_coco}
\vspace{-0.1in}
\end{figure*}

\subsection{Efficiency Analysis}
\label{sec: gen_efficiency_sec}
Tab.\,\ref{tab: gen_efficiency} compares our forget vector with the generative unlearning baselines in terms of the parameters modified for an unlearning request, the storage needed per request, the training budget, the measured time and peak GPU memory of unlearning, and the additional cost at inference. \textbf{First}, our forget vector has $3{,}200$, $5{,}248$, and $75{,}520$ parameters for CIFAR-10 DDPM, DiT-XL/2, and SD v1.4, whereas the weight-editing baselines modify entire modules with $19.2$M to $860$M parameters.
Correspondingly, a forget vector needs $13$--$302$\,KB of storage per request, compared with $77$\,MB to $3.4$\,GB for a modified copy of the edited module. Saliency-based methods (SalUn, SFR-on) update only a subset of the weights, but still produce a full modified model. 
\textbf{Second}, our forget vector requires no additional denoiser evaluation at inference (Extra NFE of $0$ in Tab.\,\ref{tab: gen_efficiency}), since $\bdelta_x$ and $\bdelta_c$ are added to the inputs of the two existing evaluations of classifier-free guidance; the same holds for the weight-editing methods and the negative prompt. In contrast, SLD and SAFREE (with latent re-attention) require a third denoiser evaluation per sampling step, which increases the denoiser computation of every generated image by $50\%$. Training-free methods thus avoid the cost of unlearning (a time of $0$) at the expense of a permanent overhead at inference, whereas our forget vector incurs a one-time optimization cost and leaves the cost of inference unchanged.
\textbf{Third}, the measured time depends on both the training budget and the pre-processing. Each iteration of our forget vector back-propagates through the frozen denoiser, as weight-editing methods do, so that its time per step is of the same order as theirs. On CIFAR-10, our forget vector ($9.1$\,mins) is comparable to SalUn ($10.6$\,mins) and slower than SFR-on ($4.5$\,mins), which trains for only $50$ steps. On DiT and SD, it is the fastest among the trained methods ($6.7$ \textit{vs.}\ $10.7$--$48.0$\,mins and $23.4$ \textit{vs.}\ $53.2$--$86.5$\,mins), since the baselines train for more steps (SA and SalUn on DiT), perform more expensive iterations (on SD, $1.4$\,s per step for our forget vector \textit{vs.}\ $2.2$--$4.7$\,s for ESD, SalUn, and SFR-on, \textit{e.g.}, ESD samples an intermediate image in every iteration), or require a costly pre-processing of saliency masks or Fisher information (SalUn and SFR-on). 
\textbf{Fourth}, our forget vector has the lowest peak GPU memory in all three settings ($10.2$--$11.2$ \textit{vs.}\ $14.5$--$28.5$\,GB for the baselines), since it keeps no gradients or optimizer states of the model weights. The efficiency advantage of our forget vector in image generation thus lies in the number of parameters, the storage, the memory, and the inference cost, while its training time is comparable to or lower than that of the baselines, consistent with image classification (Sec.\,\ref{sec: efficiency analysis}).

\end{document}